\documentclass{article}

\makeatletter
\newcommand*{\addFileDependency}[1]{% argument=file name and extension
  \typeout{(#1)}
  \@addtofilelist{#1}
  \IfFileExists{#1}{}{\typeout{No file #1.}}
}
\makeatother

\usepackage{lipsum}
\usepackage{amsfonts}
\usepackage{graphicx}
\usepackage{epstopdf}
\usepackage[algo2e,ruled,linesnumbered]{algorithm2e}
\usepackage[flushleft]{threeparttable}
\usepackage{booktabs}
\usepackage{marginnote}

\newcommand{\flippedtodo}[1]{\let\marginpar\marginnote
  \reversemarginpar
  \renewcommand{\baselinestretch}{0.8}
  \todo{#1}}
\ifpdf
  \DeclareGraphicsExtensions{.eps,.pdf,.png,.jpg}
\else
  \DeclareGraphicsExtensions{.eps}
\fi
\newcommand{\email}[1]{{#1}}

\title{An AO-ADMM approach to constraining PARAFAC2 on all modes\thanks{\textbf{Funding:} This work was supported in part by the Research Council of Norway through project 300489 (IKTPLUSS) and ANR JCJC project LoRAiA ANR-20-CE23-0010.
}
}
\date{}

\author{
Marie Roald\thanks{Department of Data Science and Knowledge Discovery, Simula Metropolitan Center for Digital Engineering, Oslo, Norway \& Faculty of Technology, Art and Design, Oslo Metropolitan University, Oslo, Norway (\email{mariero@simula.no}, \email{carla@simula.no}).}
\and Carla Schenker\footnotemark[2]
\and Vince D. Calhoun\thanks{Department of Psychology, Georgia State University,  Atlanta, GA, USA (\email{vcalhoun@gsu.edu})}
\and T\"{u}lay Adal\i\thanks{Department of Computer Science and Electrical Engineering, UMBC, Baltimore, MD (\email{adali@umbc.edu}).}
\and Rasmus Bro\thanks{Department of Food Science, University of Copenhagen, Copenhagen, Denmark (\email{rb@food.ku.dk}).}
\and Jeremy E. Cohen\thanks{Univ Lyon, INSA-Lyon, UCBL, UJM-Saint Etienne, CNRS, Inserm,
CREATIS UMR 5220, U1206, F-69100 Villeurbanne, France (\email{jeremy.cohen@cnrs.fr}).}
\and Evrim Acar\thanks{Department of Data Science and Knowledge Discovery, Simula Metropolitan Center for Digital Engineering, Oslo, Norway (\email{evrim@simula.no}).}
}

\usepackage{amsopn}

\usepackage{subfiles}
\usepackage{subcaption}
\usepackage{todonotes}
\usepackage[a4paper, margin=1.571in]{geometry}

\usepackage{amsmath}
\usepackage{amssymb}
\usepackage{amsthm}

\usepackage{stmaryrd}

\usepackage[mathscr]{eucal}

\usepackage{amsbsy}

\usepackage{bm}

\usepackage{bbding}

\usepackage{array}

\usepackage{tabularx}

\usepackage[neveradjust]{paralist}

\usepackage{fancyvrb}

\usepackage{caption}

\usepackage{placeins}

\usepackage{ifpdf}
\ifpdf
\DeclareGraphicsExtensions{.pdf,.png}
\else
\DeclareGraphicsExtensions{.eps}
\fi

\usepackage{ulem}
\usepackage{xcolor}

\usepackage[hyperfootnotes=false]{hyperref}
\usepackage[capitalize,noabbrev]{cleveref}

\newcommand{\Real}{{\mathbb R}}

\newcommand{\All}{\emph{:}}

\newcommand{\Tra}{^{{\sf T}}} % transpose
\newcommand{\V}[1]{{\bm{\mathbf{\MakeLowercase{#1}}}}} % vector
\newcommand{\Oprod}{\circ} % outer product

\newcommand{\M}[1]{{\bm{\mathbf{\MakeUppercase{#1}}}}} % matrix
\newcommand{\MC}[2]{\V{#1}_{#2}} % matrix column
\newcommand{\MR}[2]{\V{#1}_{#2 \All}} % matrix row
\newcommand{\Mn}[2]{\M{#1}^{(#2)}} % n-th matrix
\newcommand{\MnTra}[2]{\M{#1}^{(#2){{\sf T}}}} % n-th matrix transpose
\newcommand{\Khat}{\odot} %Khatri-Rao
\newcommand{\T}[1]{\boldsymbol{\mathscr{\MakeUppercase{#1}}}} %tensor
\newcommand{\TFS}[2]{\M{#1}_{{#2}}} % tensor frontal slice
\newcommand{\norm}[1]{\left\lVert \, #1 \, \right\rVert}
\newcommand{\fnorm}[1]{\left\lVert \, #1 \, \right\rVert_{F}}

\usepackage{optidef}
\Crefname{algocf}{Algorithm}{Algorithms}

\DeclareMathOperator*{\argmin}{arg~min}

\DeclareMathOperator{\st}{subject~to}

\newcommand{\X}{\T{X}}

\newcommand{\FMS}{\text{FMS}}
\newcommand{\FMSA}{\text{FMS}_\A}
\newcommand{\FMSB}{\text{FMS}_\B}
\newcommand{\FMSC}{\text{FMS}_\C}

\newcommand{\CWSNR}[1]{\text{cwSNR}_{#1}}

\renewcommand{\Mn}[2]{\M{#1}_{#2}} % n-th matrix
\renewcommand{\MnTra}[2]{\M{#1}_{#2}^{\sf T}} % n-th matrix transpose
\newcommand{\proxF}[1]{\text{prox}_{#1}}

\newcommand{\prox}[2]{\proxF{#1}\hspace{-0.2em}\left(#2\right)}

\newcommand{\DiagEntries}[1]{\text{Diag}\left( #1 \right)}

\newcommand{\Trace}[1]{\text{Tr}\left( #1 \right)}

\newcommand{\vectorize}[1]{\text{vec}\left(#1\right)}

\newcommand{\ForAllK}[1]{\left\{ #1 \right\}_{k \leq K}}

\newcommand{\A}{\M{A}}
\newcommand{\Bk}{\Mn{B}{k}}
\newcommand{\BAll}{\ForAllK{\Bk}}
\newcommand{\C}{\M{C}}

\newcommand{\Dk}{\Mn{D}{k}}
\newcommand{\DAll}{\ForAllK{\Dk}}
\newcommand{\Pk}{\Mn{P}{k}}
\newcommand{\PAll}{\ForAllK{\Mn{P}{k}}}
\newcommand{\blueprint}{\M{\Delta_B}}
\newcommand{\B}{\M{B}}

\newcommand{\Aux}[1]{\M{Z}_{#1}}

\newcommand{\AAux}{\Aux{\A}}
\newcommand{\BkAux}{\Aux{\Bk}}

\newcommand{\BkConstraint}{\M{Y}_{\Bk}}
\newcommand{\BAllConstraint}{\ForAllK{\BkConstraint}}

\newcommand{\DkAux}{\Aux{\Dk}}

\newcommand{\DualVar}[1]{\M{\mu}_{#1}}
\newcommand{\ADual}{\DualVar{\A}}
\newcommand{\BkDual}{\DualVar{\BkAux}}

\newcommand{\blueprintkDual}{\DualVar{\blueprint_k}}

\newcommand{\DkDual}{\DualVar{\Dk}}

\newcommand{\PSet}{\mathscr{P}}
\newcommand{\Ortho}[2]{\mathcal{O}_{#1, #2}}

\newcommand{\Xk}{\Mn{X}{k}}

\newcommand{\IndicatorF}[1]{\iota_{#1}}
\newcommand{\ConstraintIndicator}[1]{\iota_{\PSet} \left( #1 \right)}
\newcommand{\ConstraintIndicatorF}{\iota_{\PSet}}

\newcommand{\regF}[1]{g_{#1}}
\newcommand{\reg}[2]{\regF{#1}\left( #2 \right)}
\newcommand{\lossF}[1]{f_{#1}}
\newcommand{\loss}[2]{\lossF{#1}\left( #2 \right)}

\newtheorem{theorem}{Theorem}
\numberwithin{theorem}{section}
\newtheorem{proposition}{Proposition}
\numberwithin{proposition}{section}
\ifpdf
\hypersetup{
  pdftitle={An AO-ADMM approach to constraining PARAFAC2 on all modes},
  pdfauthor={M. Roald, C. Schenker, R. Bro, V. D. Calhoun, T. Adali, J. E. Cohen, and E. Acar}
}
\fi

\usepackage[style=ieee]{biblatex}
\begin{document}

\maketitle

\begin{abstract} Analyzing multi-way measurements with variations across one mode of the dataset is a challenge in various fields including data mining, neuroscience and chemometrics. For example, measurements may evolve over time or have unaligned time profiles. The PARAFAC2 model has been successfully used to analyze such data by allowing the underlying factor matrices in one mode (i.e., the evolving mode) to change across slices. The traditional approach to fit a PARAFAC2 model is to use an alternating least squares-based algorithm, which handles the constant cross-product constraint of the PARAFAC2 model by
implicitly estimating the evolving factor matrices.
This approach makes imposing regularization on these factor matrices challenging. There is currently no algorithm to flexibly impose such regularization with general penalty functions and hard constraints. In order to address this challenge and to avoid the implicit estimation, in this paper, we propose an algorithm for fitting PARAFAC2 based on alternating optimization with the alternating direction method of multipliers (AO-ADMM). 
With numerical experiments on simulated data, we show that the proposed PARAFAC2 AO-ADMM approach allows for flexible constraints, recovers the underlying patterns accurately, and is computationally efficient compared to the state-of-the-art. We also apply our model to two real-world datasets from neuroscience and chemometrics, and show that constraining the evolving mode improves the interpretability of the extracted patterns.
\end{abstract}

\section{Introduction}
For many applications in different domains, measurements are obtained in the form of sequences of matrices, which can be arranged as a third-order tensor. Tensor decomposition, which is the higher-order extension of matrix decomposition, is a well-known and useful tool for analysis of such multi-way data. A popular tensor decomposition model is the CANDECOMP/PARAFAC (CP) decomposition \cite{CaCh70,Ha70}, also known as the canonical polyadic decomposition \cite{Hi27a} which describes a tensor as the sum of a minimum number of rank-one components, i.e., an \(R\)-component CP model of the tensor \(\T{X} \in \Real^{I \times J \times K}\), is given by
\begin{equation}
    \T{X} \approx \sum_{r=1}^R \MC{A}{r} \Oprod\MC{B}{r} \Oprod\MC{C}{r}, \label{eq:cp.oprod} 
\end{equation}
where \(\MC{A}{r}, \MC{B}{r}\) and \(\MC{C}{r}\) are the \(r\)-th column of the factor matrices \(\A \in \Real^{I \times R}, \B \in \Real^{J \times R}\) and \(\C \in \Real^{K \times R}\). These factor matrices can be interpreted as a collection of underlying patterns. Following \eqref{eq:cp.oprod}, each frontal slice, \(\Xk\) of \(\T{X}\) can be represented as:
\begin{equation}
    \M{X}_k \approx \A \Dk \B\Tra, \label{eq:cp}
\end{equation}
where each \(\Dk \in \Real^{R \times R}\) is diagonal with diagonal entries given by the \(k\)-th row of \(\C\). The CP model has successfully extracted meaningful patterns from data tensors in various fields including chemometrics \cite{Br97}, neuroscience \cite{MoHaHePaAr06,AcBiBiBr07,WiKiWa18} and social network analysis \cite{AcDuKo10b} (see surveys on tensor factorizations for more applications \cite{KoBa09, AcYe09, PaFaSi16}).

\begin{figure}
    \centering
    \includegraphics[trim=0 0.90cm 0 0.5cm,clip]{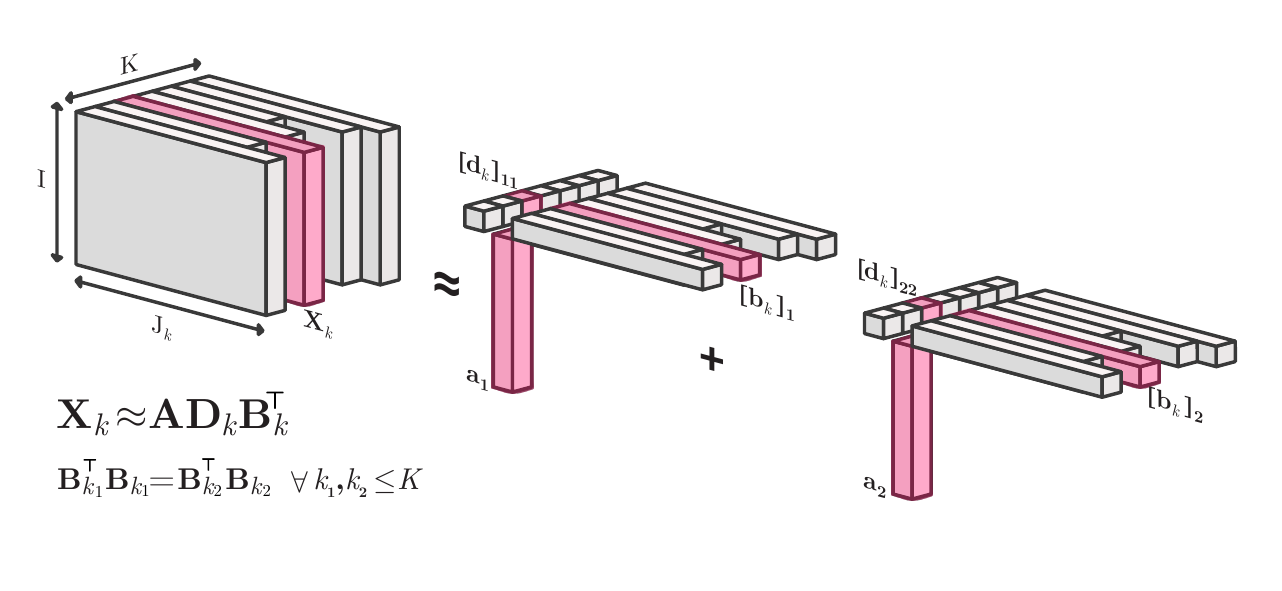}
    \caption{Illustration of a two-component PARAFAC2 model for a ragged tensor.}
    \label{fig:parafac2}
\end{figure}

However, CP cannot describe patterns that vary across one mode without following the strict assumption in (\ref{eq:cp}). For such evolving patterns, the more general PARAFAC2 model \cite{Ha72} is better suited as it models the slices of a tensor by
\begin{equation}
\Xk \approx \A \Dk \Bk\Tra,
\end{equation}
where  \(\Dk \in \Real^{R \times R}\) is diagonal, \(\A \in \Real^{I \times R}, \Bk \in \Real^{J \times R}\) and \(\smash{\MnTra{B}{k_1}\Mn{B}{k_1}} = {\MnTra{B}{k_2}\Mn{B}{k_2}}\) for all \(k_1, k_2 \leq K\).
Thus, PARAFAC2 allows the \(\Bk\) factor matrices to vary across tensor slices, under the constraint that their cross product is constant, which is more general than CP, where \(\Mn{B}{k_1} = \Mn{B}{k_2}\). Letting \(\Bk\) vary in this way also enables decomposing ragged tensors (i.e., stacks of matrices with varying size, as illustrated in \cref{fig:parafac2}).

Letting factor matrices evolve across one mode has proven advantageous for applications in many domains. PARAFAC2 has, for example, shown an exceptional ability to analyze chromatographic data with unaligned elution profiles \cite{BrAnKi99}. The PARAFAC2 model has also been successfully applied to resolve unaligned temporal profiles in electronic health records \cite{AfPePaSeHoSu18} and to retrieve information across languages from a multi-language corpus \cite{ChBaBr07}. In neuroscience, PARAFAC2 has been used to model functional connectivity in functional Magnetic Resonance Imaging (fMRI) data \cite{MaChMo17}, for tracing time-evolving networks of brain connectivity \cite{RoShJi19} and for jointly analyzing fMRI data from multiple tasks \cite{LeAc22}.

Online (also referred to as incremental or adaptive) tensor methods can also be used to capture evolving components, e.g., through updates of the CP model \cite{NiSi09,VaVeLa17}, in particular, in the context of streaming data. While related, modelling assumptions of online tensor methods vs. the PARAFAC2 model are different. Online methods update the whole model, i.e., factor matrices in all modes, as new data slices arrive at new time points, for instance through efficient stochastic gradient descent (SGD)-based approaches \cite{MaMaGi15, MaHaKa16, KoHo20}. On the other hand, PARAFAC2 jointly analyzes data slices summarizing them all using a common $\M{A}$ factor matrix (e.g., when a \emph{subjects} by \emph{voxels} by \emph{time windows} fMRI tensor is analyzed using PARAFAC2, the same subject coefficients, $\M{A}$, are assumed for all time windows \cite{RoShJi19}) that can only change up to a scaling across different slices while letting $\M{B}$ factor matrices change from one slice to another. The distinction between modelling assumptions becomes even more clear when multi-task fMRI data (in the form of a \emph{subjects} by \emph{voxels} by \emph{tasks} tensor) is analyzed. PARAFAC2 assumes the same subject coefficients and different voxel-mode (evolving) components for each task \cite{LeAc22} whereas online tensor methods are not suitable for the analysis of such data.

One challenge with PARAFAC2 is that the traditional alternating least squares (ALS) approach for fitting the model handles its constant cross-product constraint by implicitly estimating the evolving \(\Bk\) factor matrices via the reparametrization \(\Bk = \Pk \blueprint\) with \(\Pk\Tra\Pk = \M{I}\) \cite{KiTeBr99}. This reparametrization makes it difficult to impose constraints on the evolving \(\Bk\)-matrices in a flexible way. Nevertheless, constraints are essential to obtain uniqueness and interpretability in matrix factorizations \cite{WaZh12}. Additionally, constraints and regularization can improve the interpretability of components obtained from CP models \cite{BrJo97,FrHa08}, and the non-evolving modes of PARAFAC2 models \cite{AfPePaSeHoSu18}. Recent studies have also demonstrated the benefits of constraining the evolving mode of PARAFAC2  \cite{He17,CoBr18,AfPePaSeHoSu18,BeKeGiDe20}.
One way to constrain the \(\Bk\)-matrices is by specifying a linear subspace their columns should be contained in. In \cite{He17}, Helwig showed that the data can be preprocessed by projecting it onto the subspace of interest, before fitting the PARAFAC2 model and that this approach can constrain the evolving components to be spanned by a B-spline basis, thus also constraining them to be smooth \cite{He17}.
However, a downside of this scheme is that it only supports linear constraints and that we need to know the linear subspace (e.g., the spline knots) of the components a priori, which may be difficult in practice. Another approach to have constraints on \(\Bk\)-matrices, in particular, non-negativity constraints, is the flexible coupling approach \cite{CoBr18}. This approach fits a non-negative coupled matrix factorization model with a regularization term based on the PARAFAC2 constraint. While this approach, in theory, can be used with any constrained least squares solvers, the algorithm in \cite{CoBr18} uses hierarchical alternating least squares (HALS) and supports only non-negativity constraints. Other notable approaches include LogPar \cite{YiAfHoChZhSu20} using another regularization penalty based on PARAFAC2 to improve uniqueness properties of regularized coupled non-negative matrix factorization for binary data, and CANDELINC2  \cite{BeKeGiDe20}, which uses a scheme combining CANDELINC (Canonical Decomposition with Linear Constraints) \cite{CaPrKr80,BrAn98,Ki98} and non-negativity constraints for PARAFAC2. Despite these recent efforts, available methods for PARAFAC2 are still limited in terms of the type of constraints they can impose.

In this paper, we propose fitting PARAFAC2 using an alternating optimization (AO) scheme with the alternating direction method of multipliers (ADMM) to facilitate the use of a wider set of constraints in all modes (both evolving and non-evolving). Recently, Huang et al. introduced the AO-ADMM scheme for constrained CP models \cite{HuSiLi16}, and Schenker et al. extended this framework to regularized linearly coupled matrix-tensor decompositions \cite{ScCoAc20,ScCoAc20b}. AO-ADMM has been successfully used to impose proximable constraints on the non-evolving factor matrices of the PARAFAC2 model \cite{AfPePaSeHoSu18,ReLoXiHo20}. However, using that approach, evolving factor matrices are still modelled implicitly by \(\Bk = \Pk \blueprint\) as in \cite{KiTeBr99}, with AO-ADMM being used for the CP step of the algorithm. Thus, only linear constraints can be imposed directly on the evolving mode, still limiting the possible constraints on the evolving factor matrices. Unlike earlier studies, we present an AO-ADMM based framework for the PARAFAC2 model with ADMM updates that widen the possible constraints to any proximable penalty function in all modes. With numerical experiments on both simulated and real data, we demonstrate the performance of this framework in terms of  

\begin{itemize}
    \item \textbf{Flexibility:} The scheme allows for regularization of the evolving components with any proximable penalty function. 
    \item \textbf{Efficiency:} The scheme can fit non-negative PARAFAC2 models faster than the flexible coupling with HALS approach introduced in \cite{CoBr18}.
    \item \textbf{Accuracy:} Applying suitable constraints for the evolving components can improve performance compared to only constraining the non-evolving modes.
\end{itemize}

This paper extends our preliminary study \cite{RoScCoAc21}, where we introduced and showed the promise of AO-ADMM for constraining PARAFAC2 models using non-negativity constraints, graph Laplacian and total variation regularization. In this paper, we provide a more detailed derivation and a theoretical discussion of both the AO-ADMM scheme and the regularized PARAFAC2 problem. We include extensive numerical experiments using more general simulation setups and one new constraint (unimodal component vectors). Finally, we demonstrate the usefulness of the proposed algorithmic approach on two applications from neuroimaging and chemometrics.

\Cref{sec:notation} briefly states the notation used in this paper,
and \cref{sec:parafac2.intro} introduces PARAFAC2 and states some observations on the PARAFAC2 constraint and available methods to fit PARAFAC2 models. In \cref{sec:parafac2.aoadmm}, we introduce the new AO-ADMM scheme that enables regularization on all modes.
Numerical experiments evaluating the proposed scheme on seven simulation setups and two real datasets are presented in \cref{sec:experiments}. Finally, \cref{sec:discussion} discusses our results and possible future work.

\section{Notation}\label{sec:notation}
We specify some key tensor concepts and notations here and refer to \cite{KoBa09} for a thorough review of the nomenclature and background theory. A tensor can be considered a multidimensional array that extends the concept of matrices to higher dimensions. The dimensions of a tensor are called modes, and the number of modes is referred to as the order. Thus, vectors are first-order tensors, matrices are second-order tensors, ``cubes'' of numbers are third-order tensors and so forth. Tensors with three modes or more are often called \emph{higher-order tensors}. We indicate vectors by bold lowercase letters, e.g., \(\V{V}\), matrices by bold uppercase letters, e.g., \(\M{M}\), and higher-order tensors by bold uppercase calligraphic letters, e.g., \(\T{X}\). Furthermore, we denote the  \(r\)-th column of a matrix,  \(\M{X}\), by \(\V{x}_{r}\) and the \(r\)-th column of a matrix \(\Mn{X}{k}\) by \(\left[\smash{\V{x}}_k\right]_{r}\). Sets of matrices are denoted by calligraphic letters (e.g., \(\Ortho{n}{m} = \{\M{P} \in \Real^{n \times m} : \M{P}\Tra\M{P} = \M{I}\}\)) and sets of sets of matrices are denoted by script letters (e.g., \(\PSet\)). 

All tensors we consider in this work have three modes or less\footnote{But our work extends to any order trivially. For orders above three, the PARAFAC2 constraint is not needed for uniqueness \cite{BrAnKi99} since a fourth-order PARAFAC2 model is equivalent to a coupled CP model. Still, the PARAFAC2 constraint can be used to enforce a constant relationship between the components for each frontal slab (higher-order slice) of the data tensor.}. The Khatri-Rao product (column-wise Kronecker product) is denoted by \(\Khat\), the Hadamard product (element-wise multiplication) is denoted by \(*\), the vector outer product is denoted by \(\Oprod\) and the Frobenius norm is denoted by \(\fnorm{\cdot}\). The matrix inverse and transpose are denoted by \(^{-1}\) and \(\Tra\), respectively. The number of components for tensor decomposition models is denoted by \(R\), and \(I, J\) and \(K\) denote the number of horizontal, lateral and frontal slices in a tensor. For ragged tensors (i.e., stacks of matrices with varying number of columns), \(J_k\) represents the number of columns for the \(k\)-th slice.

\section{PARAFAC2 \& ALS}\label{sec:parafac2.intro}
Recall that the PARAFAC2 model represents \(\TFS{X}{k}\) as: 
\begin{equation}
    \TFS{X}{k} \approx \M{A} \Dk \Bk\Tra, \label{eq:pf2}
\end{equation}
where the \(\Bk \in \Real^{J \times R}\) matrices follow the \emph{PARAFAC2 constraint}. Specifically, \(\smash{\MnTra{B}{k_1}\Mn{B}{k_1}} = {\MnTra{B}{k_2}\Mn{B}{k_2}}\) for all \(k_1, k_2 \leq K\) (henceforth, \(\BAll \in \PSet\)). If one considers the columns of the factor matrices \(\BAll\) as some underlying patterns or sources underlying measurements, then PARAFAC2 allows these patterns to rotate and mirror for each slice \(k\) while the cross product between these patterns stays constant. 
Moreover, the constant cross-product constraint is equivalent to a constant correlation constraint if the columns of the \(\Bk\) matrices sum to zero (which can be ensured using the method in \cite{He17}). 
PARAFAC2 has also been shown to extract meaningful patterns, even when the data does not fully follow the PARAFAC2 constraint \cite{BrAnKi99,RoShJi19}.
The next section states some further mathematical properties of this constraint.

\subsection{Observations on the PARAFAC2 constraint} \label{sec:parafac2.observations}

The set
\begin{equation}
    \PSet = \{\{\Bk \}_{k=1}^{K}~|~ \MnTra{B}{k_1}\Mn{B}{k_1} = {\MnTra{B}{k_2}\Mn{B}{k_2}} \; \forall k_1, k_2 \leq K \}
\end{equation}
defines the PARAFAC2 constraint, and restricts the angle between component vectors of \(\Bk\) to be constant for all \(k\). Without this constraint, the PARAFAC2 model would be equivalent to the rank \(R\) matrix factorization of the stacked matrix \( [\TFS{X}{1}, \ldots, \TFS{X}{K} ] \) whose parameters are not identifiable in general.
The PARAFAC2 constraint ensures unique components given enough frontal slices, \(\Xk\) \cite{HaLu96,TeJoKi96,KiTeBr99}.
Despite its importance, the set \( \PSet \) and its properties have received little attention in the PARAFAC2 literature.

Below, we make a few key observations on the PARAFAC2 constraint. First, as shown by Kiers et al.\cite{KiTeBr99}, given a collection of matrices \( \{\Bk \}_{k=1}^{K} \in \PSet \), there exists a matrix \( \blueprint \in \Real^{R\times R} \) and a collection \( \{ \Pk\}_{k=1}^{K} \) of orthogonal matrices such that 
\begin{equation}
    \forall k\leq K, \; \Bk = \Pk\blueprint.
\end{equation}
This means that \(\PSet \) is related to \( \Ortho{J_1}{R} \times \ldots \times \Ortho{J_K}{R} \times \Real^{R\times R}  \). We use this fact in the supplementary material to prove that \(\PSet\) is closed.
\begin{proposition}\label{prop:pset.closed}
    The set \(\PSet\) is closed.
\end{proposition}
As a direct corollary, the projection on \( \PSet \) is well-defined and generally unique \cite{FrOt14}. Projecting on \( \PSet\) will be of importance in \cref{sec:parafac2.aoadmm}.

Second, it may also be noted that \(\PSet \) is the set of all discretizations with \(K \) points of elements in \( \Omega = \{\Ortho{R}{R}\blueprint | \blueprint\in\Real^{R\times R} \} \)\footnote{One may assume \textit{w.l.o.g.} that \( J_k=R \), up to dimensionality reduction of each slice \(\TFS{X}{k}\)}, where \( \Ortho{R}{R}\blueprint = \{ \Bk\in\Real^{R\times R} ~|~ \exists \Pk\in \Ortho{R}{R},\; \Bk = \Pk\blueprint \} \). Studying this continuous set \( \Omega \) rather than the discrete \(\PSet \) provides another interesting point of view. Indeed, projecting on the PARAFAC2 constraint also means finding the set \( \Ortho{R}{R}\blueprint \) which is the closest to all matrices \( \Bk \).  In fact, noticing that the point-set distance satisfies
\begin{equation}
    d(\Bk,\Ortho{R}{R}\blueprint ) = \min_{\Pk\in\Ortho{R}{R}} \|\Bk - \Pk\blueprint  \|_F^2 = d(\blueprint, \Ortho{R}{R}\Bk),
\end{equation} one may rewrite the projection problem on \( \PSet \) as the following minimization problem:
\begin{equation}
    \min_{\blueprint} \sum_{k=1}^{K} d(\blueprint, \Ortho{R}{R}\Bk).
\end{equation}
Intuitively, projecting on the PARAFAC2 constraint therefore amounts to finding the closest point \( \blueprint \) to a set of curves \( \Ortho{R}{R}\Bk \). Because the projection on \(\PSet \) is in fact a bilevel optimization problem, this justifies the empirical use of an alternating algorithm to compute the projection on \( \PSet\) which is detailed in \cref{sec:parafac2.aoadmm}. We also hope this geometric point of view can lead to a better understanding of the PARAFAC2 constraint. A striking example is the rank one case \(R=1 \), where \( \Omega\) simply becomes the set of all spheres, and projecting on \( \PSet \) amounts to projecting each \( \Bk \) on the average sphere of radius \(\frac{1}{K}\sum_{k=1} \|\Bk\|_2\). Some illustrations and the computations for the rank one case are provided in the supplementary materials.

\subsection{Recoverability of the evolving components} \label{sec:pf2.recoverability}
While the PARAFAC2 model has many similarities with the CP model, it comes with some extra challenges. One notable challenge is the additional sign indeterminacy of the PARAFAC2 model. Specifically, for any component, \(r\), we can reverse the sign of \([d_k]_{r r}\) together with \([\V{b}_k]_r\), so long as the corresponding signs for all components, \(s\), that satisfy \([\V{b}_k]_r\Tra [\V{b}_k]_{s} \neq 0\) are also reversed \cite{Ha72,BrLeJo13}. There are several ways to handle this sign indeterminacy, e.g. by imposing non-negativity constraints on \(\DAll\) \cite{Ha72}, aligning the components with the direction of the data \cite{BrLeJo13} or by using additional information \cite{He13}. 

Additionally, for real data matrices that are affected by noise, the recoverability of the \(r\)-th column of \(\Bk\) is directly affected by the magnitude of the \(r\)-th diagonal entry in \(\Dk\). To see this, consider estimating \(\Bk\) from noisy data \(\tilde{\M{X}}_k = \Xk + \M{E}_k\) with \(\Xk = \A \Dk \Bk\Tra\) and known \(\A\) and \(\Dk\). Using the normal equation for \(\Bk\), we get
\begin{equation}
    \hat{\M{B}}_k \Tra = \Dk^{-1} \left(\A\Tra \A\right)^{-1} \A\Tra (\Xk + \Mn{E}{k}) = \Bk\Tra + \Dk^{-1} \left(\A\Tra \A\right)^{-1} \A\Tra\Mn{E}{k}.
\end{equation}
This means that the estimate of the \(r\)-th column of \(\Bk\), \([\hat{\V{b}}_k]_r\), is affected by the noise scaled by \(1/[d_k]_{rr}\) (for simplicity, we omit \(\left(\smash{\A\Tra \A}\right)^{-1} \A\), since it is constant for all \(k\)). Thus, if an element of \(\Dk\) is small compared to the noise strength \(\norm{\Mn{E}{k}}_F\), then the corresponding column of \(\Bk\) will be poorly estimated.

To quantify the effect of the noise on recovering a column of \(\Bk\), we define the \emph{column-wise signal-to-noise ratio} (cwSNR), defined by
\begin{equation}
    \CWSNR{kr} = \frac{[d_k]_{rr}^2}{\norm{\Mn{E}{k}}^2},
\end{equation}
with the factor matrices scaled so \(\A\) and \(\Bk\) have unit norm columns. We show that the cwSNR is a good predictor on the accuracy of \(\Bk\) estimates in \cref{sec:cwsnr.experiment}.

\subsection{ALS for PARAFAC2 (unregularized)} \label{sec:pf2.als}
There is no closed form solution for fitting PARAFAC2 models. Instead, one has to solve a difficult, non-convex, optimization problem. This is usually done by estimating the solution using \emph{alternating optimization} (AO), which, instead of solving the optimization problem directly, fixes all but one factor matrix in each step. This single factor matrix can then be updated efficiently by solving a quadratic surrogate problem.

We cannot easily use AO directly on \cref{eq:pf2}, as the PARAFAC2 constraint leads to a non-convex optimization problem for the \(\Bk\) updates. However, as discussed in \cref{sec:parafac2.observations}, the PARAFAC2 model can be reformulated as:
\begin{equation}
    \TFS{X}{k} \approx \A \Dk \blueprint\Tra \Pk\Tra, \label{eq:pf2.kiers}
\end{equation}
where \(\blueprint \in \Real^{R \times R}\) and \(\Pk \in \Ortho{J_k}{R}\). This problem can be solved efficiently using an alternating least squares procedure. In this case, the \(\Pk\) updates are performed by solving an orthogonal Procrustes problem, resulting in \cref{alg:pf2.als} \cite{KiTeBr99}.

\begin{algorithm2e}
\DontPrintSemicolon
\SetAlgoLined
\KwResult{\(\A, \ForAllK{\Pk}, \blueprint, \DAll\)}
 Initialize \(\A\), \(\blueprint\) and \(\DAll\)\;
 \While{stopping conditions are not met and max no. iterations not exceeded}{
  
  \For{\(k \gets 1\) \KwTo \(K\)}{
    Compute a rank \(R\) trucated SVD: \(\Mn{U}{k} \Mn{S}{k} \Mn{V}{k} \Tra = \Xk \Tra \A \Dk \blueprint \Tra\) \;
     \( \Pk \gets \Mn{U}{k} \Mn{V}{k} \Tra \) \;
     \( \Mn{T}{k} \gets \Xk \Pk \) \;
   }
   Estimate \(\A, \blueprint \) and \(\DAll\) by running a few iterations (e.g., 5) of ALS to fit an \(R\)-component CP model to the tensor \(\T{T}\) with frontal slices given by \(\Mn{T}{k}\). \label{alg.line:pf2cpstep}}
 \caption{PARAFAC2 ALS \cite{KiTeBr99}}
 \label{alg:pf2.als}
\end{algorithm2e}

\subsection{Constrained PARAFAC2 with flexible coupling}
The ALS formulation of PARAFAC2 does not lend itself for constraining or regularizing the \(\Bk\) components. Currently the only way to impose non-negativity on these components is with the \emph{flexible coupling approach} by Cohen and Bro \cite{CoBr18}. This approach works by fitting a coupled matrix factorization with a regularization term that penalizes the distance between \(\BAll\) and a point on a curve \( \Ortho{R}{R}\blueprint \) whose coordinates are also parameters to estimate, yielding the following optimization problem:
\begin{equation}
\begin{aligned}
    \min_{\A, \DAll, \BAll, \blueprint, \PAll} & \quad \left[\sum_{k=1}^K \fnorm{\TFS{X}{k} - \M{A} \Dk \Bk\Tra}^2 + \mu_t \fnorm{\Bk - \Pk \blueprint}^2 \right]\\
    \st & \quad \Pk\Tra \Pk = \M{I} \qquad \forall k \leq K, \\
         & \quad \A \geq 0, \\
         & \quad \Bk, \Dk \geq 0 \qquad \forall k \leq K,
\end{aligned}
\end{equation}
which can be solved with HALS \cite{GiGl12}, also known as column-wise updates \cite{BrSi98,Br98}. To ensure that the components satisfy the PARAFAC2 constraint, the regularization parameter, \(\mu_t\), is adaptively selected and increased after every iteration. The word ``flexible’’ in the name of this approach stems from the fact that the \(\Bk\) matrices are not directly parametrized by \(\Pk\blueprint\), but instead encouraged to be close to this product through a ``flexible coupling''. While the trick of gradually increasing the regularization strength to obtain models that follow the PARAFAC2 constraint could, in theory, be used with any constrained least squares solver, the algorithm in \cite{CoBr18} uses HALS and supports only non-negativity constraints. Furthermore, to the best of our knowledge, this formulation has only ever been used with non-negativity constraints.

\section{PARAFAC2 AO-ADMM}\label{sec:parafac2.aoadmm}
Consider the regularized PARAFAC2 problem:
\begin{equation}
\begin{aligned}
    \min_{\A, \ForAllK{\Bk, \Dk}} ~ & \left[f\left(\A, \BAll, \ForAllK{\Dk} \right)  + \reg{\A}{\A} + \sum_{k=1}^K\left\{ \reg{\Bk}{\Bk} + \reg{\Dk}{\Dk}\right\}\right] \label{eq:reg.pf2}\\
    \st ~ & \BAll \in \PSet,
\end{aligned}
\end{equation}
where \(f\) is the sum of squared errors data fidelity function, given by
\begin{equation}
    f\left(\A, \BAll, \ForAllK{\Dk} \right) = \sum_{k=1}^K \norm{\A \Dk \Bk \Tra - \Xk}_F^2,
\end{equation}
and \(\regF{\A}, \regF{\Bk}\), \(\regF{\Dk}\) are proximable regularization penalties.

As mentioned above, we cannot easily solve \cref{eq:reg.pf2} using the formulation in \cref{eq:pf2.kiers}, since it would require solving a regularized least squares problem with orthogonality constraints. Thus, we instead propose to solve \cref{eq:reg.pf2} with a scheme based on ADMM, solving each subproblem approximately. Specifically, we use an AO-ADMM algorithm.

AO-ADMM has been used for fitting constrained CP models \cite{HuSiLi16} and coupled matrix and tensor factorizations \cite{ScCoAc20,ScCoAc20b}. It has also been used for fitting PARAFAC2 models, by using AO-ADMM to update \(\A, \blueprint\) and \(\C\) (corresponding to using AO-ADMM instead of ALS to fit the CP model in Line \ref{alg.line:pf2cpstep} of \cref{alg:pf2.als}) \cite{AfPePaSeHoSu18}. However, with such a scheme, the \(\Bk\) matrices can only be constrained linearly.

We instead propose to use AO-ADMM to solve \cref{eq:reg.pf2} without reparametrization, thus allowing for proximal regularization of all factor matrices including \(\BAll\). For this, we introduce an algorithm for projecting onto \(\PSet\), which makes it feasible to use ADMM for solving regularized least squares problems with the PARAFAC2 constraint, and obtain the PARAFAC2 AO-ADMM algorithm described in \cref{alg:parafac2.aoadmm}.

In the following sections, we describe how each step of this algorithm is conducted in detail. First, we summarize ADMM in \cref{sec:admm} and AO-ADMM in \cref{sec:aoadmm}. 
Then, \cref{sec:B.updates} and \cref{sec:AD.updates}, describes ADMM update steps for lines \ref{alg.line:pf2.aoadmm.B} and \ref{alg.line:pf2.aoadmm.A}-\ref{alg.line:pf2.aoadmm.D} of \cref{alg:parafac2.aoadmm}, respectively.
\Cref{sec:stopping.feasibility} outlines stopping conditions and a heuristic to compute appropriate feasibility penalty parameters. Finally, \cref{sec:complexity} discusses the computational complexity for each step of the algorithm and \cref{sec:prox} lists some proximable penalties (with more details in Section SM3) and notes some important considerations when using penalty-based regularization. 

\begin{algorithm2e}
\SetAlgoLined
\DontPrintSemicolon
\KwResult{\(\A, \ForAllK{\Bk, \Dk}\)}
Initialize \(\A, \AAux, \ADual, \Bk, \BkAux, \BkDual, \blueprint, \Pk, \blueprintkDual, \Dk, \DkAux,\) and \(\DkDual\) \;
 \While{stopping conditions are not met and max no. iterations not exceeded}{
    Update \( \ForAllK{\Bk, \BkAux, \Pk, \BkDual, \blueprintkDual}\) and \(\blueprint \) using \cref{alg:admm.b} \label{alg.line:pf2.aoadmm.B}\;
    Update \( \A, \AAux \) and \(\ADual\) using \cref{alg:admm.a} \label{alg.line:pf2.aoadmm.A}\;
    Update \(\ForAllK{ \Dk, \DkAux, \DkDual}\) using \cref{alg:admm.cmf.d} \label{alg.line:pf2.aoadmm.D}\;
 }
 \caption{AO-ADMM for PARAFAC2}\label{alg:parafac2.aoadmm}
\end{algorithm2e}

\subsection{Summary of ADMM}\label{sec:admm}
ADMM solves optimization problems in the form:
\begin{equation}
\begin{aligned}
    \min_{\V{x}, \V{y}} \quad & \left[ f(\V{x}) + g(\V{y}) \right] \\
    \st \quad & \M{M}\V{x} + \M{N}\V{y} = \V{c}, \label{eq:admm.form}
\end{aligned}
\end{equation}
where \(\M{M}\) and \(\M{N}\) are known matrices and \(\V{c}\) is a known vector. For ADMM, we first formulate the augmented Lagrange dual problem of \cref{eq:admm.form}:
\begin{equation}
    \max_{\V{\nu}} \left[ \quad \min_{\V{x}, \V{y}} \quad \left[ f(\V{x}) + g(\V{y}) + \V{\nu}\Tra \left(\M{M}\V{x} + \M{N}\V{y} - \V{c}\right) + \frac{\rho}{2}\norm{\M{M}\V{x} + \M{N}\V{y} - \V{c}}_2^2 \right] \right], \label{eq:augmented.lagrangian}
\end{equation}
where \(\rho\) is a penalty parameter that defines the penalty for infeasible solutions. Then, we solve the saddle-point problem \cref{eq:augmented.lagrangian} with AO on \(\V{x}\) and \(\V{y}\) and gradient ascent on \(\V{\nu}\). Defining the scaled dual-variable, \(\V{\mu} = \V{\nu} / \rho\), we obtain \cref{alg:admm}.

One benefit is that whenever \(\M{M}\) or \(\M{N}\) are identity matrices, then the corresponding update step reduces to evaluating the scaled proximal operator (shown for \(g\)):
\begin{equation}
    \prox{\frac{g}{\rho}}{-(\M{M}\V{x} - \V{c} + \V{\mu})} =  \argmin_{\V{y}} \left[ g(\V{y}) + \frac{\rho}{2}\norm{\V{y} + \M{M}\V{x} - \V{c} + \V{\mu}}_2^2 \right], \label{eq:prox.def}
\end{equation}
which can be evaluated efficiently for a large family of functions. The update steps for \(\regF{\A}\), \(\regF{\Bk}\), and \(\regF{\Dk}\) in \cref{eq:reg.pf2} can always be reduced to this form, enabling any proximable regularization on all modes. See \cref{sec:prox} for details on proximal operators and examples of proximable penalties. For a thorough review of ADMM, we refer to \cite{BoPaChPeEc11}.

\begin{algorithm2e}
\DontPrintSemicolon
\SetAlgoLined
\KwResult{\(\V{x}, \V{y}, \V{\mu}\)}
 Initialize \(\V{x}\), \(\V{y}\) and \(\V{\mu}\)\;
 \While{stopping conditions are not met and max no. iterations not exceeded}{
  \(\V{x} \gets \argmin_{\V{x}} \left[ f(\V{x}) + \frac{\rho}{2}\norm{\M{M}\V{x} + \M{N}\V{y} - \V{c} + \V{\mu}}_2^2 \right]\) \;
  \(\V{y} \gets \argmin_{\V{y}} \left[ g(\V{y}) + \frac{\rho}{2}\norm{\M{M}\V{x} + \M{N}\V{y} - \V{c} + \V{\mu}}_2^2 \right]\) \;
  \(\V{\mu} \gets \M{M}\V{x} + \M{N}\V{y} - \V{c} + \V{\mu}\) \;
 }
 \caption{ADMM \cite{BoPaChPeEc11}}
 \label{alg:admm}
\end{algorithm2e}
\subsection{Overview of AO-ADMM}\label{sec:aoadmm}

Consider now a problem in the form
\begin{equation}
	\min_{\V{x}, \V{y}} \left[ f(\V{x}, \V{y}) + g_x(\V{x}) + g_y(\V{y}) \right],
\end{equation}
where the proximal operator for \(f\) is costly to evaluate, but the proximal operators of \(f_\V{x}(\V{y}) = f(\V{x}, \V{y})\), \(f_\V{y}(\V{x}) = f(\V{x}, \V{y})\), \(g_\V{x}\) and \(g_\V{y}\) are easy to evaluate. In that case, we can use an AO-scheme, where we fix \(\V{y}\) and update \(\V{x}\) with several iterations of ADMM, and vice versa. Such a scheme is called an AO-ADMM scheme \cite{HuSiLi16}.

There are, to the best of our knowledge, no convergence guarantees for AO-ADMM. For AO, it is known that if \(f_\V{y}(\V{x}) + g_x(\V{x})\) and \(f_\V{x}(\V{y}) + g_y(\V{y})\) always have unique minima and these minima are attained for each iteration, then all limit points of the algorithm are stationary points (see Proposition 2.7.1 in \cite{Be99} or the discussion of convergence in \cite{HuSiLi16}). Thus, if we impose a strictly convex regularization penalty (e.g., a ridge penalty), the AO-ADMM algorithm would provably converge to a stationary point given infinitely many ADMM iterations for each block update. While we often have convex regularization for updating the \(\A\)- and \(\C\)-matrix, the PARAFAC2 constraint for the \(\Bk\)-matrices is not convex, which makes it hard to conclude about the convergence of the ADMM updates for \(\BAll\), and therefore also the AO-ADMM algorithm.
Still, in our experience, the AO-ADMM algorithm converges in most cases.
\subsection{An ADMM scheme for \(\ForAllK{\mathbf{B}_k}\) in \cref{alg:parafac2.aoadmm}} \label{sec:B.updates}
Our main contribution is introducing an ADMM scheme to solve the non-convex optimization problem
\begin{equation}
\begin{aligned}
    \min_{\BAll} \quad & \left[ \sum_{k=1}^K \loss{\Bk}{\Bk} + \reg{\Bk}{\Bk} \right]\\
    \st \quad & \BAll \in \PSet, 
\end{aligned}
\end{equation}
where \(\loss{\Bk}{\Bk} = \norm{\smash{\A \Dk \Bk \Tra - \Xk}}_F^2\). We introduce the following splitting scheme
\begin{equation}
\begin{aligned}
    \min_{\BAll} \quad & \left[ \sum_{k=1}^K \left\{\loss{\Bk}{\Bk} + \reg{\Bk}{\BkAux}\right\} + \ConstraintIndicator{\BAllConstraint} \right]\\
    \st \quad & \Bk = \BkAux \qquad \forall k \leq K,\\
        \quad & \Bk = \BkConstraint \qquad \forall k \leq K,
\end{aligned} \label{eq:B.split}
\end{equation}
where \(\ConstraintIndicator{\smash{\BAllConstraint}} = 0\) if \(\BAllConstraint \in \PSet\), and \(\infty\) otherwise. This splitting scheme is in the standard form for problems that are solvable with ADMM, which we can utilize to obtain \cref{alg:admm.b}. Next, we describe each line in this algorithm in detail.

\begin{algorithm2e}
\DontPrintSemicolon
\SetAlgoLined
\KwResult{\(\Bk, \BkAux, \BkConstraint, \BkDual, \blueprintkDual\)}
 \While{stopping conditions are not met and max no. iterations not exceeded}{
  \For{\(k \gets 1\) \KwTo \(K\)}{
    \(
    \begin{aligned}
        \Bk 
        \xleftarrow{\cref{eq:B.loss.update}} 
        \argmin_{\Bk} \left[ \begin{aligned} \loss{\Bk}{\Bk} 
        &+ \frac{\rho_{\Bk}}{2}\norm{\Bk - \BkAux + \BkDual}^2 \\
        &+ \frac{\rho_{\Bk}}{2}\norm{\Bk - \BkConstraint+ \blueprintkDual}^2 \end{aligned}\right]
    \end{aligned}\) \label{alg.line:admm.b.update}\;
    \(\BkAux \gets \prox{ \frac{\regF{\M{B}}}{\rho_{\Bk}} }{\Bk + \BkDual}\) \label{alg.line:admm.b.aux.update} \;
  }
  \(\BAllConstraint \xleftarrow{\text{Alg.}~\ref{alg:constraint.prox}} \prox{\ConstraintIndicatorF{}}{\ForAllK{\Bk + \blueprintkDual}} \)  \label{alg.line:admm.b.pf2.update}\;
  \For{\(k \gets 1\) \KwTo \(K\)}{
    \(\BkDual \gets \Bk - \BkAux + \BkDual\) \;
    \(\blueprintkDual \gets \Bk - \BkConstraint + \blueprintkDual\) \;
  }
 }
 \caption{ADMM updates for the \(\Bk\)-matrices}
 \label{alg:admm.b}
\end{algorithm2e}

Two main lines in \cref{alg:admm.b} need to be derived: the update rules for \(\BAll\) in Line~\ref{alg.line:admm.b.update} and the update rules for \(\BAllConstraint\) in Line~\ref{alg.line:admm.b.pf2.update}. The proximal operator for \(\regF{\Bk}\) in Line~\ref{alg.line:admm.b.aux.update} is specified by the regularization function. The update rule for \(\Bk\) requires us to solve a least squares problem, which has a closed form solution:
\begin{equation}
    \Bk \gets \left(\Xk\Tra \A \Dk + \frac{\rho_{\Bk}}{2}\left(\BkAux - \BkDual + \BkConstraint - \blueprintkDual\right)\right)\left( \Dk\A\Tra  \A \Dk + \rho_{\Bk} \M{I} \right)^{-1}. \label{eq:B.loss.update}
\end{equation}
while the update rules for \(\BAllConstraint\) require an efficient way to estimate \(\proxF{\ConstraintIndicatorF}\).

Evaluating \(\proxF{\ConstraintIndicatorF}\) is equivalent to evaluating a projection onto \(\PSet\), which we estimate using the parametrization of \(\PSet\) discussed in \cref{sec:parafac2.observations}, yielding:
\begin{equation}
\begin{aligned}
    \min_{\blueprint, \PAll} \quad & \left[ \sum_{k=1}^K \frac{\rho_{\Bk}}{2}\norm{\Bk - \Pk \blueprint + \blueprintkDual}^2  \right]\\
    \st \quad & \Pk\Tra \Pk = \M{I} \qquad \forall k \leq K.
\end{aligned}\label{eq:pset.proj}
\end{equation}
This equation can be approximated efficiently with an AO scheme, where the orthogonal Procrustes problem for each \(\Pk\) is solved independently. With this, we obtain \cref{alg:constraint.prox}. In our experience, one iteration in \cref{alg:constraint.prox} for each ADMM iteration in \cref{alg:admm.b} is sufficient.

\begin{algorithm2e}
\DontPrintSemicolon
\SetAlgoLined
\KwResult{\(\ForAllK{\Pk}, \blueprint\)}
 \While{stopping conditions are not met and max no. iterations not exceeded}{
  \For{\(k \gets 1\) \KwTo \(K\)}{
    Compute ``economy style'' SVD: \( \left(\Bk + \blueprintkDual\right)\blueprint \Tra = \M{U}^{(k)} \M{\Sigma}^{(k)} {\M{V}^{(k)}}\Tra\) \;
    \( \Pk \gets \M{U}^{(k)}{\M{V}^{(k)}}\Tra\) \;
   }
    \(\blueprint \gets \frac{1}{\sum_{k=1}^K \rho_{\Bk}} \sum_{k=1}^K \rho_{\Bk} \Pk\Tra \left( \Bk + \blueprintkDual \right)\) \;
    
 }

 \caption{Approximate projection onto \(\PSet\)}\label{alg:constraint.prox}
\end{algorithm2e}

\subsection{ADMM updates for \(\mathbf{A}\) and \(\ForAllK{\mathbf{D}_k}\) in \cref{alg:parafac2.aoadmm}\label{sec:AD.updates}}
ADMM updates for \(\mathbf{A}\) and \(\ForAllK{\mathbf{D}_k}\) can be obtained in multiple ways. Here, we introduce ADMM schemes based on the coupled matrix decomposition interpretation of PARAFAC2 (CMF-based updates). However, by assuming primal feasibility in \cref{eq:B.split} (i.e., \(\Bk = \Pk \blueprint\)), we can use CP-based updates for \(\A\) and \(\DAll\) \cite{AfPePaSeHoSu18}. For discussions and comparisons with the CP-based scheme, see Section SM5.

For the CMF-based update steps, we want to solve the optimization problems
\begin{equation}
\min_{\A} \left[ \loss{\A}{\A} + \reg{\A}{\A} \right] \qquad \text{and} \qquad \min_{\Dk} \left[ \loss{\Dk}{\Dk} + \reg{\Dk}{\Dk} \right],
\end{equation}
for updating \(\A\) and \(\DAll\), respectively. \(\lossF{\A}\) and \(\lossF{\Dk}\) are the sum of squared errors data fidelity function for \(\A\) and \(\DAll\), respectively, and \(\regF{\A}\) and \(\regF{\Dk}\) are regularization penalties. 
We apply ADMM directly to these optimization problems.

The proximal operators for the data fidelity functions consist of solving least squares problems. The proximal operator with respect to \(\A\), is given by
\begin{equation}
    \prox{\frac{\lossF{\A}}{\rho_\A}}{\M{M}} = \left( \sum_{k=1}^K \Xk \Mn{\Gamma}{k}  + \frac{\rho_\A}{2}\M{M} \right) \left( \sum_{k=1}^K \Mn{\Gamma}{k} \Tra \Mn{\Gamma}{k} + \frac{\rho_\A}{2} \M{I} \right)^{-1}, \label{eq:A.cmf.loss.update}
\end{equation}
where \(\Mn{\Gamma}{k} = \Bk \Dk\). To compute \(\prox{\frac{\lossF{\Dk}}{\rho_{\Dk}}}{\V{v}}\), we consider the vectorized problem:
\begin{equation}
\prox{\frac{\lossF{\Dk}}{\rho_{\Dk}}}{\V{v}} = \argmin_{\MR{C}{k}} \left[ \fnorm{\left(\A \Khat \Bk\right)\MR{C}{k}\Tra - \vectorize{\Xk}}^2 + \frac{\rho_{\Dk}}{2}\fnorm{\MR{C}{k}\Tra - \V{v}}^2 \right],
\end{equation}
where \(\MR{C}{k}\Tra\) is a vector containing the diagonal entries of \(\Dk\). Then, we use \((\A \Khat \Bk)\Tra(\A \Khat \Bk) = (\A\Tra\A * \Bk\Tra\Bk)\) to obtain
\begin{equation}
\prox{\frac{\lossF{\Dk}}{\rho_{\Dk}}}{\V{v}} = \left( \A\Tra \A * \Bk\Tra\Bk + \frac{\rho_{\Dk}}{2} \M{I} \right)^{-1}  \left( \DiagEntries{\A\Tra \Xk \Bk} + \frac{\rho_{\Dk}}{2}\V{v} \right), \label{eq:C.cmf.loss.update}
\end{equation}
where \(\DiagEntries{\A\Tra \Xk \Bk}\) is the vector containing the diagonal entries of \(\A\Tra \Xk \Bk\).

\begin{algorithm2e}
\SetAlgoLined
\DontPrintSemicolon
\KwResult{\(\A, \AAux, \ADual\)}
 \While{stopping conditions are not met and max no. iterations not exceeded}{
    \( \A \xleftarrow{\cref{eq:A.cmf.loss.update}} \prox{\frac{\lossF{\A}}{\rho_\A}}{\AAux - \ADual}\) \;
    \( \AAux \gets \prox{\frac{ \regF{\A} }{ \rho_\A }}{\A + \ADual} \) \;
    \( \ADual \gets \ADual + \A - \AAux \) \;
 }
 \caption{CMF-based ADMM updates for the \(\A\)-matrix}\label{alg:admm.a}
\end{algorithm2e}

\begin{algorithm2e}
\SetAlgoLined
\DontPrintSemicolon
\KwResult{\(\Dk, \DkAux, \DkDual\)}
 \While{stopping conditions are not met and max no. iterations not exceeded}{
   \For{\(k \gets 1\) \KwTo \(K\)}{
        \( \Dk \xleftarrow{\cref{eq:C.cmf.loss.update}} \prox{\frac{\lossF{\Dk}}{\rho_{\Dk}}}{\DkAux - \DkDual}\) \;
        \( \DkAux \gets \prox{\frac{ \regF{\Dk} }{ \rho_{\Dk} }}{\Dk + \DkDual} \) \;
        \( \DkDual \gets \DkDual + \Dk - \DkAux \) \;
    }
 }
 \caption{CMF-based ADMM updates for the \(\Dk\)-matrices (\(\C\)-matrix)}\label{alg:admm.cmf.d}
\end{algorithm2e}

\subsection{Stopping conditions and feasibility penalties}\label{sec:stopping.feasibility}
For the inner ADMM loops (\cref{alg:admm.a,alg:admm.b,alg:admm.cmf.d}), we follow \cite{BoPaChPeEc11} and use stopping conditions based on the primal and dual residuals:
\begin{equation}
    \frac{\norm{\V{x}^{(t, q)} - \V{z}^{(t, q)}}}{\norm{\V{x}^{(t, q)}}} \leq \epsilon, \qquad \text{and} \qquad \frac{\norm{\V{z}^{(t, q)} - \V{z}^{(t, q-1)}}}{\norm{\V{z}^{(t, q)}}} \leq \epsilon. \label{eq:admm.residuals}
\end{equation}
In our case, \(\V{z}\) is replaced with an auxiliary factor matrix (e.g., \(\BkAux\) or \(\BkConstraint\)) and \(\V{x}\) is replaced with the corresponding factor matrix (e.g., \(\Bk\)). The \((t, q)\) superscript represents the current outer and inner iteration number, respectively. For stopping the inner loops, we require that either all stopping conditions are fulfilled, or that a predefined number of iterations have been performed. In our experience, a low number (e.g., five) of inner iterations is sufficient.

For the outer loop, in \cref{alg:parafac2.aoadmm} we have two types of stopping conditions: loss decrease conditions and feasibility conditions. We stop once both the regularized sum of squared error and the relative feasibility gaps are below a given threshold, or their relative decrease is below a given threshold. That is, we ensure that 
\begin{equation*}
    f^{(t)} + g^{(t)} < \epsilon^{\text{abs}} \qquad \text{or} \qquad \left|f^{(t-1)} + g^{(t-1)} - f^{(t)} - g^{(t)}\right| < \epsilon^{\text{rel}}\left(f^{(t-1)} + g^{(t-1)}\right),
\end{equation*}
where \(f^{(t)}\) is the sum of squared errors after \(t\) outer iterations and \(g^{(t)}\) is the sum of all regularization penalties after \(t\) outer iterations, are satisfied. We also ensure that
\begin{equation*}
    \V{r}^{(t)}_\V{z} = \norm{\V{x}^{(t)} - \V{z}^{(t)}} < \epsilon^{\text{abs}} \qquad \text{or} \qquad |\V{r}^{(t)}_\V{z} - \V{r}^{(t-1)}_\V{z}| < \epsilon^{\text{rel}} \V{r}^{(t-1)}_\V{z},
\end{equation*}
is satisfied for all auxiliary variables \(\V{z}\). After exiting the AO-ADMM algorithm, it is important to verify that all primal feasibility gaps are sufficiently small. Otherwise, we may end up with components that violate the constraints.

In order to select the penalty parameters, we used the heuristic in \cite{HuSiLi16,ScCoAc20}, setting
\begin{equation}
\begin{aligned}
    \rho_\A &= \frac{1}{R}\Trace{\sum_{k=1}^K \Dk \Bk\Tra \Bk \Dk}\\
    \rho_{\Bk} &= \frac{1}{R}\Trace{\Dk \A\Tra \A \Dk}, \\
    \rho_{\Dk} &= \frac{1}{R}\Trace{\A\Tra \A * \Bk\Tra \Bk},
\end{aligned}
\end{equation}
For convex regularization penalties, the ADMM-subproblems are guaranteed to converge with any \(\rho\). However, for non-convex regularization penalties (e.g., the \(\ConstraintIndicatorF\) in the \(\Bk\) updates) the value of \(\rho\) can affect ADMM's convergence properties \cite{BoPaChPeEc11}. 

\subsection{Computational complexity of the AO-ADMM update steps} \label{sec:complexity}
\begin{table}
\centering
\caption{Computational complexities for the different AO-ADMM update steps. \(I\), \(J\) and \(K\) denote the tensor size, \(R\) denotes the number of components and \(Q\) the number of iterations.} \label{tab:complexities}
\begin{tabular}{@{}ll@{}}
\toprule
          & Computational complexity                          \\
\midrule
\(\A\)    & \(O(IJKR + JKR^2 + R^3 + IR^2Q)\)          \\
\(\DAll\) & \(O(IJKR + IR^2 + JKR^2 + KR^3 + KR^2Q)\)  \\
\(\BAll\) & \(O(IJKR + IR^2 + KR^3 + JKR^2Q)\)         \\
\bottomrule
\end{tabular}
\end{table}

The computational complexities for the AO-ADMM update steps are given in \cref{tab:complexities}\footnote{for the computational complexities of the CP-based updates, see supplementary Table SM1}. To minimize the computational complexity, we compute parts of the right-hand side as well as the Cholesky factorization of the left hand side of all normal equations (e.g.  \(\Xk\Tra \A \Dk\) and the Cholesky factorization of \((\Dk\A\Tra  \A \Dk + \rho_{\Bk} \M{I})\) for \cref{eq:B.loss.update}) only once per outer iteration and re-use it for the inner iterations.  

\subsection{Constraints} \label{sec:prox}
In this section, we review some background on proximal operators and list some useful proximable regularization penalties. 
If \(g\) is a proper lower semi-continuous convex function, the proximal operator, given by \cref{eq:prox.def}, has a unique solution. For non-convex \(g\), \cref{eq:prox.def} may not have a unique solution, and in those cases, we select one of the possible solutions.
A large variety of regularization penalties admit closed form solutions or efficient algorithms for evaluating their proximal operator \cite{PaBo14,Be17,Co13,St08}. In this work, we evaluate the efficency of our AO-ADMM scheme for four such penalty functions, given in \cref{tab:proxes}. In particular, we impose non-negativity, graph Laplacian regularization, total variation (TV) regularization and unimodality. For more details on these penalties, see Section SM3 in the supplement.

\begin{table}[]
    \centering
    \caption{Some proximable regularization penalties}
    \begin{tabular}{@{}lrr@{}}
    \toprule
        Structure & Penalty & Proximal operator \\
    \midrule
        Non-negativity & \(\IndicatorF{\Real_+}\) & \(\prox{\IndicatorF{\Real_+}}{x} = \max (0, x)\) \\
        Graph Laplacian regularization & \(\V{x}\Tra\M{L}\V{x}\) & \(\prox{\V{x}\Tra\M{L}\V{x}}{\V{x}} = \left(\M{L} + 0.5 \M{I}\right)^{-1}\mathbf{x}\) \\
        TV regularization & \(\sum_i \left| x_i - x_{i-1} \right|\) & \cite{Co13,Co17} \\
        Unimodality & \(\IndicatorF{\mathcal{U}}\)  & \cite{St08} \\
        Unimodality and non-negativity & \(\IndicatorF{\mathcal{U} \cap \Real_+} \) & \cite{BrSi98} \\
        PARAFAC2 constraint & \(\IndicatorF{\PSet} \) & \cref{alg:constraint.prox} \\
    \bottomrule
    \end{tabular}
    \label{tab:proxes}
\end{table}

\subsubsection{The scale-indeterminacy of penalty-based regularization}\label{sec:scale.reg}
Many penalty functions, such as graph Laplacian regularization and TV regularization, scale with the norm of the components. However, the PARAFAC2 loss function is invariant to scaling\footnote{The observations in this section also hold for the CP decomposition.}. For example, \(\A\) can be multiplied with a constant, \(\epsilon\), so long as we multiply either all \(\Bk\)-matrices, or all \(\Dk\)-matrices by \(1/\epsilon\). Thus, if we let \(\epsilon \to 0\) we can obtain an arbitrarily small regularization penalty without affecting the components recovered by the algorithm (ignoring numerical difficulties arising as \(\epsilon \to 0\)).

To circumvent the scaling indeterminacy, we must regularize the norm of all factor matrices whenever the regularization for one factor matrix is norm-dependent. A straightforward way of regularizing the norm of the other factor matrices is with ridge regularization, which we can incorporate into the proximal operator for the data-fidelity term. We will, for brevity's sake, only show how the proximal operator for the \(\A\)-matrix is altered by this change (see supplementary for \(\Bk\) and \(\Dk\) alterations): 
\begin{equation}
   \prox{\frac{\lossF{\A} + \gamma\fnorm{\cdot}^2}{\rho_\A}}{\M{M}} = \left( \sum_{k=1}^K \Xk \Mn{\Gamma}{k}  + \frac{\rho_\A}{2}\M{M} \right) \left( \sum_{k=1}^K \Mn{\Gamma}{k} \Tra \Mn{\Gamma}{k} + \frac{2\gamma + \rho_\A}{2} \M{I} \right)^{-1},
\end{equation}
where \(\Mn{\Gamma}{k} = \Bk \Dk\) and \(\gamma\) is the ridge regularization penalty.
Including ridge regularization this way does not necessarily increase the number of tunable parameters, which is apparent from the following theorem (proof in Section SM1):
 
\begin{theorem}\label{th:reg.scaling}
Let \(f: \Real^{n} \times \Real^{m} \to \Real\) be a function satisfying \(f(a\V{u}, a^{-2}\V{v}) = f(\V{u}, \V{v})\), and let \(r_u : \Real^{n} \to \Real\) and \(r_v : \Real^{m} \to \Real\) be two (absolutely) homogeneous functions of degree \(d_u\) and \(d_v\) respectively. That is \(r_u(a \V{u}) = |a|^{d_u}r_u(\V{u})\) and  \(r_v(a \V{v}) = |a|^{d_v}r_v(\V{v})\). Then the optimization problems
\begin{equation}
    \min_{\V{u}, \V{v}} \left[f(\V{u}, \V{v}) + a r_u(\V{u}) + r_v(\V{v}) \right]
\end{equation}
and 
\begin{equation}
    \min_{\V{u}, \V{v}} \left[f(\V{u}, \V{v}) + r_u(\V{u}) + a^{2 \frac{d_v}{d_u}} r_v(\V{v}) \right]
\end{equation}
are equivalent\footnote{Two optimization problems are equivalent if the solution to one can easily be transformed to the solution of the other \cite{BoVa04}.} for any positive a.
\end{theorem}
Consequently, if we impose ridge regularization on two modes (e.g., \(\A\) and \(\DAll\)) with the same regularization parameter, and a homogeneous regularization penalty on the remaining mode (\(\Bk\)), then scaling the ridge penalty is equivalent to scaling the \(\Bk\)-penalty (set \(\V{u} = \{\A, \DAll\}\) and \(\V{v} = \BAll\) and apply \cref{th:reg.scaling}).

\section{Experiments}\label{sec:experiments}
Here, we evaluate our proposed AO-ADMM scheme for constrained PARAFAC2 on a variety of experiments. We assess the performance in terms of accuracy (i.e., how well the underlying factors are captured), and computational efficiency on simulated datasets, and show the benefits of imposing constraints on the evolving mode of a PARAFAC2 model with real-world applications from chemometrics and neuroscience. Specifically, we use seven simulation setups. In Setup 1, 2, and 3, we demonstrate that the AO-ADMM scheme is both as accurate as and faster than the flexible coupling with HALS scheme for setups with non-negativity in all modes. Setup 1 also demonstrates the performance of the methods in the case of over- and under-estimation of the number of components. In Setup 3, we demonstrate that imposing unimodality constraints on \(\BAll\) can improve accuracy and that the AO-ADMM scheme can be effective even when \(\BAll \not\in \PSet\). 
Next, in Setup 4 and 5, we respectively use a graph Laplacian- and TV-penalty to improve factor recovery. In Setup 6, we show that the cwSNR is an accurate predictor of factor recovery. Finally, in Setup SM1 in the supplementary, we compare the CP-based and CMF-based update steps on non-negative components, and show that they are similar in speed and accuracy. For Setup 1-6, we compare with a baseline based on PARAFAC2 ALS, which allows for constraints on \(\A\) and \(\DAll\), but not on \(\BAll\). 

\subsection{Experimental setup}
For both the PARAFAC2 AO-ADMM algorithm and the baselines, we used our Python implementations linked in the GitHub repository for the paper\footnote{\url{https://github.com/MarieRoald/PARAFAC2-AOADMM-SIMODS/}} (There is also a MATLAB implementation linked in the same repository). The flexible coupling PARAFAC2 with HALS algorithm was implemented in Python closely following the MATLAB implementation by Cohen and Bro \cite{CoBr18} (with some caching and reordering of computations for increased efficiency). To compute the proximal operator of the TV seminorm, we used the publicly available C implementation  \cite{Co17} of the improved direct TV denoising algorithm presented in \cite{Co13}. Finally, the projection onto the set of non-negative unimodal vectors was implemented by thresholding the isotonic regression-vectors obtained with the algorithm from \cite{St08}.

We set all stopping tolerances for the inner iterations to \(10^{-5}\) with a maximum of 5 iterations (except for the unimodality setup, see \cref{sec:simulation.unimodality}). For the outer loop, we set all relative tolerances to \(10^{-8}\) and all absolute tolerances to \(10^{-7}\).

For all experiments, we initialized the factor matrices, auxiliary matrices and scaled dual variables by drawing their elements from a uniform distribution between 0 and 1 (except the \(\Pk\)-matrices which were initialized as the first \(R\) columns of an identity matrix, thus ensuring that they are orthogonal). The same initial factor matrices were used for both the AO-ADMM experiments and the ALS experiments. We initialized the \(\Pk\)- and \(\blueprint\)-matrices for ALS equally to the corresponding auxiliary variables for AO-ADMM. For all experiments, 20 initializations were used (except for Setup 6, where we used 50 initializations, and Setup 1 and SM1, where we used 10 initializations). We selected, for each model, the initialization that provided the lowest regularized sum of squared error among those that satisfied the stopping conditions (to ensure a low feasibility gap for the AO-ADMM components). If none satisfied the stopping conditions (i.e. all initializations reached the maximum number of iterations), we selected the initialization with the lowest regularized sum of squared error. All experiments imposed non-negativity on \(\DAll\) to resolve the special sign indeterminacy of the PARAFAC2 model \cite{He13}. 
\subsection{Simulations} 
For all simulation experiments, we first constructed simulated factor matrices. Then, after constructing a simulated tensor, \(\X\), from those factor matrices (following \cref{eq:pf2}), we added artificial noise as follows. First, we created a (possibly ragged) tensor \(\T{E}\) with elements drawn from a normal distribution. Then, \(\T{E}\) was scaled to have magnitude \(\eta \norm{\X}_F\) and added to \(\T{X}\) following
\begin{equation}
    \X_\text{noisy} = \X + \eta \norm{\X}_F \frac{\T{E}}{\norm{\T{E}}_F}. \label{eq:noise.level}
\end{equation}

Since this work introduces a scheme for regularizing the \(\Bk\)-matrices, we generated \(\Bk\)-matrices with structures tailored to various constraints.
The elements of \(\A\) were drawn from a truncated normal distribution (i.e. \(a_{ir} = \max (0, \breve{a}_{ir})\), where \(\breve{a}_{ir} \sim \mathcal{N}(0, 1)\)), and the elements of all \(\Dk\)-matrices were uniformly distributed ($\mathcal{U}(0.1, 1.1)$, avoiding near-zero elements, which can impede the recovery of \(\Bk\)).

To evaluate the different methods' performance on simulated data, we used the relative sum of squared errors (Rel. SSE) given by
\begin{equation}
    \text{Rel. SSE} = \frac{\sum_ {k=1}^K \fnorm{\Xk - \A\Dk\Bk\Tra}^2}{\sum_ {k=1}^K \fnorm{\Xk}^2} ,
\end{equation}
and the factor match score (FMS), which is defined as
\begin{equation}\label{eq:fms}
\FMS = \frac{1}{R} \sum_{r=1}^R 
\left| \MC{a}{r} \Tra \hat{\V{a}}_{r}
\tilde{\V{b}}_{r} \Tra \hat{\tilde{\V{b}}}_{r}
\MC{c}{r} \Tra \hat{\V{c}}_{r} \right|,
\end{equation}
with all component vectors normalized and the superscript \(\hat{}\) represents the estimated vectors. \(\smash{\tilde{\V{b}}_{r}}\) and \(\smash{\hat{\tilde{\V{b}}}_{r}}\) are vectors containing the concatenation of the \(r\)-th column of all \(\Bk\)-matrices and \(\smash{\hat{\M{B}}_k}\)-matrices, respectively. Likewise, \(\V{c}_r\) and \(\hat{\V{c}}_{r},\) contain the \(r\)-th diagonal entry of all \(\Dk\)-matrices and \(\smash{\hat{\M{D}}_k}\)-matrices. The FMS ranges from zero to one where one indicates fully recovered components.
Since the PARAFAC2 decomposition is only unique up to permutation and scaling, we find the optimal permutation of the components before computing the FMS.
For the more difficult setups (Setup 2-3), we also evaluate the recovery of each factor matrix independently by computing \(\FMSA, \FMSB\) and \(\FMSC\) only considering the relevant terms of \cref{eq:fms}, i.e.:
\begin{equation}
\FMSA = \frac{1}{R} \sum_{r=1}^R \left|\MC{a}{r} \Tra \hat{\V{a}}_{r} \right|.
\end{equation}

\subsubsection{Setup 1: Shifting non-negative components}\label{sec:flex.comparison}
\paragraph{Data generation}
To compare the AO-ADMM scheme with the flexible coupling with HALS scheme in terms of speed and accuracy, we used a simulation setup inspired by the simulations in \cite{CoBr18}. Specifically, we generated data by drawing the elements of \(\tilde{\B}\) from a truncated normal distribution. Then, to obtain \(\Bk\), we shifted elements of \(\tilde{\B}\) by \(k\) indices, cyclically, setting \([\Bk]_{j, r} = \tilde{\B}_{(j + k) \text{mod} J, r}\). 
This shifting yields non-negative \(\Bk\)-matrices that vary in a way that follows the PARAFAC2 constraint. Using this approach, we generated 50 different 3-component datasets, and 50 different 5-component datasets, each of size \(30 \times 40 \times 50\), before adding noise with \(\eta = 0.33\). 

\paragraph{Experiment settings}
We used AO-ADMM and flexible coupling with HALS\footnote{We also ran experiments where we replaced HALS with other non-negative least squares algorithms \cite{BeKe04,KiPa11} and found that the relative performance of the AO-ADMM compared to the flexible coupling schemes did not change (results not shown).} to decompose the data tensors with non-negativity imposed on all modes. We also compared with ALS with non-negativity on \(\A\) and \(\DAll\). For the flexible coupling with HALS scheme, we used the same initialization of \(\A, \BAll, \DAll\), \(\PAll\) and \(\blueprint\) as the AO-ADMM scheme. All models were run until the stopping conditions were satisfied or a maximum of 2000 iterations was reached. For the 3-component data, we also fitted models with two and four components to compare the methods’ robustness to over- and under-estimation of the number of components. For these experiments, we set the maximum number of iterations to 3000 and the absolute feasibility gap tolerance to \(10^{-5}\).

\begin{figure}
    \centering
    \begin{subfigure}{0.48\textwidth}
    \includegraphics{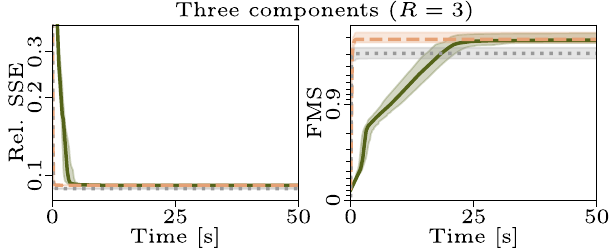}
    \end{subfigure}
    \begin{subfigure}{0.48\textwidth}
    \includegraphics{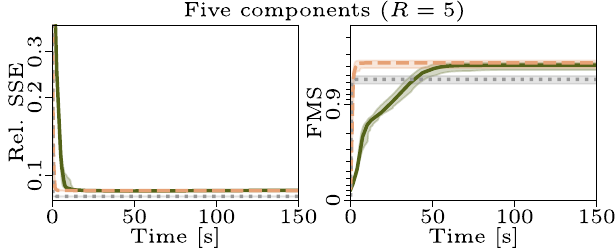}
    \end{subfigure}
    \begin{subfigure}{0.48\textwidth}
    \includegraphics{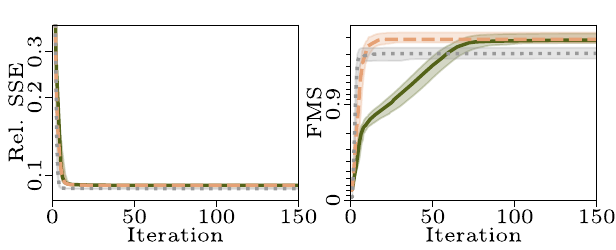}
    \end{subfigure}
    \begin{subfigure}{0.48\textwidth}
    \includegraphics{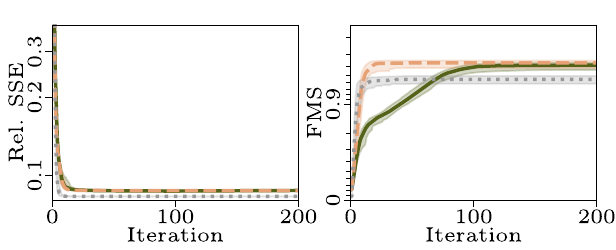}
    \end{subfigure}
    \begin{subfigure}{\textwidth}
    \includegraphics{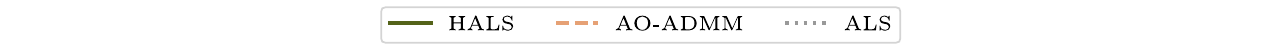}
    \end{subfigure}
    \caption{Setup 1: Diagnostic plots for the simulated data tensors. Median values are shown as solid lines, and the shaded area shows the interquartile range. The top row shows performance as a function of time, and the bottom row shows the performance as a function of iterations. The x-axes are cropped at 60 s ($R$=3) and 150 s ($R$=5) for the top row, and 150 iterations ($R$=3) and 200 iterations ($R$=5) for the bottom row. Subsequent iterations only decreased the feasibility gap (for AO-ADMM) and regularization penalty (for flexible coupling with HALS), not the FMS or relative SSE.}
    \label{fig:flex.comparison}
\end{figure}

\begin{figure}
    \centering
    \begin{subfigure}{0.48\textwidth}
    \includegraphics{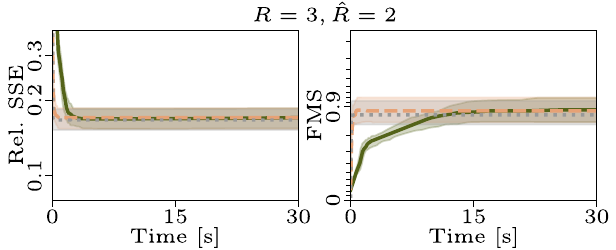}
    \end{subfigure}
    \begin{subfigure}{0.48\textwidth}
    \includegraphics{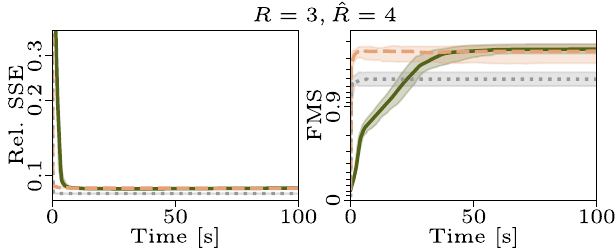}
    \end{subfigure}
    \begin{subfigure}{0.48\textwidth}
    \includegraphics{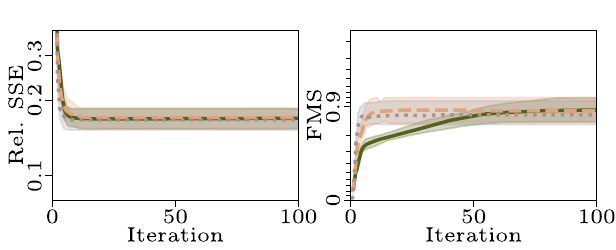}
    \end{subfigure}
    \begin{subfigure}{0.48\textwidth}
    \includegraphics{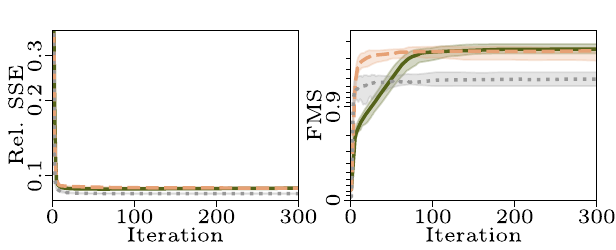}
    \end{subfigure}
    \begin{subfigure}{\textwidth}
    \includegraphics{figures/paper/M145003_setup1_legend.pdf}
    \end{subfigure}
    \caption{Setup 1: Diagnostic plots for the simulated data tensors with under- and over-estimation of the number of components. \(R\) represents the true number of components and \(\hat{R}\) represents the estimated number of components. To compute the FMS between two decompositions with different numbers of components, we use the subset of \(\min(R,\hat{R})\) components that provides the highest FMS.}
    \label{fig:flex.over.under.estimation.rank}
\end{figure}

\paragraph{Results}
\Cref{fig:flex.comparison} shows that the AO-ADMM scheme attains as good FMS as the flexible coupling with HALS scheme but faster. The AO-ADMM scheme was slightly slower than ALS in terms of speed but got a higher FMS. In \cref{fig:flex.over.under.estimation.rank}, we see that the relative performance of the methods are the same in overfactoring and underfactoring with ALS getting lower FMS, and HALS being slower than AO-ADMM.

\subsubsection{Setup 2: Non-negative components}
\paragraph{Data generation}
Setup 1 focuses on shifting \(\Bk\)-matrices as in \cite{CoBr18}. However, this implicitly assumes \(\Bk = \Pk \blueprint\), with non-negative \(\blueprint \in \Real^{J \times R}\) and non-negative \(\Pk \in \Ortho{J}{J}\). Thus, the \(\Pk\)-matrices are assumed to have disjoint support. Here, we evaluate the different PARAFAC2 fitting algorithms on non-negative datasets without these implicit assumptions. Specifically, we construct a non-negative cross-product matrix, \(\M{X}\Tra \M{X}\), where \(\M{X} \in \Real^{100 \times R}\) has elements from a truncated normal distribution and obtain the \(\Bk\) factor matrices by solving:
\begin{equation}
\begin{aligned}
\min_{\Bk} \quad & \left[ ||\Bk \Tra \Bk - \M{X}\Tra\M{X}||^2 \right] \\
\st \quad & [\Bk]_{jr} \geq 0,
\end{aligned}
\label{eq:general.bk}
\end{equation}
using projected gradient descent with various random initializations for each \(\Bk\) (more details in the supplementary Subsection SM6.1). Following this, we constructed 50 three-component ragged tensors, consisting of 15 frontal slices, of size \(50 \times J_k\), where \(J_k\) was a random integer between \(50\) and \(100\). We also evaluated non-negativity for all modes with increased collinearity in the \(\C\)-matrix, which we obtained by using \(\tilde{\M{D}}_k\) instead of \(\Dk\), with \([\tilde{\M{D}}_k]_{rr} = 0.5 [\Dk]_{11} + 0.5 [\Dk]_{rr}\). Finally, we added noise following \eqref{eq:noise.level} with 10 \(\eta\)-values logarithmically spaced between 0.5 and 2.5.

\paragraph{Experiment settings} We decomposed each data tensor using PARAFAC2 AO-ADMM and HALS with non-negativity on all modes and PARAFAC2 ALS (with non-negativity on \(\A\) and \(\DAll\)), both with a maximum of 6000 iterations.

\begin{figure}
    \centering
    \begin{subfigure}{\textwidth}
        \includegraphics{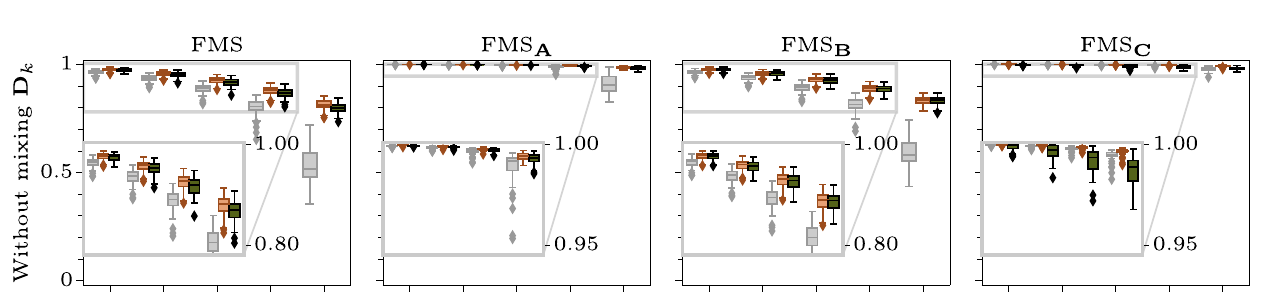}
    \end{subfigure}
    \begin{subfigure}{\textwidth}
        \includegraphics{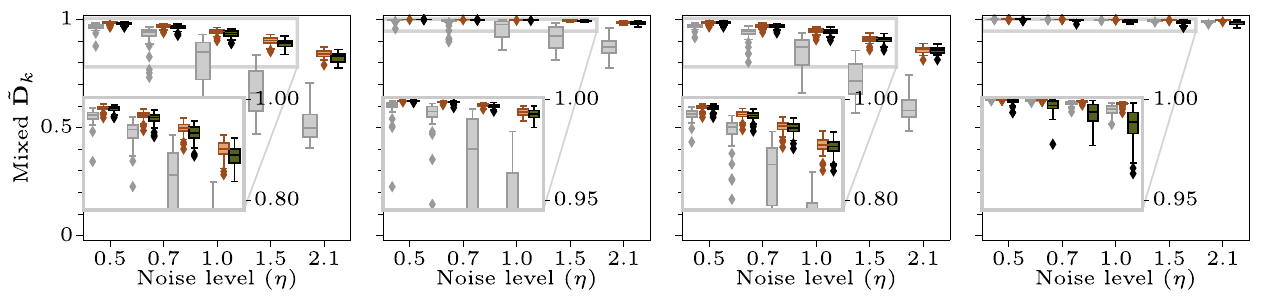}
    \end{subfigure}
    \begin{subfigure}{\textwidth}
        \includegraphics{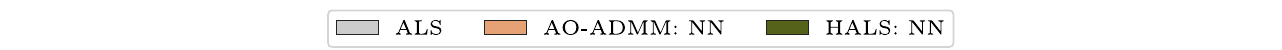}
    \end{subfigure}
    \caption{Setup 2: Boxplots showing the FMS for the different datasets. ALS refers to models fitted with ALS and non-negativity only on \(\A\) and \(\DAll\), AO-ADMM: NN refers to models fitted with AO-ADMM with non-negativity imposed on all modes and HALS: NN refers to models fitted with HALS with non-negativity imposed on all modes. For the sake of space, this figure only shows every other noise level, see supplementary Figure SM4 for the remaining noise levels.}
    \label{fig:sim.nn.boxplot}
\end{figure}

\paragraph{Results}
For some initializations, the ALS algorithm obtained degenerate solutions (solutions where two components are highly correlated in all modes but point in the opposite direction, and therefore cancel each other out) \cite{ZiKi02}. These initializations were excluded from the analysis. Furthermore, we also discarded the datasets where ALS gave degenerate solutions for all 20 initializations (these datasets were also removed for AO-ADMM)\footnote{See supplementary Table SM4 for an overview of the number of datasets excluded because ALS gave degenerate solutions.}. To identify degenerate solutions, we used the minimum triple-cosine (TC) based on the normalized component vectors \cite{ZiKi02}:
\begin{equation}
\text{TC} = \min_{r, s} \quad \V{a}_r\Tra\V{a}_s [\V{b}_1]_r\Tra[\V{b}_2]_s \V{c}_r\Tra\V{c}_s,
\end{equation}
excluding solutions with \(\text{TC} < -0.85\)~\cite{Br97}.
\Cref{fig:sim.nn.boxplot} shows the resulting distributions of FMS for all setups with different noise levels. We observe that the fully constrained models obtained a higher \(\FMS\) than the ALS algorithm for all setups. 
The largest improvement is for the \(\FMSB\) score, 
which is reasonable since the \(\BAll\) components are the most challenging components to recover. However, the AO-ADMM and HALS models also improve the \(\FMSA\) compared to ALS (AO-ADMM improves the \(\FMSC\) as well). The improvement grows for higher noise levels, demonstrating that imposing non-negativity for all modes instead of just the non-evolving modes makes the model more robust to noise. Furthermore, as for Setup 1, HALS was slower in terms of both time and iterations (see Figures SM5 and SM6, respectively).

\subsubsection{Setup 3: Unimodality constraints}\label{sec:simulation.unimodality}
\paragraph{Data generation}
For the unimodality constraints, we generated unimodal \(\Bk\) factor matrices similar to the approach in \cite{BeKeGiDe20}. However, we also wanted to examine the performance for data that does not exactly follow the PARAFAC2 constraint. For this, we generated the \(r\)-th column of \(\Bk\) as the normal probability density function
\begin{equation}
    [\V{b}_k]_r = \V{P}_\text{normal}(\mu_{r} + 0.41k, \sigma_{kr}),
\end{equation}
where \(\V{P}_\text{normal}(\mu, \sigma)\) is the PDF of a normal distribution sampled with 50 linearly spaced points between \(-10\) and \(10\) and with given mean and standard deviation, \(\mu_{r} \sim \mathcal{U}(-7, 0)\), \(\sigma_{kr} = \sigma_r + n_{kr}\) with \(\sigma_r \sim \mathcal{U}(0.5, 1)\) and \(n_{kr} \sim \mathcal{N}(0, 0.1)\). The factor 0.41 was chosen since that is equivalent to a shift of one index. Using a Gaussian PDF gives non-negative, unimodal components and increasing \(\mu\) for each slice produces shifting components, similar to setup 1. Furthermore, the random variation in \(\sigma_{kr}\) for each \(k\) leads to the components changing shape somewhat across slices in a way that will violate the PARAFAC2 constraint. Examining components that violate the PARAFAC2 constraint is interesting because real data might also not follow the PARAFAC2 constraint perfectly. The components are also highly collinear since the Gaussian curves are likely to overlap. Following this procedure, we constructed 50 datasets with five components, tensor sizes \(10 \times 50 \times 15\) and \(\eta = 0.33\). Subsection SM6.3 in the supplementary materials contains results from running the same experiment with \(\sigma_{kr} = \sigma_r\) so the components follow the PARAFAC2 constraint.

\paragraph{Experiment settings}
For the constrained model, we imposed non-negativity on \(\A\) and \(\DAll\), and unimodality as well as non-negativity on \(\BAll\). As baselines, we used AO-ADMM and HALS with non-negativity imposed on all modes, and ALS with non-negativity imposed only on \(\A\) and \(\DAll\). Each fitting algorithm ran until the stopping conditions were met or for a maximum of 2000 iterations.

During initial experiments, we observed that the feasibility gap (\(\fnorm{\Bk - \Pk \blueprint}\) and \(\fnorm{\Bk - \BkAux}\)) was too large for the AO-ADMM scheme to converge. This problem was mainly observed for the unimodality-constrained models, but it also occurred for the non-negativity constrained models. Therefore, we introduced three changes to the algorithm: We initialized the \(\A, \DAll, \blueprint\) and \(\Pk\)-matrices by running the ALS algorithm for one iteration before AO-ADMM, increased the automatically selected penalty parameters for the evolving mode, \(\rho_{\Bk}\)-s, by a factor of 10 and ran the inner (ADMM) iterations for at most 20 iterations instead of 5.

\begin{figure}
    \centering
    \includegraphics{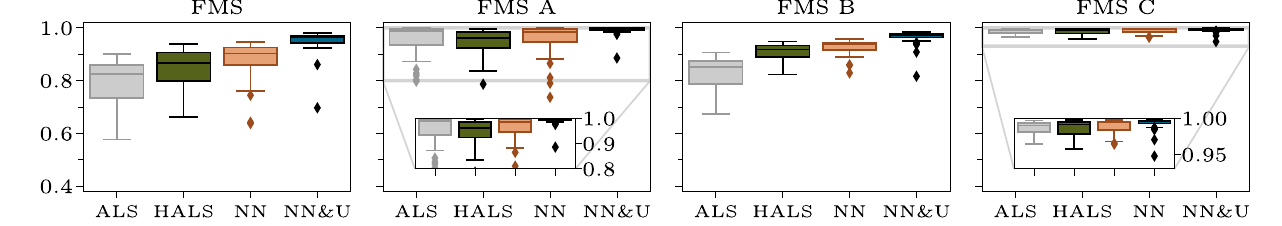}
    
    \caption{Setup 3: Boxplots showing the FMS for different models fitted to datasets with unimodal \(\Bk\)-matrices. ALS represents models fitted with non-negativity imposed on \(\A\) and \(\DAll\), HALS and NN represents models fitted with non-negativity imposed on all modes using HALS and AO-ADMM respectively, and NN\&U represents models fitted with non-negativity on all modes and unimodality imposed on \(\BAll\) using AO-ADMM.}
    \label{fig:sim.unimodal.boxplot}
\end{figure}

\begin{figure}
    \centering
    \begin{subfigure}{\textwidth}
        \includegraphics{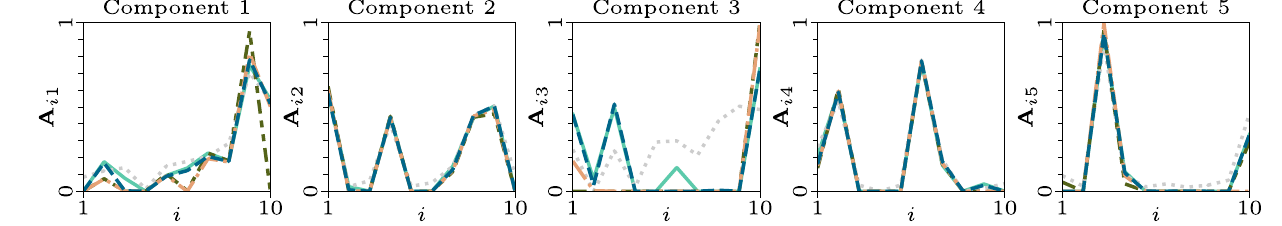}
    \end{subfigure}
    \begin{subfigure}{\textwidth}
        \includegraphics{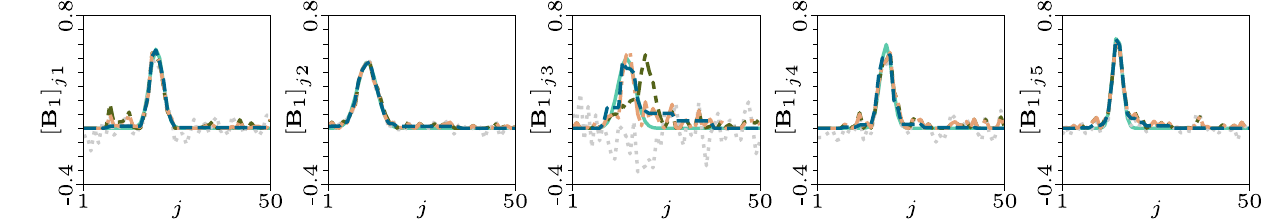}
    \end{subfigure}
    \begin{subfigure}{\textwidth}
        \includegraphics{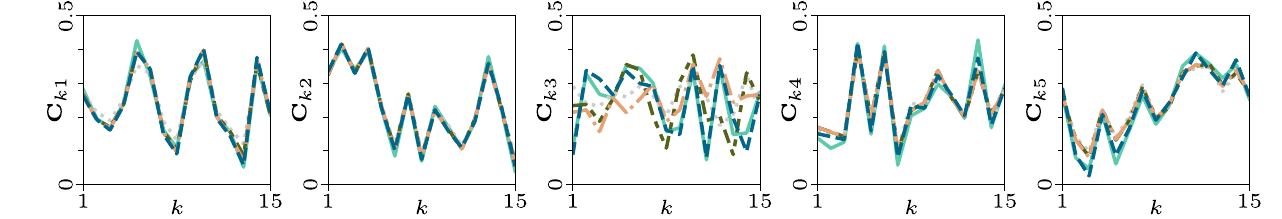}
    \end{subfigure}
    \begin{subfigure}{\textwidth}
        \includegraphics{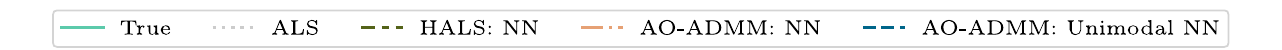}
    \end{subfigure}
    
    \caption{Setup 3: Plots showing the true and estimated components for one of the datasets. AO-ADMM: NN and HALS: NN represents the model fitted with non-negativity imposed on all modes using AO-ADMM and HALS, respecitvely, Unimodal NN represents the model fitted with non-negativity on all modes and unimodality imposed on \(\BAll\) using AO-ADMM, and ALS represents the model fitted with non-negativity imposed on \(\A\) and \(\DAll\). The cwSNRs (left to right) are \(-4.4, -2.6, -12, -11\) and \(-4.1\)~dB for this frontal slice.}
    \label{fig:sim.unimodal.factors}
\end{figure}

\begin{figure}
    \centering
    \includegraphics{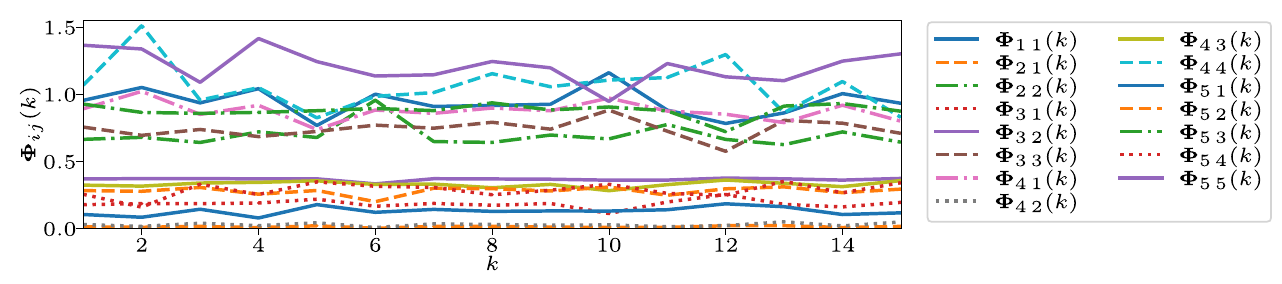}
    
    \caption{Setup 3: Plot showing how the cross product, \(\M{\Phi}(k) = \Bk\Tra\Bk\), changes across frontal slices for the same dataset as \cref{fig:sim.unimodal.factors}. }
    \label{fig:sim.unimodal.crossprod}
\end{figure}

\paragraph{Results}
As with Setup 2, we observed that the ALS scheme sometimes provided degenerate solutions, which were disregarded using the same heuristic. Datasets where all initializations fitted with ALS were degenerate were excluded from analysis, leaving 43 out of 50 datasets. \Cref{fig:sim.unimodal.boxplot} shows that models fitted with AO-ADMM obtained a higher FMS than those fitted with ALS and HALS\footnote{For timing plots, see Figure SM7 in the supplement.}. Furthermore, imposing non-negativity and unimodality on \(\BAll\) improved the FMS compared to only imposing non-negativity and substantially improved FMS compared to imposing no constraints on \(\BAll\). Also, we observed an increase in the recovery of all factor matrices for models fitted with unimodality constraints compared to those fitted without. In \cref{fig:sim.unimodal.factors}, we see the true and estimated factors for one of the datasets and the cross product of \(\Bk\) for each slice, \(k\), in this dataset is shown in \cref{fig:sim.unimodal.crossprod}. 
We see from the figure that the cross product is not constant. However, despite this, the components are recovered, which demonstrates that the PARAFAC2 AO-ADMM scheme works well even with slight violations of the PARAFAC2 constraint.


\subsubsection{Setup 4: Graph Laplacian regularization}
\paragraph{Data generation}
To assess graph Laplacian regularization, we simulated \emph{smooth} components, using a simulation setup inspired by \cite{He17}. For this setup, we first select a \(D\)-dimensional space of smooth vectors, \(\mathcal{V}\), and construct an orthonormal matrix \(\M{M}\) whose columns span \(\mathcal{V}\). Next, we observe that if we have a collection of matrices, \(\ForAllK{\smash{\tilde{\M{B}}_k}} \in \PSet\), then we can multiply each factor matrix, \(\tilde{\M{B}}_k\), with \(\M{M}\), thus obtaining a new collection of matrices, \(\ForAllK{\smash{{\M{B}}_k}} = \ForAllK{\smash{\M{M}\tilde{\M{B}}_k}} \in \PSet\), whose columns are smooth.

We constructed \(\M{M}\) from a basis for cubic polynomials on the interval \([-1, 1]\),
orthogonalized using the SVD. To generate random \(\ForAllK{\smash{\tilde{\M{B}}_k}} \in \PSet\), we set \(\smash{\tilde{\M{B}}_k} = \Pk \blueprint\), drawing the elements of \(\blueprint\) from a standard normal distribution and generated the \(\Pk\)-matrices as random orthogonal matrices (obtained from the QR factorization: \(\Pk \M{R} = \M{S}_k\), where \([\M{S}_k]_{ij}\sim\mathcal{N}(0, 1)\)). Following this procedure, we set \(R=3\) and constructed 20 different \(30 \times 200 \times 30\) simulated data tensors with \(\eta = 0.5\).

\paragraph{Experiment settings}
Since the components are smooth, it is reasonable to penalize local differences. Thus, we constructed the similarity graph, such that neighbouring indices in the vectors had a similarity score of 1, and all non-neighbouring indices of the vectors had a similarity score of 0. 
To decide the strength of the regularization penalty, we conducted a grid-search of four regularization parameters logarithmically spaced between \(1\) and \(1000\). Also, following \cref{th:reg.scaling}, we used ridge regularization on \(\A\) and \(\DAll\), setting the regularization strength to 0.1. We compared with models without ridge regularization to show that a penalty-based regularization on one mode requires penalizing the norm of the remaining factor matrices. In addition, we compared with an ALS baseline without regularization. All models were ran until the stopping conditions were met or they reached 5000 iterations.

\paragraph{Results}
From \cref{fig:sim.penalty.boxplot,fig:sim.penalty.factors}, we see that imposing smoothness regularization yielded smoother components and a higher FMS compared to using ALS without regularization. Furthermore, ridge regularization on \(\A\) and \(\DAll\) was essential for obtaining improvement with graph Laplacian regularization on the \(\Bk\)-matrices, which is expected from the observations in \cref{sec:scale.reg}. To ensure that this was not an artifact from the initialization selection scheme, we also compared the initializations that obtained the highest FMS for each dataset, which showed the same behavior (see supplement, Figure SM12). Another notable observation is that none of the models fitted without ridge regularization on \(\A\) and \(\DAll\), but with regularization on \(\Bk\) converged (see supplement, Table SM5).
Moreover, when we imposed regularization on only the \(\Bk\)-matrices, the regularization penalty (when calculated after normalizing the \(\Bk\)-matrices) decreased only in early iterations before approaching the same value as the unregularized (ALS) model. Figure SM10 in the supplement shows a more detailed visualization. This behavior demonstrates that if we use penalty based regularization on the \(\Bk\)-matrices without penalizing the norm of the \(\A\)-matrix and \(\Dk\)-matrices, then the regularization will have little effect.

\subsubsection{Setup 5: Total variation regularization}
\paragraph{Data generation}
For assessing TV regularization, we simulated piecewise constant components for ragged tensors. To construct these components, we used the following scheme: For each \(\Bk\)-matrix, we generated a random binary matrix with orthogonal columns, \(\hat{\M{Q}}_k \in \Real^{J_k \times 4}\), and at most two discontinuities per column (i.e., piecewise constant). Then, we normalized these matrices, obtaining orthonormal matrices, \(\M{Q}_k\), and set \(\Bk = \M{Q}_k \M{\Omega}\), where \(\M{\Omega} \in \Real^{4 \times R}\) has two non-zero elements per column, drawn from a standard normal distribution. This scheme led to piecewise constant components with at most four jumps. We  set \(R=3\) and constructed 20 such simulated ragged data tensors with 30 frontal slices of size \(30 \times J_k\), with \(J_k\) drawn uniformly between \(200\) and \(250\). Following \cref{eq:noise.level}, we added noise with \(\eta = 0.5\).

\paragraph{Experiment settings}
We imposed TV regularization on the \(\Bk\)-matrices and ridge regularization on \(\A\) and \(\DAll\). Setting the ridge regularization parameter to both \(0\) and \(0.1\), we performed a grid search for the TV regularization parameter with five logarithmically spaced values between \(0.001\) and \(10\). As baseline we used ALS without regularization on \(\BAll\). All methods were run until the stopping conditions were fulfilled or they reached 5000 iterations.

\begin{figure}
    \centering
    \begin{subfigure}{0.49\textwidth}
     \includegraphics{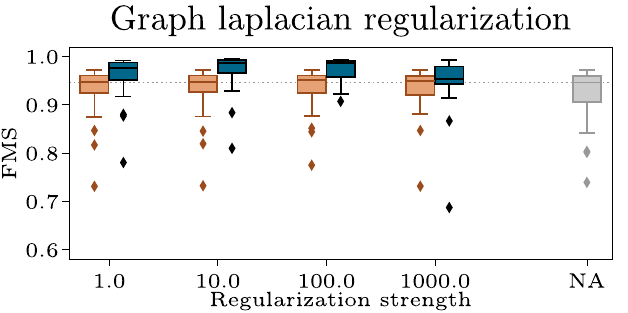}
    \end{subfigure}
    \begin{subfigure}{0.49\textwidth}
     \includegraphics{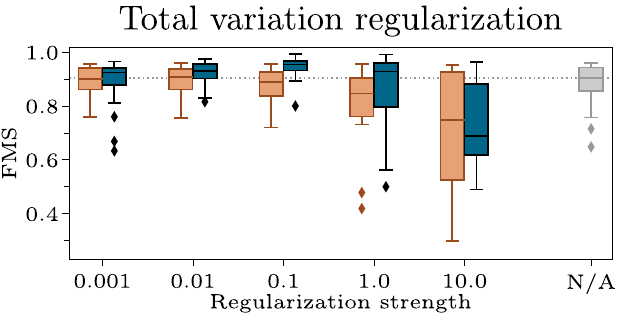}
    \end{subfigure}
     \includegraphics{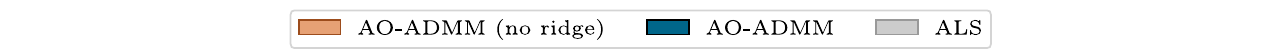}
    \caption{Setup 4 and 5: Boxplots showing the FMS obtained with different levels of regularization.}
    \label{fig:sim.penalty.boxplot}
\end{figure}

\begin{figure}
    \centering
    \begin{subfigure}{\textwidth}
     \centering
     \includegraphics{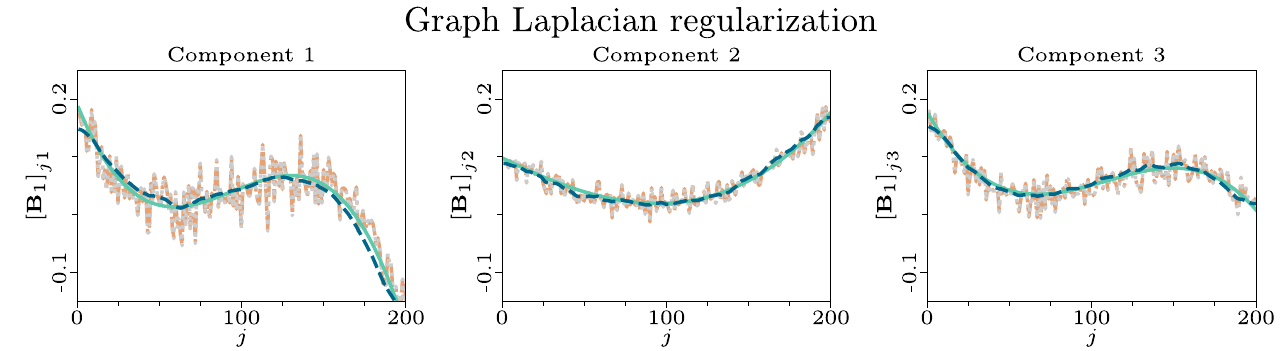}
    \end{subfigure}
    \begin{subfigure}{\textwidth}
     \centering
     \vspace{0.5em}
     \includegraphics{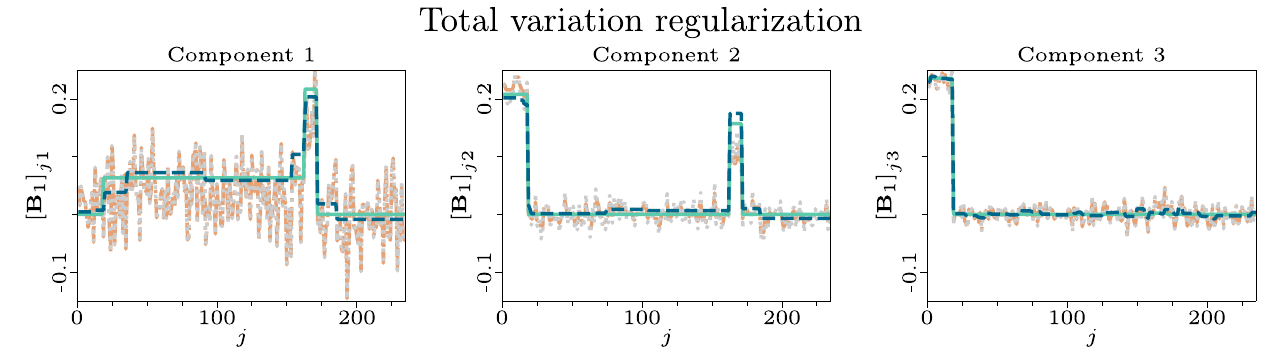}
    \end{subfigure}
     \includegraphics{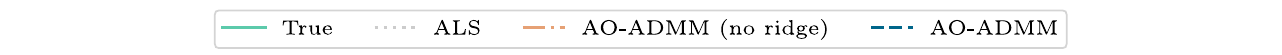}
    \caption{Setup 4 and 5: Plots showing true and estimated components. On the top, we see one of the datasets where graph Laplacian regularization was used to obtain smooth components and on the bottom we see another dataset where TV regularization was used to obtain piecewise constant components. The cwSNRs for the top row are \(-8.6,  0.031\) and \(-3.7\), respectively. For the bottom row, the cwSNRs are \(-14,  -5.6\) and \(0.36\), respectively.}
    \label{fig:sim.penalty.factors}
\end{figure}

\paragraph{Results}
\Cref{fig:sim.penalty.boxplot} shows that the TV-regularized PARAFAC2 AO-ADMM algorithm performs better than the unregularized ALS algorithm for most regularization strengths. Moreover, from \cref{fig:sim.penalty.factors} we see that the TV regularized factors are piecewise constant when ridge was imposed on \(\A\) and \(\DAll\). Similar to setup 4, we observe that for AO-ADMM, none of the initializations converged without ridge regularization on \(\A\) and \(\DAll\) (see supplement, Table SM6) and the regularization (when calculated after normalizing the \(\Bk\)-matrices) decreased only in early iterations before increasing (see supplement Figure SM11). 

\subsubsection{Setup 6: Evaluating the cwSNR}\label{sec:cwsnr.experiment}
\paragraph{Dataset generation}
To evaluate the cwSNR ability to assess the recovery of the \(\Bk\) components, we simulated five different 5-component datasets of size \(30 \times 40 \times 100\). For the \(\Bk\) factor matrices, we drew the elements of \(\B_1\) from a truncated normal distribution and then shifted them cyclically by \(k\) entries. After constructing the data tensors, we added noise following \cref{eq:noise.level} with \(\eta = 0.1\), \(\eta = 0.33\) and \(\eta = 0.5\).

\paragraph{Experiment setup}
Each model was ran for 2000 iterations or until the stopping conditions were satisfied. Then, to assess the degree of recovery for a given \(\left[\smash{\hat{\V{b}}}_k\right]_{r},\) we used the cosine similarity (SIM), given by (for normalized component vectors)
\begin{equation}
    \text{SIM}_{kr} = {\left[\smash{\V{b}}_k\right]_{r}\Tra} \left[\smash{\hat{\V{b}}}_k\right]_{r}.
\end{equation}

\begin{figure}
    \centering
    \includegraphics{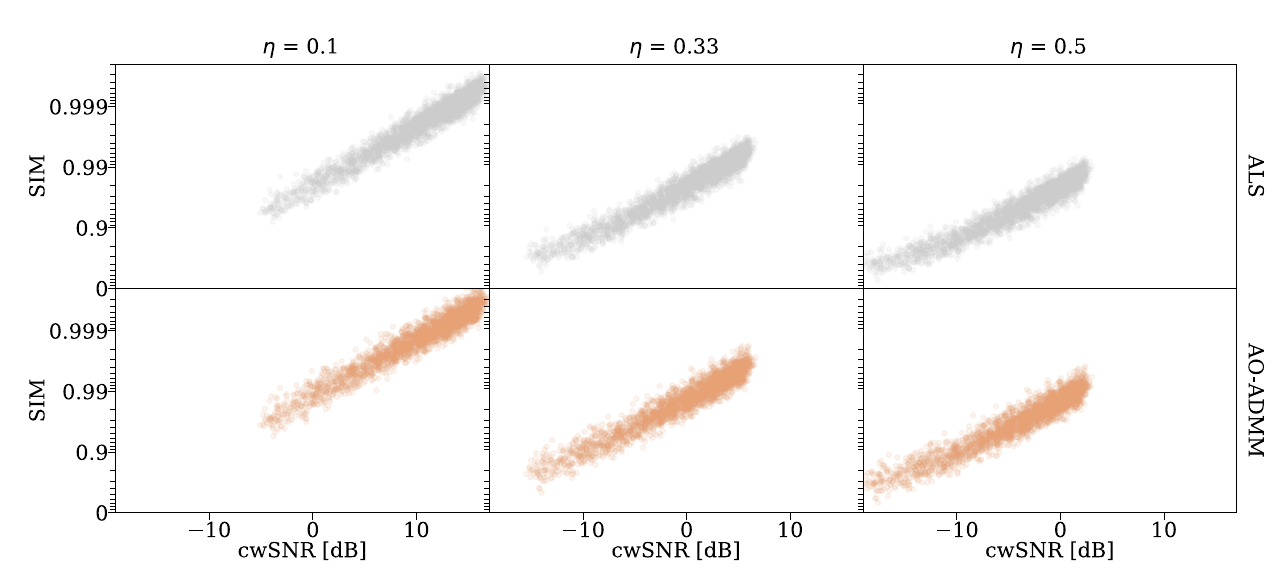}  
    \caption{Setup 6: Scatter plot showing factor recovery (SIM) plotted against cwSNR for models fitted with non-negativity on all modes (AO-ADMM) and non-negativity only on \(\A\) and \(\C\) (ALS). Each dot represents the cosine similarity score for one column of one \(\Bk\) factor matrix for one dataset (all five datasets are aggregated in each plot).}
    \label{fig:noise.vs.c}
\end{figure}

\paragraph{Results}
\Cref{fig:noise.vs.c} shows the results from these experiments. The top row shows the cosine similarity scores obtained with non-negativity on \(\A\) and \(\DAll\) (ALS), the bottom row shows the recovery scores obtained with non-negativity on all modes (AO-ADMM) and the columns correspond to the different noise levels. We see that the expected recovery score grows monotonically with the cwSNR for both models.

\subsection{Neuroimaging application}
\begin{figure}
    \centering
    \includegraphics{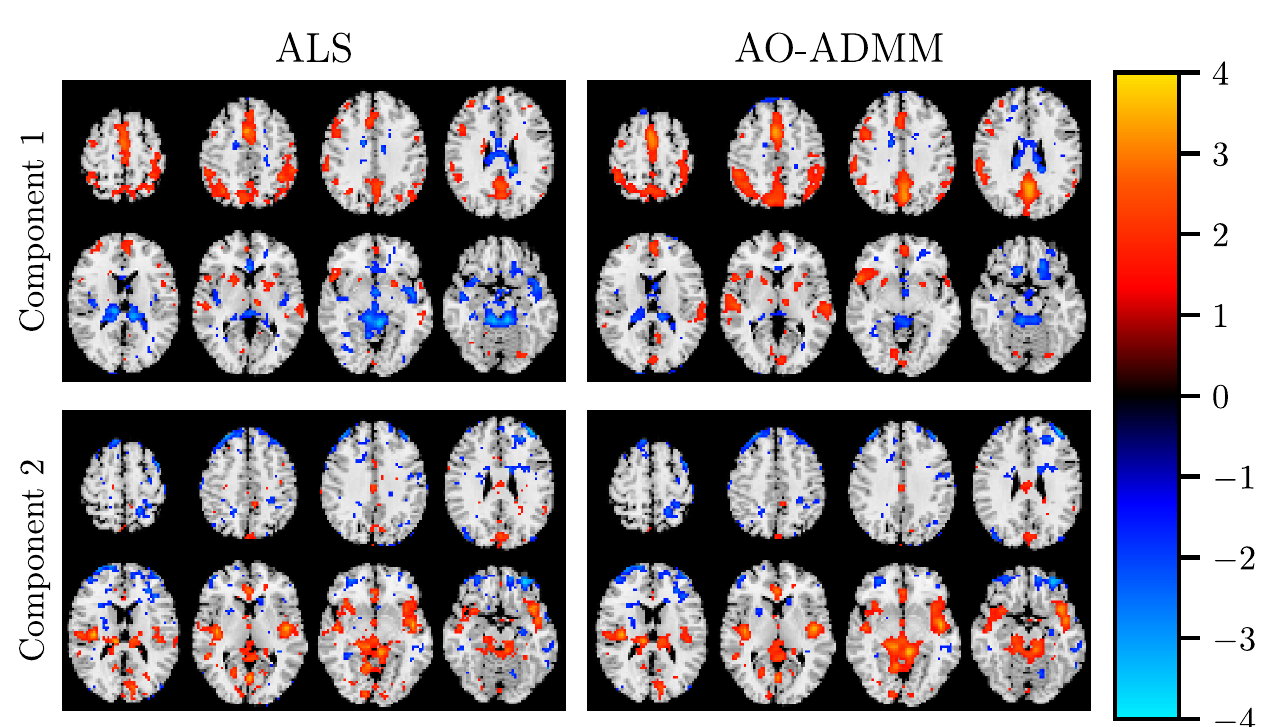}
    \caption{Voxel-mode components of the two-component unregularized (left) and regularized (right) PARAFAC2 models fitted to the fMRI data. The composite plots are generated by z-scoring the components and only showing voxels with activation \(>1.5\). 
    The regularization penalties (after normalization) for the voxel-mode components are 10.4 and 17.6 for the regularized and unregularized model, respectively, which demonstrates that the regularization quantifiably increases smoothness. We see that the regularized model has less speckle noise and more defined regions of activation. Furthermore, the regularized components are more focal (i.e., have high intensity in the mid-parts), which is expected of spatial fMRI maps, and thus they lend themselves to better association with physical quantities of interest. For the component that shows a statistically significant difference between the two groups (component 1), we see activation in the primary and secondary motor and cerebellum, and also auditory cortex, which are regions expected to be activated by the SM task. This activation is less noisy and easier to discern for the regularized model.}
    \label{fig:fmri}
\end{figure}

For neuroimaging applications, our motivation for imposing constraints on PARAFAC2 components is to improve their interpretability. Previous work has shown that PARA\-FAC2 can reveal dynamic networks (spatial dynamics) from fMRI data arranged as a \emph{subjects} by \emph{voxels} by \emph{time windows} tensor \cite{RoShJi19}. However, the model is prone to noise affecting the extracted networks (see \cref{fig:fmri}). To investigate if smoothness inducing regularization can improve interpretability, we used images from the MCIC collection \cite{fMRI:MCIC}, which contains fMRI-scans from healthy controls and patients with schizophrenia. We used the sensory motor (SM) task data, with the same feature extraction and preprocessing as  \cite{RoShJi19}.

We analyzed the fMRI data tensor using both regularized PARAFAC2 fitted with AO-ADMM and unregularized PARAFAC2 fitted with ALS, both ran for at most 8000 iterations. The \(\A\), \(\Bk\) and \(\DAll\) factor matrices represented the subject-mode, voxel-mode and time-mode components, respectively. This configuration allows the voxel-mode components that indicate activation networks to evolve over time. To reduce the noise in the components, we chose a graph Laplacian regularizer based on the image gradient, thus penalizing large differences in neighboring voxels. We imposed ridge regularization on \(\A\) and \(\DAll\) to resolve the scaling indeterminacy, and non-negativity constraints on \(\DAll\) to resolve the sign indeterminacy.

The spatial regularizer encourages similar activation for neighboring voxels, which is a reasonable assumption for fMRI images \cite{BaWaTsWaDa17}. Specifically, we used graph Laplacian regularization where \(w_{ij}=1\) if voxel \(i\) and \(j\) are neighboring voxels (in the von Neumann sense) and \(w_{ij}=0\) otherwise. The proximal operator of this penalty function involves solving a large linear system, similar to that obtained when solving Helmholtz’ equation with a finite difference scheme. To solve this system, we, therefore, used the conjugate gradient method, with a smoothed aggregation preconditioner, setting the near-nullspace component to a vector consisting only of ones.

We fixed the ridge penalty to \(0.1\), and performed a grid-search for the optimal smoothness penalty. The selected regularization parameter, \(10\),
provided the strongest smoothing-effect while also not deteriorating the interpretability of the model. 

\Cref{fig:fmri} shows a comparison of the unregularized and the regularized model. We see that imposing regularization yields smoother, more interpretable, components.
The regularized sensorimotor component (i.e., component 1) shows the same significantly stronger activation in healthy controls compared to patients as the unregularized component
(the $p$-values obtained using a two-sample t-test on the corresponding subject-mode component vector is \(2.29 \times 10^{-4}\) for AO-ADMM and \(2.98 \times 10^{-4}\) for ALS) while the regions of activation are less affected by noise, and thus the interpretation is clearer. 

\subsection{Chemometrics application}
Another application where PARAFAC2 is well-suited is the analysis of gas chromatography mass spectrometry (GC-MS) measurements \cite{AmSkBrCoMa08,BrAnKi99}. In GC-MS measurements, a sample is injected into a column that lets different molecules pass through it at different velocities depending amongst others on their affinity to the column (GC). At distinct time points at the end of the column, the mass spectrum is measured (MS). From this, we obtain a three-way tensor where one mode represents the mass spectrum, one mode represents the time after injecting the sample (retention time), and one mode represents the different samples. When we analyze the GC-MS data with PARAFAC2, we obtain components that represent the mass spectrum (\(\A\)) and elution profile (\(\BAll\)) for the different chemical compounds as well as their concentration in different samples (\(\DAll\)).

The dataset used here stems from a project on fermentation of apple wine using different microorganisms. The data is typical for untargeted chemical profiling as commonly done in flavor research. Samples were taken every eight hours from the headspace of the fermentation tank throughout the fermentation and measured on an Agilent 5973 MSD, Agilent Technologies, Santa Clara, USA.

Gathering GC-MS measurements can be time-consuming, and it is therefore of interest to enable the analysis of smaller GC-MS datasets. The quality of the analysis may depend on the number of samples, \(K\). 
For this dataset, if we analyze all of the samples, constraints on \(\Bk\) are not needed for recovery and PARAFAC2 with ALS is, therefore, sufficient.
However, constraining all modes may reduce the number of samples required for accurately capturing the underlying components. To evaluate whether constraining \(\BAll\) reduces the number of samples needed, we constructed the data tensor from only a third of the 57 samples (i.e., the first 19 samples), forming a \(286 \times 95 \times 19\) tensor. We then analyzed the GC-MS tensor using PARAFAC2 fitted with both AO-ADMM and ALS, both ran for at most 6000 iterations. As for the simulations, we ran 20 random initializations, selecting the runs with the lowest SSE.

PARAFAC2 is well suited because it allows the elution profiles to differ across different samples. Thus, PARAFAC2 can account for retention shifts and other shape changes that can occur between different samples containing the same chemical.  Neither concentrations, elution profiles nor mass spectra should contain negative entries, so we impose non-negativity constraints on all factor matrices for PARAFAC2 with AO-ADMM  and on the \(\A\) and \(\DAll\) factor matrices for PARAFAC2 with ALS.

\begin{figure}
    \centering
    \begin{subfigure}{\textwidth}
    \centering
        \includegraphics[width=0.95\textwidth]{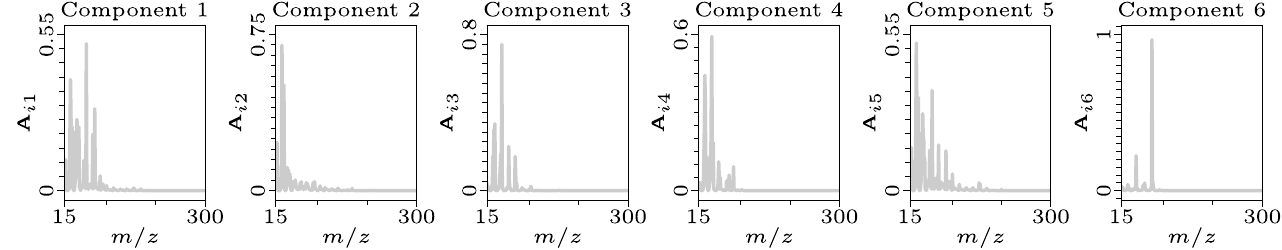}
    \end{subfigure}
    \begin{subfigure}{\textwidth}
    \centering
        \includegraphics[width=0.95\textwidth]{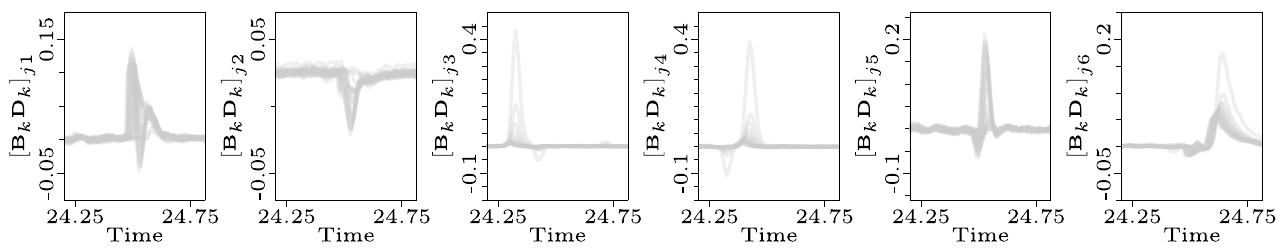}
        \caption{ALS (non-negativity on \(\A\) and \(\DAll\)).}
    \end{subfigure}
    \begin{subfigure}{\textwidth}
    \centering
        \includegraphics[width=0.95\textwidth]{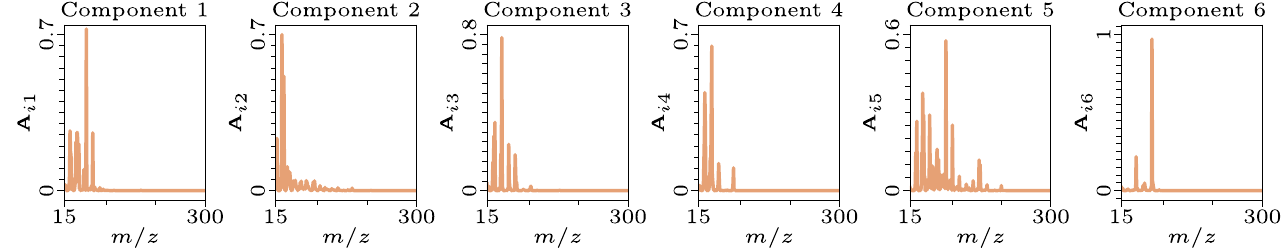}
    \end{subfigure}
    \begin{subfigure}{\textwidth}
    \centering
        \includegraphics[width=0.95\textwidth]{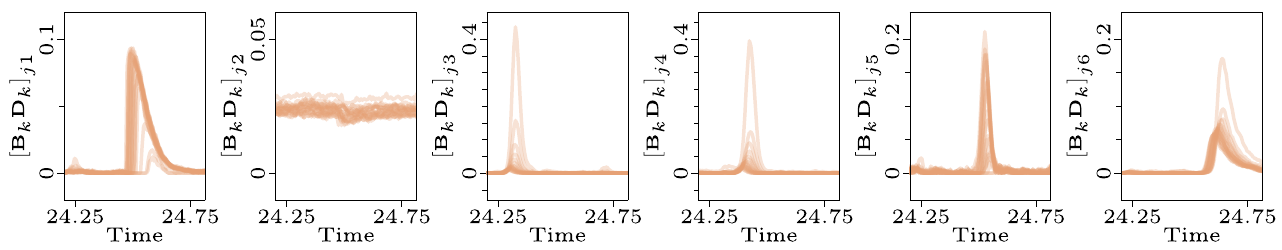}
        \caption{AO-ADMM (non-negativity on all modes).}
    \end{subfigure}
    \caption{Mass spectra (top) and elution profiles scaled by concentration (bottom). For both models, the second component represents noise and the other components represent chemical information.}
    \label{fig:gcms.factorplot}
\end{figure}

\Cref{fig:gcms.factorplot} shows the 6-component models obtained with ALS and AO-ADMM. We see that both models include one noise component and five components that contain chemical information. However, the model fitted without non-negativity on \(\Bk\) yields unphysical elution profiles (negative peaks and multiple peaks), which demonstrates the need for more samples if we do not impose constraints on \(\BAll\). For the model fitted with non-negativity on all modes, these phenomena are much less apparent, and we only observe minor secondary peaks, particularly for low-concentration samples. Thus, constraining the \(\Bk\)-matrices can improve the interpretative value of the PARAFAC2 components and reduce the number of samples required for analysis.

\section{Discussion}\label{sec:discussion}
In this paper, we have proposed an efficient algorithmic framework based on AO-ADMM that allows us to fit PARAFAC2 models with any proximable regularization penalty on all modes. Furthermore, we show that our approach can successfully apply hard constraints, e.g., non-negativity and unimodality, or penalty based regularization such as TV and graph Laplacian regularization on \(\BAll\). Our numerical experiments show that the AO-ADMM framework is accurate, and faster than the flexible coupling approach for non-negative PARAFAC2 and that their relative performance was unchanged with under- and over-estimation of the rank.

With both simulated data and real applications, we demonstrated that constraining the \(\Bk\)-matrices, can improve recovery and interpretability. In particular, constraints can be necessary to recover the components accurately from noisy data when the component vectors are highly collinear. We also showed that the proposed scheme recovered the components even in cases where the PARAFAC2 constraint is slightly violated. 
Furthermore, we demonstrated that all modes should be regularized when the regularization penalty of one mode is dependent on the norm.

Nevertheless, there are possibilities for improvement. For example, the ADMM update steps are based on the projection \(\proxF{\ConstraintIndicatorF}\), for which we proposed an efficient approximation algorithm based on AO. However, the mathematical properties of \(\PSet\) are, to the best of our knowledge, still not well understood, and future analysis may improve upon this projection.
Another possible extension is support for other loss functions, such as the KL-divergence or weighted least squares. Our scheme could also be extended to other models such as block term decomposition 2 (BTD2) \cite{ChKoMoTh19}, which is a four-way extension of the block term decomposition (BTD) \cite{De08} with evolving factor matrices that follow the PARAFAC2 constraint. AO-ADMM has already been used to fit BTD models \cite{GuPaPa20}, and our work could be straightforwardly adapted for fitting constrained BTD2 models with AO-ADMM.
It may also be beneficial to use Nesterov-type extrapolation, which has been successfully used to speed up ALS schemes for CP \cite{MiYeSt20,AnCoKhGi20} as well as ALS and flexible coupling schemes for PARAFAC2 \cite{YuAuBr21}.

\appendix

\section*{Acknowledgments}
We would like to thank Jesper Løve Hinrich and Remi Cornillet for insightful discussions on constrained PARAFAC2. This work has benefited from the Experimental Infrastructure for Exploration of Exascale Computing (eX3), which is supported by the Research Council of Norway under contract 270053.

\printbibliography

@Article{AcBiBiBr07,
  author =        {Evrim Acar and Canan A. Bingol and Haluk Bingol and Rasmus Bro and Bulent Yener},
  title =         {Multiway Analysis of Epilepsy Tensors},
  journal =       {Bioinformatics},
  year =          {2007},
  volume =        {23},
  number =        {13},
  pages =         {i10-i18}
}

@ARTICLE{Hi27a,
  author = {F. L. Hitchcock},
  title = {The Expression of a Tensor or a Polyadic as a Sum of Products},
  journal = {J. Math. Phys.},
  year = {1927},
  volume = {6},
  pages = {164--189},
  number = {1},
}

@Article{AcYe09,
  author =        {Evrim Acar and Bulent Yener},
  title =         {Unsupervised Multiway Data Analysis: A Literature Survey},
  journal =       {IEEE Trans. Knowl. Data Eng.},
  year =          {2009},
  volume =        {21},
  number =        {1},
  pages =         {6--20},
  month =         jan,
}

@Article{KoHo20,
  author  = {T. G. Kolda and D. Hong},
  title   = {Stochastic Gradients for Large-Scale Tensor Decomposition},
  journal = {SIAM J. Math. Data Sci.},
  year    = {2020},
  volume  = {2},
  number  = {4},
  pages   = {1066–1095},
}

@InProceedings{MaHaKa16,
  author    = {T. Maehara and K. Hayashi and K. Kawarabayashi},
  title     = {Expected Tensor Decomposition with Stochastic Gradient Descent},
  booktitle = {AAAI'16: Proc. 30th AAAI Conf. Artif. Intell.},
  year      = {2016},
  pages     = {1919–1925},
}

@Article{MaMaGi15,
  author  = {M. Mardani and G. Mateos and G. B. Giannakis},
  title   = {Subspace Learning and Imputation for Streaming Big Data Matrices and Tensors},
  journal = {IEEE Trans. Signal Process.},
  year    = {2015},
  volume  = {63},
  number  = {10},
  pages   = {2663-2677},
}

@InProceedings{VaVeLa17,
  author    = {M. Vandecappelle and N. Vervliet and L. De Lathauwer},
  title     = {Nonlinear Least Squares Updating of the Canonical Polyadic Decomposition},
  booktitle = {EUSIPCO'17: Proc. 25th Eur. Signal Proc. Conf.},
  organization={EURASIP},
  year      = {2017},
  pages     = {663-667},
}

@Article{NiSi09,
  author    = {D. Nion and N. D. Sidiropoulos},
  title     = {Adaptive Algorithms to Track the {PARAFAC} Decomposition of a Third-Order Tensor},
  journal   = {IEEE Trans. Signal Process.},
  year      = {2009},
  volume    = {57},
  pages     = {2299-2310}
}

@Article{AcDuKo10b,
  author =        {Daniel M. Dunlavy and Tamara G. Kolda and Evrim Acar},
  title =         {Temporal Link Prediction using Matrix and Tensor Factorizations},
  journal =       {ACM TKDD},
  year =          {2011},
  volume =        {5},
  number =        {2},
  pages =         {Article 10},
}

@Article{AmSkBrCoMa08,
  author  = {José Manuel Amigo and Thomas Skov and Rasmus Bro and Jordi Coello and Santiago Maspoch},
  title   = {Solving {GC-MS} problems with {PARAFAC2}},
  journal = {TrAC Trends Anal. Chem.},
  year    = {2008},
  volume  = {27},
  number  = {8},
  pages   = {714 - 725},
  issn    = {0165-9936},
}

@Article{MaChMo17,
  author   = {Madsen, K.H. and Churchill, N.W. and Mørup, M.},
  title    = {Quantifying functional connectivity in multi-subject {fMRI} data using component models},
  journal  = {Human Brain Mapping},
  year     = {2017},
  volume   = {38},
  number   = {2},
  pages    = {882-899}
}

@Article{PaFaSi16,
  author =  {E. E. Papalexakis and C. Faloutsos and N. D. Sidiropoulos},
  title =   {Tensors for Data Mining and Data Fusion: Models, Applications, and Scalable Algorithms},
  journal = {ACM Trans. Intell. Syst. Technol.},
  year =    {2016},
  volume =  {8},
  number =  {2},
  pages =   {Article 16}
}

@inproceedings{YiAfHoChZhSu20,
  title={{LogPar}: Logistic {PARAFAC2} Factorization for Temporal Binary Data with Missing Values},
  author={Yin, Kejing and Afshar, Ardavan and Ho, Joyce C and Cheung, William K and Zhang, Chao and Sun, Jimeng},
  booktitle={KDD'20: Proc. 26th ACM SIGKDD Int. Conf. Knowl. Discov. and Data Mining},
  pages={1625--1635},
  year={2020}
}

@misc{Co17,
    year={2019},
    title={Software},
    author={Condat, Laurent},
    url={https://lcondat.github.io/software.html},
    note={[Accessed: 2020-10-19]}
}

@article{Co13,
  title={A direct algorithm for 1-{D} total variation denoising},
  author={L. Condat},
  journal={IEEE Signal Process. Letters},
  volume={20},
  number={11},
  pages={1054--1057},
  year={2013},
  publisher={IEEE}
}

@InProceedings{ScCoAc20b,
  author       = {Carla Schenker and J. E. Cohen and E. Acar},
  title        = {An optimization framework for regularized linearly coupled matrix-tensor factorization},
  booktitle    = {EUSIPCO'20: Proc. 2020 28th Eur. Signal Process. Conf.},
  organization = {EURASIP},
  year         = {2021},
  pages        = {985-989},
}

@article{ScCoAc20,
  title={A Flexible Optimization Framework for Regularized Matrix-Tensor Factorizations with Linear Couplings},
  author={Schenker, Carla and Cohen, Jeremy E and Acar, Evrim},
  journal={IEEE J. Sel. Top. Signal Process.},
  volume={15},
  number={3},
  pages={506--521},
  year={2021},
  publisher={IEEE}
}

@article{He13,
  title={The special sign indeterminacy of the direct-fitting {P}arafac2 model: Some implications, cautions, and recommendations for Simultaneous Component Analysis},
  author={N. E. Helwig},
  journal={Psychometrika},
  volume={78},
  number={4},
  pages={725--739},
  year={2013},
  publisher={Springer}
}

@article{KoBa09,
  title={Tensor decompositions and applications},
  author={T. G. Kolda and B. W. Bader},
  journal={SIAM Rev.},
  volume={51},
  number={3},
  pages={455--500},
  year={2009},
  publisher={SIAM}
}

@Article{Ha72,
  author  = {R. A. Harshman},
  journal = {UCLA working papers in phonetics},
  title   = {{PARAFAC2}: {M}athematical and technical notes},
  year    = {1972},
  pages   = {30--44},
  volume  = {22},
}

@InProceedings{CoBr18,
  author    = {J. E. Cohen and R. Bro},
  booktitle = {LVA/ICA'18},
  title     = {Nonnegative {PARAFAC2}: A Flexible Coupling Approach},
  year      = {2018},
  pages     = {89--98},
}

@Article{Ha70,
  author  = {R. A. Harshman},
  journal = {UCLA working papers in phonetics},
  title   = {Foundations of the {PARAFAC} procedure: Models and conditions for an ``explanatory'' multi-modal factor analysis},
  year    = {1970},
  pages   = {1-84},
  volume  = {16},
}

@Article{CaCh70,
  author  = {J. D. Carroll and J. J. Chang},
  journal = {Psychometrika},
  title   = {Analysis of individual differences in multidimensional scaling via an {N}-way generalization of ``{Eckart-Young}'' decomposition},
  year    = {1970},
  issn    = {1860-0980},
  number  = {3},
  pages   = {283--319},
  volume  = {35},
}

@InProceedings{AfPePaSeHoSu18,
  author    = {A. Afshar and I. Perros and E. E. Papalexakis and E. Searles and J. Ho and J. Sun},
  booktitle = {CIKM'18: Proc. ACM Int. Conf. Inf. Knowl. Manag.},
  title     = {{COPA}: {C}onstrained {PARAFAC2} for {S}parse \& {L}arge {D}atasets},
  year      = {2018},
  pages     = {793--802},
  numpages  = {10},
}

@Article{BrAnKi99,
  author   = {R. Bro and C. A. Andersson and H. A. L. Kiers},
  journal  = {J. Chemom.},
  title    = {{PARAFAC2 - Part II. Modeling chromatographic data with retention time shifts}},
  year     = {1999},
  issn     = {0886-9383},
  number   = {3-4},
  pages    = {295--309},
  volume   = {13},
}

@Article{HuSiLi16,
  author    = {K. Huang and N. D. Sidiropoulos and A. P. Liavas},
  journal   = {IEEE Trans. Signal Process.},
  title     = {A flexible and efficient algorithmic framework for constrained matrix and tensor factorization},
  year      = {2016},
  number    = {19},
  pages     = {5052--5065},
  volume    = {64},
  publisher = {IEEE},
}

@article{PaBo14,
  title={Proximal algorithms},
  author={Parikh, Neal and Boyd, Stephen},
  journal={Found. Trends Mach. Learn.},
  volume={1},
  number={3},
  pages={127--239},
  year={2014},
  publisher={Now Publishers Inc. Hanover, MA, USA}
}

@Article{BoPaChPeEc11,
  author  = {S. Boyd and N. Parikh and E. Chu and B. Peleato and J. Eckstein},
  journal = {Found. Trends Mach. Learn.},
  title   = {Distributed optimization and statistical learning via the alternating direction method of multipliers},
  year    = {2011},
  month   = jan,
  number  = {1},
  pages   = {1--122},
  volume  = {3},
}

@Article{KiTeBr99,
  author  = {H. A. L. Kiers and J. M. F. {Ten Berge} and R. Bro},
  journal = {J. Chemom.},
  title   = {{PARAFAC2} - {Part I. A} direct fitting algorithm for the {PARAFAC2} model},
  year    = {1999},
  number  = {3-4},
  pages   = {275--294},
  volume  = {13},
}

@Article{He17,
  author   = {N. E. Helwig},
  journal  = {Biom. J.},
  title    = {Estimating latent trends in multivariate longitudinal data via {Parafac2} with functional and structural constraints},
  year     = {2017},
  number   = {4},
  pages    = {783-803},
  volume   = {59},
}

@InProceedings{LeAc22,
  author    = {I. Lehmann and E. Acar and T. Hasija and M. A.B.S. Akhonda and V. D. Calhoun and P. J. Schreier and T. Adali},
  title     = {Multi-task {fMRI} Data Fusion using {IVA} and {PARAFAC2}},
  booktitle = {ICASSP'22: Proc. Int. Conf. Acoust., Speech, Signal Process.},
  year      = {2022},
}

@Article{WiKiWa18,
  author  = {A. H. Williams and T. H. Kim and F. Wang and S. Vyas and S. I. Ryu and K. V. Shenoy and M. Schnitzer and T. G. Kolda and S. Ganguli},
  title   = {Unsupervised Discovery of Demixed, Low-Dimensional Neural Dynamics across Multiple Timescales through Tensor Component Analysis},
  journal = {Neuron},
  year    = {2018},
  volume  = {98},
  number  = {6},
  pages   = {1099-1115},
}

@InProceedings{RoShJi19,
  author      = {M. Roald and S. Bhinge and C. Jia and V. Calhoun and T. Adali and E. Acar},
  title       = {Tracing Network Evolution using the {PARAFAC2} model},
  booktitle   = {ICASSP'20: Proc. Int. Conf. Acoust., Speech, Signal Process.},
  year        = {2020},
  institution = {arXiv:1911.02926v1},
}

@InProceedings{ChBaBr07,
  author    = {P. A. Chew and B. W. Bader and T. G. Kolda and A. Abdelali},
  booktitle = {KDD'07: Proc. 13th ACM SIGKDD Int. Conf. Knowl. Discov. and Data Mining},
  title     = {Cross-Language Information Retrieval Using {PARAFAC2}},
  year      = {2007},
  pages     = {143–152},
  numpages  = {10},
}

@article{MoHaHePaAr06,
  title={Parallel factor analysis as an exploratory tool for wavelet transformed event-related {EEG}},
  author={M{\o}rup, Morten and Hansen, Lars Kai and Herrmann, Christoph S and Parnas, Josef and Arnfred, Sidse M},
  journal={NeuroImage},
  volume={29},
  number={3},
  pages={938--947},
  year={2006},
  publisher={Elsevier}
}

@article{Br97,
  title={{PARAFAC}. Tutorial and applications},
  author={Bro, Rasmus},
  journal={Chemom. Intel. Lab. Syst.},
  volume={38},
  number={2},
  pages={149--172},
  year={1997},
  publisher={Amsterdam; New York: Elsevier Science Pub. Co., 1986-}
}

@article{GiGl12,
  title={Accelerated multiplicative updates and hierarchical {ALS} algorithms for nonnegative matrix factorization},
  author={N. Gillis and F. Glineur},
  journal={Neural Comput.},
  volume={24},
  number={4},
  pages={1085--1105},
  year={2012},
  publisher={MIT Press}
}

@article{fMRI:MCIC,
  title={The {MCIC} collection: a shared repository of multi-modal, multi-site brain image data from a clinical investigation of schizophrenia},
  author={Gollub, R. L. and others},
  journal={Neuroinformatics},
  volume={11},
  number={3},
  pages={367--388},
  year={2013},
  publisher={Springer}
}

@inproceedings{BaWaTsWaDa17,
  title={Unsupervised network discovery for brain imaging data},
  author={Bai, Zilong and Walker, Peter and Tschiffely, Anna and Wang, Fei and Davidson, Ian},
  booktitle={KDD'17: Proc. 23rd ACM SIGKDD Int. Conf. Knowl. Discov. and Data Mining},
  pages={55--64},
  year={2017}
}

@article{BeKeGiDe20,
  title={Getting to the core of {PARAFAC2}, a nonnegative approach},
  author={Van Benthem, Mark H and Keller, Timothy J and Gillispie, Gregory D and DeJong, Stephanie A},
  journal={Chemom. Intel. Lab. Syst.},
  volume={206},
  pages={104127},
  year={2020},
  publisher={Elsevier}
}

@article{BrSi98,
  title={Least squares algorithms under unimodality and non-negativity constraints},
  author={Bro, Rasmus and Sidiropoulos, Nicholaos D},
  journal={J. Chemom.},
  volume={12},
  number={4},
  pages={223--247},
  year={1998},
  publisher={Wiley Online Library}
}

@article{St08,
  title={Unimodal regression via prefix isotonic regression},
  author={Stout, Quentin F},
  journal={Comput. Stat. Data Anal.},
  volume={53},
  number={2},
  pages={289--297},
  year={2008},
  publisher={Elsevier}
}

@book{Be17,
  title={First-order methods in optimization},
  author={Beck, Amir},
  year={2017},
  publisher={SIAM}
}

@article{MiYeSt20,
  title={Nesterov acceleration of alternating least squares for canonical tensor decomposition: Momentum step size selection and restart mechanisms},
  author={Mitchell, Drew and Ye, Nan and De Sterck, Hans},
  journal={Numer. Linear Algebra with Appl.},
  volume={27},
  number={4},
  pages={e2297},
  year={2020},
  publisher={Wiley Online Library}
}

@INPROCEEDINGS{AnCoKhGi20,
  author={Shun Ang, Andersen Man and Cohen, Jeremy E. and Khanh Hien, Le Thi and Gillis, Nicolas},
  booktitle={ICASSP'20: Proc. Int. Conf. Acoust., Speech, Signal Process.},
  title={Extrapolated Alternating Algorithms for Approximate Canonical Polyadic Decomposition}, 
  year={2020},
  volume={},
  number={},
  pages={3147-3151}
  }

@article{YuAuBr21,
  title={Accelerating {PARAFAC2} algorithms for non-negative complex tensor decomposition},
  author={Yu, Huiwen and Augustijn, Dillen and Bro, Rasmus},
  journal={Chemom. Intel. Lab. Syst.},
  volume={214},
  pages={104312},
  year={2021},
  publisher={Elsevier}
}

@article{WaZh12,
  title={Nonnegative matrix factorization: A comprehensive review},
  author={Wang, Yu-Xiong and Zhang, Yu-Jin},
  journal={IEEE Trans. Knowl. Data Eng.},
  volume={25},
  number={6},
  pages={1336--1353},
  year={2012},
  publisher={IEEE}
}

@article{FrHa08,
  title={Computing non-negative tensor factorizations},
  author={Friedlander, Michael P and Hatz, Kathrin},
  journal={Optim. Methods Softw.},
  volume={23},
  number={4},
  pages={631--647},
  year={2008},
  publisher={Taylor \& Francis}
}

@article{BrJo97,
  title={A fast non-negativity-constrained least squares algorithm},
  author={Bro, Rasmus and De Jong, Sijmen},
  journal={J. Chemom.},
  volume={11},
  number={5},
  pages={393--401},
  year={1997},
  publisher={Wiley Online Library}
}

@article{CaPrKr80,
  title={CANDELINC: A general approach to multidimensional analysis of many-way arrays with linear constraints on parameters},
  author={Carroll, J Douglas and Pruzansky, Sandra and Kruskal, Joseph B},
  journal={Psychometrika},
  volume={45},
  number={1},
  pages={3--24},
  year={1980},
  publisher={Springer}
}

@article{BrAn98,
  title={Improving the speed of multiway algorithms: Part {II}: Compression},
  author={Bro, Rasmus and Andersson, Claus A},
  journal={Chemom. Intel. Lab. Syst.},
  volume={42},
  number={1-2},
  pages={105--113},
  year={1998},
  publisher={Elsevier}
}

@article{Ki98,
  title={A three--step algorithm for {CANDECOMP/PARAFAC} analysis of large data sets with multicollinearity},
  author={Kiers, Henk A. L.},
  journal={J. Chemom.},
  volume={12},
  number={3},
  pages={155--171},
  year={1998},
  publisher={Wiley Online Library}
}

@inproceedings{RoScCoAc21,
  title={{PARAFAC2} {AO-ADMM}: Constraints in all modes},
  author={Roald, Marie and Schenker, Carla and Cohen, Jeremy E and Acar, Evrim},
  booktitle={EUSIPCO'21: Proc. 2021 29th Eur. Signal Process. Conf.},
  year={2021},
  organization={EURASIP}
}

@article{ZiKi02,
  title={Degenerate solutions obtained from several variants of factor analysis},
  author={Zijlstra, Bonne JH and Kiers, Henk AL},
  journal={J. Chemom.},
  volume={16},
  number={11},
  pages={596--605},
  year={2002},
  publisher={Wiley Online Library}
}

@article{BeKe04,
  title={Fast algorithm for the solution of large-scale non-negativity-constrained least squares problems},
  author={Van Benthem, Mark H and Keenan, Michael R},
  journal={J. Chemom.},
  volume={18},
  number={10},
  pages={441--450},
  year={2004},
  publisher={Wiley Online Library}
}

@article{KiPa11,
  title={Fast nonnegative matrix factorization: An active-set-like method and comparisons},
  author={Kim, Jingu and Park, Haesun},
  journal={SIAM J. Sci. Comput.},
  volume={33},
  number={6},
  pages={3261--3281},
  year={2011},
  publisher={SIAM}
}

@phdthesis{Br98,
  author       = {Bro, Rasmus}, 
  title        = {Multi-way analysis in the food industry},
  school       = {Royal Veterinary and Agricultural University Denmark},
  year         = 1998,
}

@article{HaLu96,
  title={Uniqueness proof for a family of models sharing features of Tucker's three-mode factor analysis and PARAFAC/CANDECOMP},
  author={Harshman, Richard A and Lundy, Margaret E},
  journal={Psychometrika},
  volume={61},
  number={1},
  pages={133--154},
  year={1996},
  publisher={Springer}
}

@article{TeJoKi96,
  title={Some uniqueness results for {PARAFAC2}},
  author={ten Berge, Jos MF and Kiers, Henk AL},
  journal={Psychometrika},
  volume={61},
  number={1},
  pages={123--132},
  year={1996},
  publisher={Springer}
}

@article{BrLeJo13,
  title={Solving the sign indeterminacy for multiway models},
  author={Bro, Rasmus and Leardi, Riccardo and Johnsen, Lea Gi{\o}rtz},
  journal={J. Chemom.},
  volume={27},
  number={3-4},
  pages={70--75},
  year={2013},
  publisher={Wiley Online Library}
}

@article{FrOt14,
  title={The number of singular vector tuples and uniqueness of best rank-one approximation of tensors},
  author={Friedland, Shmuel and Ottaviani, Giorgio},
  journal={Found. Comp. Math.},
  volume={14},
  number={6},
  pages={1209--1242},
  year={2014},
  publisher={Springer}
}

@inproceedings{ReLoXiHo20,
  title={Robust Irregular Tensor Factorization and Completion for Temporal Health Data Analysis},
  author={Ren, Yifei and Lou, Jian and Xiong, Li and Ho, Joyce C},
  booktitle={CIKM'20: Proc. 29th ACM Int. Conf. Inf. Knowl. Manag.},
  pages={1295--1304},
  year={2020}
}

@book{BoVa04,
  title={Convex optimization},
  author={Boyd, Stephen and Vandenberghe, Lieven},
  year={2004},
  publisher={Cambridge university press}
}

@book{Be99,
  title={Nonlinear Programming},
  author={Bertsekas, D.P.},
  isbn={9781886529007},
  series={Athena Sci. Optim. Comput. Series},
  year={1999},
  publisher={Athena Scientific}
}

@article{De08,
  title={Decompositions of a higher-order tensor in block terms—Part {II}: Definitions and uniqueness},
  author={De Lathauwer, Lieven},
  journal={SIAM J. Matrix Anal. Appl.},
  volume={30},
  number={3},
  pages={1033--1066},
  year={2008},
  publisher={SIAM}
}

@inproceedings{GuPaPa20,
  title={Beyond rank-1: Discovering rich community structure in multi-aspect graphs},
  author={Gujral, Ekta and Pasricha, Ravdeep and Papalexakis, Evangelos},
  booktitle={Proc. Web Conf. 2020},
  pages={452--462},
  year={2020}
}

@article{ChKoMoTh19,
  title={Blind {fMRI} source unmixing via higher-order tensor decompositions},
  author={Chatzichristos, Christos and Kofidis, Eleftherios and Morante, Manuel and Theodoridis, Sergios},
  journal={J. Neurosci. Methods},
  volume={315},
  pages={17--47},
  year={2019},
  publisher={Elsevier}
}

@InProceedings{sup.AfPePaSeHoSu18,
  author    = {A. Afshar and I. Perros and E. E. Papalexakis and E. Searles and J. Ho and J. Sun},
  booktitle = {ACM Int. Conf. on Inf. and Knowl. Management},
  title     = {{COPA}: {C}onstrained {PARAFAC2} for {S}parse \& {L}arge {D}atasets},
  year      = {2018},
  pages     = {793--802},
  numpages  = {10},
}

@Article{sup.KiTeBr99,
  author  = {H. A. L. Kiers and J. M. F. {Ten Berge} and R. Bro},
  journal = {J. Chemom.},
  title   = {{PARAFAC2} - {Part I. A} direct fitting algorithm for the {PARAFAC2} model},
  year    = {1999},
  number  = {3-4},
  pages   = {275--294},
  volume  = {13},
}

@Article{sup.HuSiLi16,
  author    = {K. Huang and N. D. Sidiropoulos and A. P. Liavas},
  journal   = {IEEE Trans. Signal Process.},
  title     = {A flexible and efficient algorithmic framework for constrained matrix and tensor factorization},
  year      = {2016},
  number    = {19},
  pages     = {5052--5065},
  volume    = {64},
  publisher = {IEEE},
}

@article{sup.BrSi98,
  title={Least squares algorithms under unimodality and non-negativity constraints},
  author={Bro, Rasmus and Sidiropoulos, Nicholaos D},
  journal={J. Chemom.},
  volume={12},
  number={4},
  pages={223--247},
  year={1998},
  publisher={Wiley Online Library}
}

@article{sup.St08,
  title={Unimodal regression via prefix isotonic regression},
  author={Stout, Quentin F},
  journal={Comput. Stat. \& Data Anal.},
  volume={53},
  number={2},
  pages={289--297},
  year={2008},
  publisher={Elsevier}
}

@article{sup.LiBr12,
  title={Lean algebraic multigrid (LAMG): Fast graph Laplacian linear solver},
  author={Livne, Oren E and Brandt, Achi},
  journal={SIAM J. Sci. Comput.},
  volume={34},
  number={4},
  pages={B499--B522},
  year={2012},
  publisher={SIAM}
}

@article{sup.Co13,
  title={A direct algorithm for 1-{D} total variation denoising},
  author={L. Condat},
  journal={IEEE Signal Process. Letters},
  volume={20},
  number={11},
  pages={1054--1057},
  year={2013},
  publisher={IEEE}
}
\end{document}

\maketitle
\section{Proofs} \label{sup.sec:proofs}
\begin{proof}[Proof of Proposition 3.1]
Let \(\mathscr{M} = \Ortho{J_1}{R} \times \ldots \times \Ortho{J_K}{R} \times \Real^{R\times R}  \). There exists a continuous surjection between \(\mathscr{M}\) and \(\PSet\) \cite{sup.KiTeBr99}, and it is, therefore, sufficient to prove the closedness of \(\mathscr{M}\) to prove Proposition 3.1. Closedness is preserved by cartesian products, and both \(\Real^{R \times R}\) and \( \Ortho{J_k}{R}\) are closed (closedness of \( \Ortho{J_k}{R} \) follows from the preimage description \(\Ortho{J_k}{R} = f^{-1}(\{\M{I}\})\), where \(f(\M{X}) = \M{X}\Tra\M{X}\)). Thus, \(\mathscr{M}\) is closed, and consequently \(\PSet\) is also closed.
\end{proof}

\begin{proof}[Proof of Theorem 4.1]
Recall that \(f: \Real^{n} \times \Real^{m} \to \Real\) satisfies \(f(a\V{u}, a^{-2}\V{v}) = f(\V{u}, \V{v})\), and that \(r_u : \Real^{n} \to \Real\) and \(r_v : \Real^{m} \to \Real\) are two (absolutely) homogeneous functions of degree \(d_u\) and \(d_v\) respectively. We want to show that
\begin{equation}
    \min_{\V{u}, \V{v}} f(\V{u}, \V{v}) + a r_u(\V{u}) + r_v(\V{v}) \label{sup.eq:reg.scaling.proof.1}
\end{equation}
and 
\begin{equation}
    \min_{\V{u}, \V{v}} f(\V{u}, \V{v}) + r_u(\V{u}) + a^{2 \frac{d_v}{d_u}} r_v(\V{v}) \label{sup.eq:reg.scaling.proof.2}
\end{equation}
are equivalent.

We start by introducing the change of variables \(\V{u} = a^{-\frac{1}{d_u}}\tilde{\V{u}}\) and \(\V{v} = a^{\frac{2}{d_u}}\tilde{\V{v}}\) into \cref{sup.eq:reg.scaling.proof.1}:
\begin{equation}
    \min_{\V{u}, \V{v}} f(a^{-\frac{1}{d_u}}\tilde{\V{u}}, a^{\frac{2}{d_u}}\tilde{\V{v}}) + a r_u(a^{-\frac{1}{d_u}}\tilde{\V{u}}) + r_v(a^{\frac{2}{d_u}}\tilde{\V{v}}). \label{sup.eq:reg.scaling.proof.3}
\end{equation}
Next, from \(f(a\V{u}, a^{-2}\V{v}) = f(\V{u}, \V{v})\) and the homogeneity of \(r_u\) and \(r_v\), we have
\begin{equation}
    \min_{\V{u}, \V{v}} f(\tilde{\V{u}}, \tilde{\V{v}}) + a \left(a^{-\frac{1}{d_u}}\right)^{d_u} r_u(\tilde{\V{u}}) + \left(a^{\frac{2}{d_u}}\right)^{d_v}r_v(\tilde{\V{v}}), \label{sup.eq:reg.scaling.proof.4}
\end{equation}
which is equivalent to
\begin{equation}
    \min_{\tilde{\V{u}}, \tilde{\V{v}}} f(\tilde{\V{u}}, a\tilde{\V{v}}) + r_u(\tilde{\V{u}}) + a^{2\frac{d_v}{d_u}}r_v(\tilde{\V{v}}). \label{sup.eq:reg.scaling.proof.5}
\end{equation}
\end{proof}
\section{Additional observations on \(\PSet\)}
\subsection{Computations for the one-component case}
In the one-component case, we have \(\Ortho{J}{1} = \{\V{x} \in \Real^{J} | \norm{\V{x}} = 1\}\) and \(\Bk\Tra\Bk = \delta^2 \in \Real\). Consequently, \(\delta \Ortho{J}{1}\) is a 2-norm sphere with radius \(\delta\). Solving
\begin{equation}
    \min_{\delta} \sum_{k=1}^{K} d(\delta\Ortho{J}{1}, \Bk) = \min_{\delta} \sum_{k=1}^{K} \min_{\Pk\in\Ortho{J}{1}} \| \Bk - \Pk\delta  \|_F^2 \label{sup.eq:rank.1.1}
\end{equation}
is therefore a bilevel optimization problem where the inner problem is the projection onto the sphere with radius \(\delta\), which has the closed form solution (for \(\Bk \neq 0\)):
\begin{equation}
    \argmin_{\Pk\in\Ortho{J}{1}} \| \Bk - \Pk\delta  \|_F^2 = \frac{[\V{b}_k]_1}{\norm{[\V{b}_k]_1}}. \label{sup.eq:rank.1.2}
\end{equation}
Substituting \cref{sup.eq:rank.1.2} into \cref{sup.eq:rank.1.1} yields
\begin{equation}
    \min_{\delta} \sum_{k=1}^{K} d(\delta\Ortho{J}{1}, \Bk) = \min_{\delta} \sum_{k=1}^{K} \fnorm{ \Bk - \frac{[\V{b}_k]_1}{\norm{[\V{b}_k]_1}}\delta  }^2 \label{sup.eq:rank.1.3},
\end{equation}
which is a formulation of the mean. For the one-component case, we therefore have that projection onto \(\PSet\) is equivalent to projecting onto the sphere with radius given by the average norm of each \(\Bk\).

\subsection{Additional interpretations of \(\PSet\)}
One interpretation of \(\PSet\) arises from the formulation discussed in Subsection 3.1. We can interpret each \(\Pk\) matrix as an orthogonal basis for a given subspace and \(\blueprint\) as a matrix of coordinates for each component in that subspace.

\begin{figure}
    \centering
    \begin{subfigure}{0.48\textwidth}
    \centering
    \includegraphics[width=0.8\textwidth]{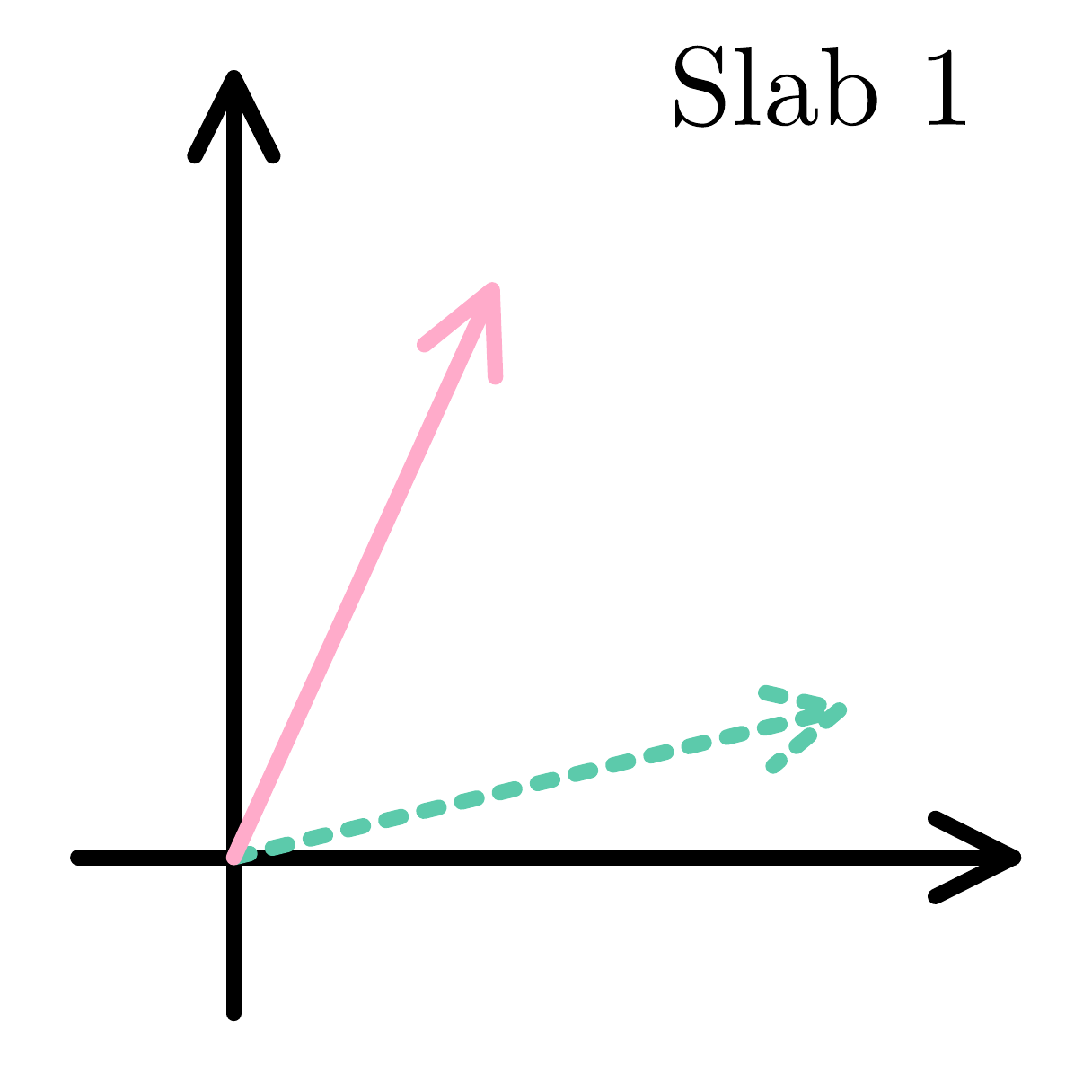}
    \end{subfigure}
    \begin{subfigure}{0.48\textwidth}
    \centering
    \includegraphics[width=0.8\textwidth]{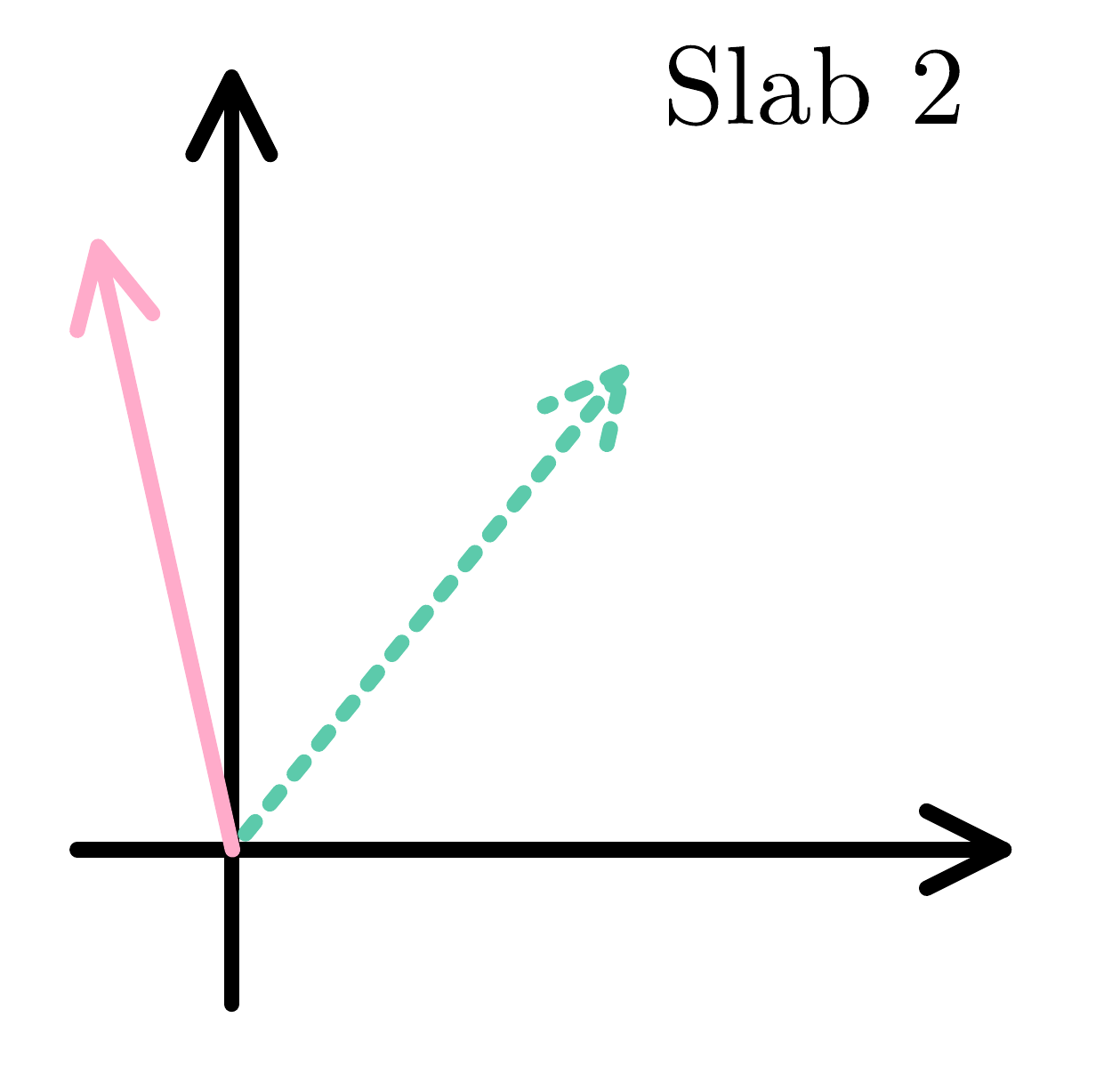}
    \end{subfigure}
    \begin{subfigure}{0.48\textwidth}
    \centering
    \includegraphics[width=0.8\textwidth]{figures/supplement/M145003_SUPPLEMENT_pset1.pdf}
    \end{subfigure}
    \begin{subfigure}{0.48\textwidth}
    \centering
    \includegraphics[width=0.8\textwidth]{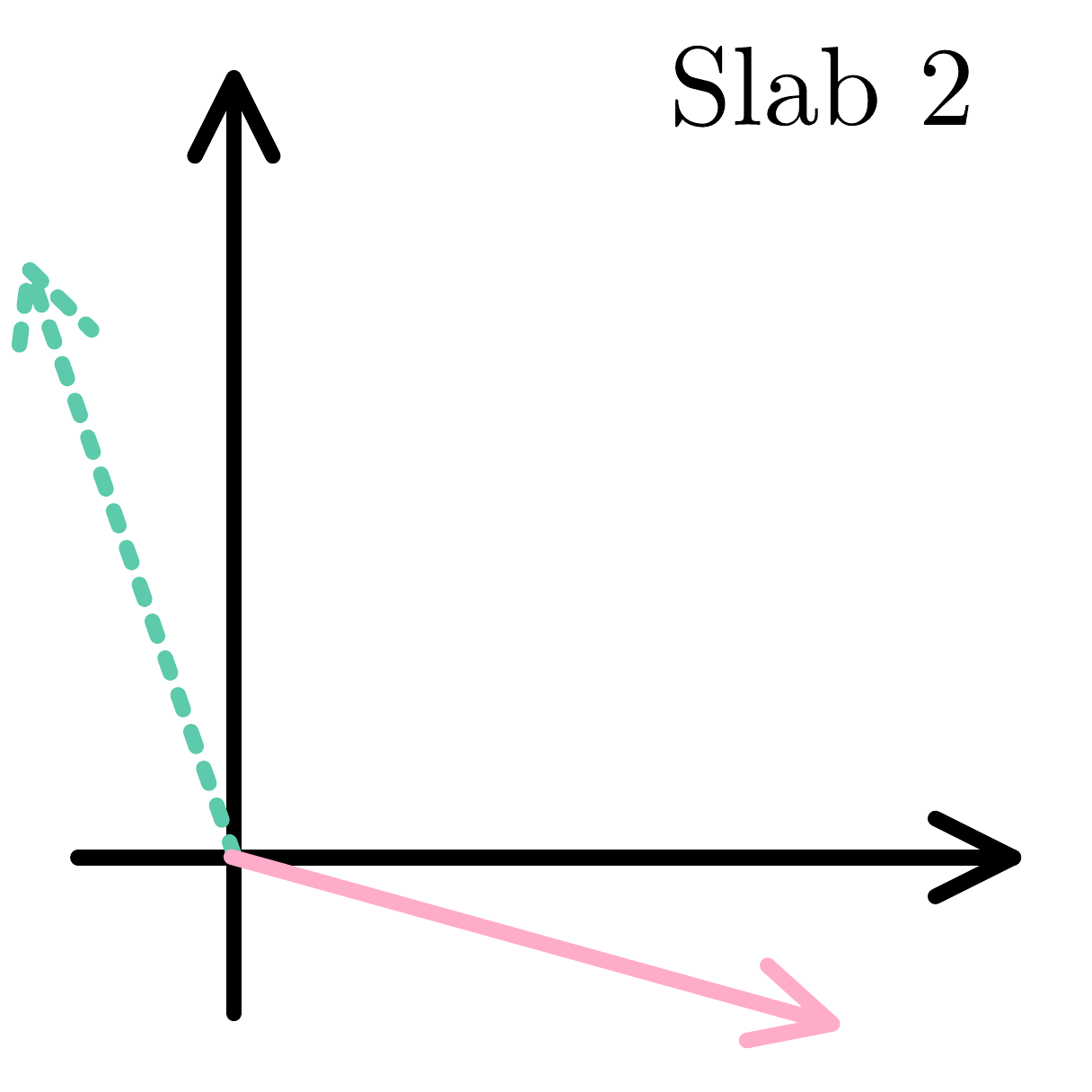}
    \end{subfigure}
    \includegraphics{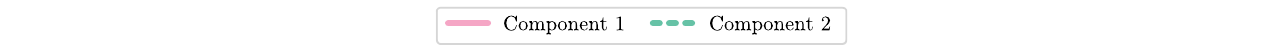}
    \caption{Illustration of the PARAFAC2 constraint with two components in two dimensions across two slabs. Each row represents one set of evolving component vectors and each column represents tensor slabs. The component vectors in the top row satisfy the PARAFAC2 constraint, since the angle between the components are constant for the first and second slab. The component vectors in the bottom row do not satisfy the PARAFAC2 constraint, since the angle between them are not constant for the two slabs.}
    \label{sup.fig:pset.twocomp}
\end{figure}

Another interpretation of \(\PSet\),  for the case where \(J_k = J\), is apparent by observing that
\begin{equation}
    \Bk = \Pk \blueprint = \Pk \M{P}_1\Tra \M{P}_1 \blueprint = \tilde{\M{P}}_k \B_1,
\end{equation}
where \(\tilde{\M{P}}_k = \Pk \M{P}_1\Tra\). Thus, every \(\Bk\) matrix can be constructed by rotating and reflecting the columns of \(\B_1\). See \cref{sup.fig:pset.twocomp} for an illustration of this with two slabs constructed from two components in \(\Real^2\).

{
\section{Some proximable constraints}\label{sup.sec:constraints}
\subsection{Hard constraints}
Imposing hard constraints, that is, restricting the component vectors to lie in a specific set, requires evaluating the projection onto that set. For some constraints, the projection has a fast closed-form solution (e.g., non-negativity). Other constraints are more involved and may not even be convex. For example, unimodality constraints require that we solve a set of isotonic regression problems for each component vector, and the proximal operator is not unique \cite{sup.BrSi98,sup.St08}. The PARAFAC2 constraint is another example of a non-convex hard constraint.

\subsection{Graph Laplacian regularization}
For graph Laplacian regularization, we penalize pairwise differences between component vector elements. With appropriately selected weight parameters \(w_{ij} \geq 0\), we get the graph Laplacian penalty
\begin{equation}
    g(\V{x}) = \sum_{i, j} w_{ij} (x_i - x_j)^2 = \V{x}\Tra \M{L} \V{x},
\end{equation}
where \(\M{L}\) is a positive semidefinite matrix with \(\ME{L}{ij} = -w_{ij}\) for \(i \neq j\) and \(\ME{L}{ii} = \sum_j w_{ij}\). The scaled proximal operator of this penalty function is given by
\begin{equation}
    \prox{\frac{\V{x}\Tra \M{L} \V{x}}{\rho}}{\V{x}} =  \left(\M{L} + \frac{\rho}{2}\M{I}\right)^{-1} \frac{\rho}{2} \V{x}, \label{sup.eq:laplace.prox}
\end{equation}
which can be evaluated using the Cholesky factorization or algebraic multigrid preconditioned conjugate gradient \cite{sup.LiBr12}.

\subsection{TV regularization}
Another popular regularization technique is TV regularization, which encourages piecewise constant components and is defined as
\begin{equation}
    g(\V{x}) = \sum_{i} |x_{i+1} - x_{i}|
\end{equation}
While there is no closed form solution for the proximal operator of the TV penalty, we can still evaluate it in linear time \cite{sup.Co13}.
}

\section{Including ridge regularization in the data fidelity updates}
Here, we state the update steps with ridge imposed for all modes. For the \(\A\)-updates, the update becomes
\begin{equation}
   \prox{\frac{\lossF{\A} + \gamma\fnorm{\cdot}^2}{\rho_\A}}{\M{M}} = \left( \sum_{k=1}^K \Xk \Mn{\Gamma}{k}  + \frac{\rho_\A}{2}\M{M} \right) \left( \sum_{k=1}^K \Mn{\Gamma}{k} \Tra \Mn{\Gamma}{k} + \frac{2\gamma + \rho_\A}{2} \M{I} \right)^{-1},
\end{equation}
if we impose ridge regularization with a strength of \(\gamma\). Likewise, the updates with respect to \(\Bk\) and \(\Dk\) become
\begin{equation}
    \Bk \gets \left(\Xk\Tra \A \Dk + \frac{\rho_{\Bk}}{2}\M{M}\right)\left( \Dk\A\Tra  \A \Dk + \left(\gamma + \rho_{\Bk}\right) \M{I} \right)^{-1},
\end{equation}
where \(\M{M} = (\BkAux - \BkDual + \BkConstraint - \blueprintkDual)\), and
\begin{equation}
\prox{\frac{\lossF{\Dk} + \gamma\fnorm{\cdot}^2}{\rho_{\Dk}}}{\V{v}} = \left( \A\Tra \A * \Bk\Tra\Bk + \frac{2\gamma + \rho_{\Dk}}{2} \M{I} \right)^{-1}  \left( \DiagEntries{\A\Tra \Xk \Bk} + \frac{\rho_{\Dk}}{2}\V{v} \right),
\end{equation}
respectively.

\section{CP-based updates for the non-evolving modes} \label{sup.sec:cp.updates}
To obtain the CP-based update steps, we use the reformulation in \cite{sup.KiTeBr99}. We start by assuming that the \(\Bk\) factors are feasible, which allows the substitution \(\Bk = \BkConstraint\). Setting \(\TFS{T}{k} = \Xk \Pk\), we see that updating \(\A\) is equivalent to solving 
\begin{align}
\min_\A \sum_{k=1}^K \fnorm{\A \Dk \blueprint \Tra - \TFS{T}{k}} + \reg{\A}{\A},
\end{align}
which is the \(\A\)-updates of a regularized CP-model fitted to \(\T{T}\) (i.e. the tensor with frontal slices given by \(\TFS{T}{k}\)). We can use a similar argument for \(\DAll\), yielding the update rules from \cite{sup.HuSiLi16,sup.AfPePaSeHoSu18}:
\begin{equation}
\prox{\frac{\lossF{\A}}{\rho_\A}}{\M{M}} = \left(\TM{T}{1}\left(\blueprint \Khat \C \right) + \frac{\rho_\A}{2}\M{M}\right)\left(\blueprint\Tra\blueprint * \C\Tra\C + \frac{\rho_\A}{2}\M{I}\right)^{-1} \label{sup.eq:A.cp.loss.update}
\end{equation}
and
\begin{equation}
\prox{\frac{\lossF{\C}}{\rho_\C}}{\M{M}} = \left(\TM{T}{3}\left(\A \Khat \blueprint \right) + \frac{\rho_\C}{2}\M{M}\right)\left(\A\Tra\A * \blueprint\Tra\blueprint + \frac{\rho_\C}{2}\M{I}\right)^{-1}, \label{sup.eq:C.cp.loss.update}
\end{equation} 
where \(\TM{T}{n}\) is the mode-\(n\) unfolding of \(\T{T}\).

For both the CP- and the CMF-based updates, we get the same overall ADMM algorithm for \(\A\); the only difference is which linear system we need to solve. However, for the updates of the \(\Dk\)-matrices, we see clear differences. For the CMF-based updates, we solve a separate system of equations for each \(\Dk\)-matrix (i.e. each row of \(\C\)), while for the CP-based updates, all \(\Dk\)-matrices are obtained by solving the same system of equations.

To select the penalty parameters with the CP-based updates, we used the following heuristic:
\begin{align}
    \rho_\A &= \frac{1}{R}\Trace{\blueprint\Tra \blueprint * \C\Tra \C}, 
    & \rho_\C &= \frac{1}{R}\Trace{\A \Tra \A * \blueprint\Tra \blueprint}.
\end{align}

\begin{algorithm2e}
\small
\SetAlgoLined
\DontPrintSemicolon
\KwResult{\(\A, \AAux, \ADual\)}
 \While{stopping conditions are not met and max no. iterations not exceeded}{
    \( \A \xleftarrow{\cref{sup.eq:A.cp.loss.update}} \prox{\frac{\lossF{\A}}{\rho_\A}}{\AAux - \ADual}\) \;
    \( \AAux \gets \prox{\frac{ \regF{\A} }{ \rho_\A }}{\A + \ADual} \) \;
    \( \ADual \gets \ADual + \A - \AAux \) \;
 }
 \caption{CMF-based and CP-based ADMM for the \(\A\)-matrices}\label{sup.alg:admm.cp.a}
\end{algorithm2e}

\begin{algorithm2e}
\small
\SetAlgoLined
\DontPrintSemicolon
\KwResult{\(\C, \CAux, \CDual\)}
 \While{stopping conditions are not met and max no. iterations not exceeded}{
        \( \C \xleftarrow{\cref{sup.eq:C.cp.loss.update}} \prox{\frac{\lossF{\C}}{\rho_{\C}}}{\CAux - \CDual}\) \;
        \( \CAux \gets \prox{\frac{ \regF{\M{C}} }{ \rho_{\C} }}{\C + \CDual} \) \;
        \( \CDual \gets \CDual + \C - \CAux \) \;
 }
 \caption{CP-based ADMM for the \(\Dk\)-matrices (\(\C\)-matrix)}\label{sup.alg:admm.cp.d}
\end{algorithm2e}

\subsection{Computational complexity}
One difference between the CP-based and CMF-based update steps is their computational complexity, which are given in \cref{sup.tab:complexities}. 

\begin{table}
\centering
\caption{Computational complexities for the different AO-ADMM update steps. \(I\), \(J\) and \(K\) denote the tensor size, \(R\) denotes the number of components and \(Q\) the number of iterations.} \label{sup.tab:complexities}
\begin{threeparttable}
\begin{tabular}{@{}lll@{}}
\toprule
          & CMF-based updates                          & CP-based updates                                      \\
          \midrule
\(\A\)    & \(O(IJKR + JKR^2 + R^3 + IR^2Q)\)          & \(O(IKR^2 + R^3 + IR^2Q)\)\tnote{*}  \\
\(\DAll\) & \(O(IJKR + IR^2 + JKR^2 + KR^3 + KR^2Q)\) & \( O(IKR^2 + R^3 + KR^2Q)\)\tnote{*} \\
\(\BAll\) & \(O(IJKR + IR^2 + KR^3 + JKR^2Q)\)         & ---                        \\
\bottomrule
\end{tabular}
\begin{tablenotes}
\item[*]The CP-based update steps also require computing the \(\T{T}\)-tensor which adds an additional \(O(IJKR)\) complexity step shared between the A and D updates.
\end{tablenotes}
\end{threeparttable}
\end{table}

From the complexities alone, it seems like the CP-based updates have lower computational complexity. However, for the CP-based updates, we also need to compute the \(\T{T}\) tensor, by multiplying the frontal slices of \(\X\) with the \(\Pk\)-matrices, which incurs an additional \(O(IJKR)\) step in the algorithm. Also, the \(\Bk\) updates have a substantial time-complexity that may dominate the \(\A\) and \(\Dk\) updates. Finally, with the CMF-based updates, we update the \(\A\)- and \(\Dk\)-matrices based on the \(\Bk\)-matrices. However, for the CP-based updates, we update these matrices based on the \(\BkConstraint\)-matrices. Thus, the two algorithms may behave differently, and it is difficult to select the ``optimal'' update steps based on computational complexity alone.

\subsection{Setup SM1: Comparing the CMF- and CP-based ADMM updates}\label{sup.sec:updatecomparison.experiment}
\paragraph{Dataset generation}
To compare the two ADMM update schemes for \(\A\) and \(\DAll\), we generated non-negative \(\Bk\)-factor matrices by first creating a non-negative \(\blueprint\) and non-negative \(\PAll\) and then setting \(\Bk = \Pk \blueprint\). The elements of \(\blueprint\) were drawn from a uniform distribution between 0 and 1. Each \(\Pk\)-matrix was generated by first dividing the indices between 1 and \(J\) into \(R\) contiguous partitions, each containing the non-zero elements of one column of \(\Pk\), and then drawing these non-zero elements from a uniform distribution between 0 and 1. The partition indices were generated from an \(R\)-category Dirichlet distribution (with each concentration parameter set to 1) scaled by \(J\), rounded to the nearest integer. We used this approach to generate 30 five-component PARAFAC2 decompositions of various sizes. Ten of size \(30 \times 10 \times 70\), ten of size  \(30 \times 100 \times 70\) and ten of size \(30 \times 1000 \times 70\). For each of these decompositions, we generated two simulated data tensors with different noise levels; one with \(\eta = 0.33\) and one with \(\eta = 0.5\).

\paragraph{Experiment setup}
Each data tensor was decomposed with ten random initialization using both the CMF-based updates and the CP-based updates. Each initialization ran until convergence or for at most 2000 outer iterations. With both schemes, we imposed non-negativity on all modes.

\begin{figure}
    \centering
    \begin{subfigure}{\textwidth}
    \includegraphics{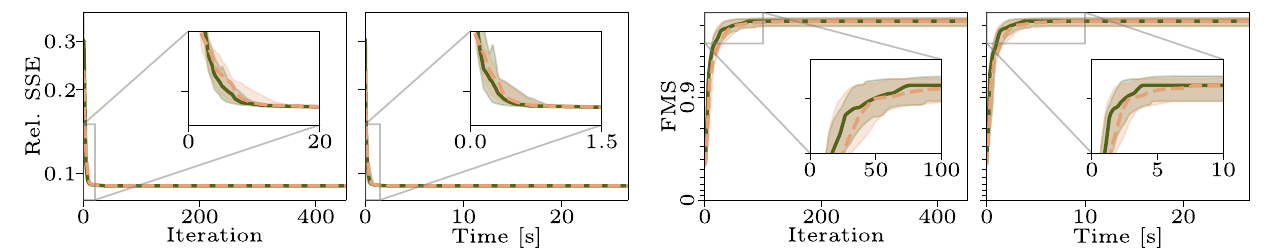}
    \caption{}
    \end{subfigure}
    \begin{subfigure}{\textwidth}
    \includegraphics{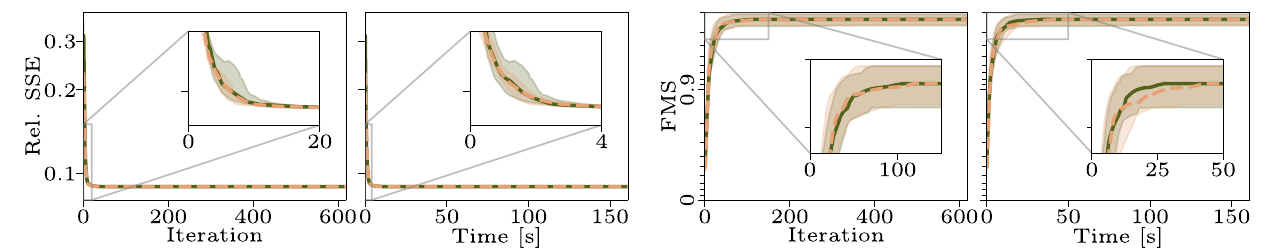}
    \caption{}
    \end{subfigure}
    \begin{subfigure}{\textwidth}
    \includegraphics{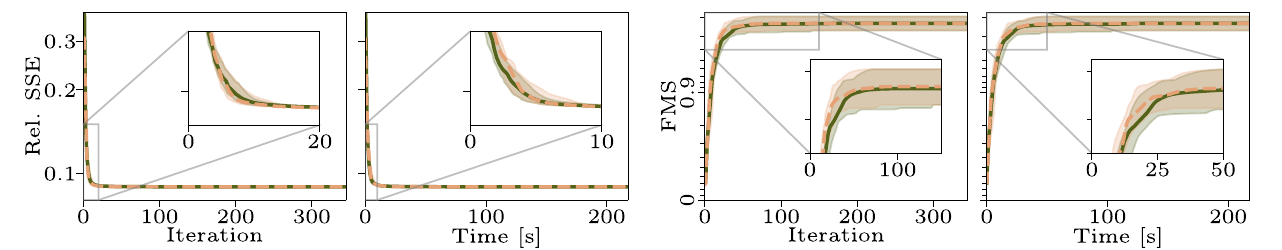}
    \caption{}
    \end{subfigure}
    \begin{subfigure}{\textwidth}
    \includegraphics{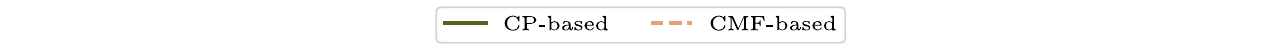}
    \end{subfigure}
    \caption{Setup SM1: Performance plots on the data tensors with \(\eta=0.33\) for the CMF-based and CP-based AO-ADMM schemes. J=10, 100, and 1000 in (a), (b) and (c), respectively}
    \label{sup.fig:update.comparison.noise0.33}
\end{figure}

\begin{figure}
    \centering
    \begin{subfigure}{\textwidth}
    \includegraphics{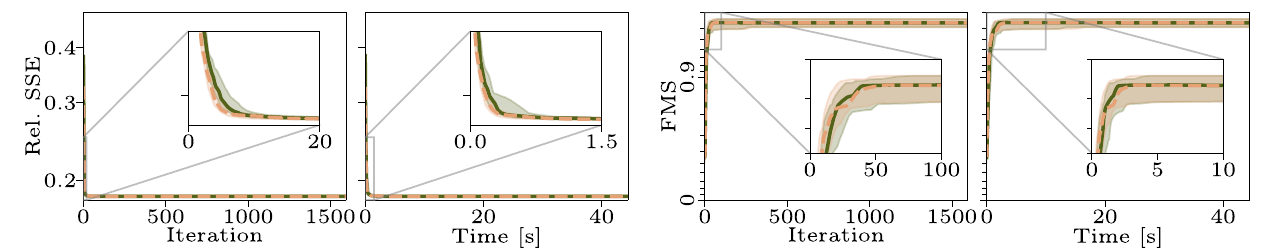}
    \caption{}
    \end{subfigure}
    \begin{subfigure}{\textwidth}
    \includegraphics{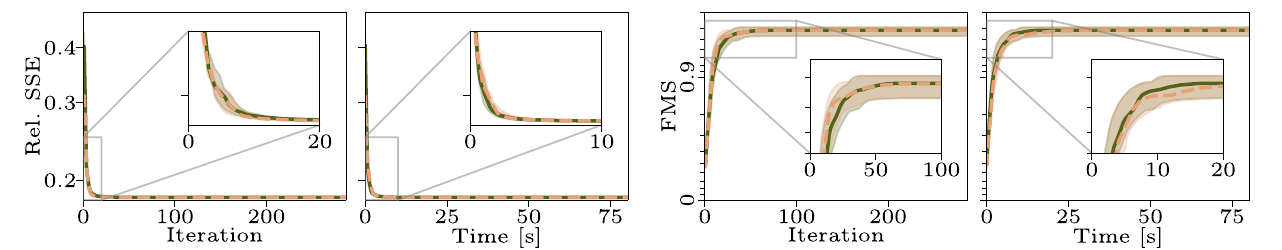}
    \caption{}
    \end{subfigure}
    \begin{subfigure}{\textwidth}
    \includegraphics{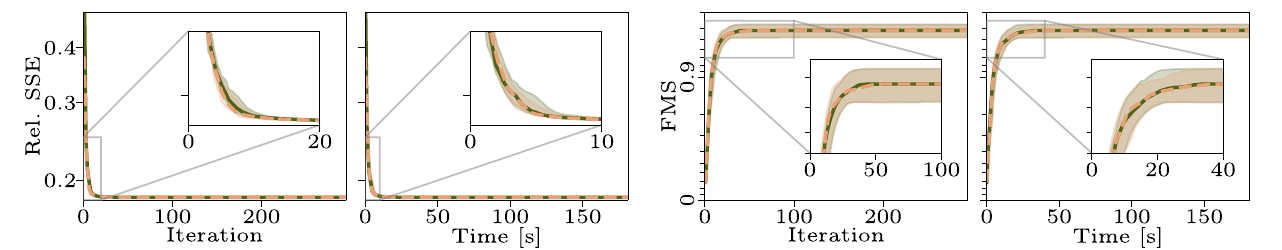}
    \caption{}
    \end{subfigure}
    \begin{subfigure}{\textwidth}
    \includegraphics{figures/supplement/M145003_SUPPLEMENT_setupSM1_legend.pdf}
    \end{subfigure}
    \caption{Setup SM1: Performance plots on the data tensors with \(\eta=0.5\) for the CMF-based and CP-based AO-ADMM schemes. J=10, 100, and 1000 in (a), (b) and (c), respectively}
    \label{sup.fig:update.comparison.noise0.5}
\end{figure}

\paragraph{Results}
\Cref{sup.tab:update.comparison.selected.inits,sup.tab:update.comparison.all.inits} and \cref{sup.fig:update.comparison.noise0.33,sup.fig:update.comparison.noise0.5} demonstrate that the two AO-ADMM schemes are comparable in terms of time and FMS.

\begin{table}[]
    \centering
    \caption{Mean performance (\(\pm\) one standard deviation) for the selected initialization for all simulated datasets with the CMF-based and CP-based AO-ADMM scheme.}
    \begin{tabular}{@{}r@{\hspace{1em}}rl@{\hspace{0.5em}}ll@{\hspace{0.5em}}ll@{\hspace{0.5em}}l@{}}
    \toprule
    &  & \multicolumn{2}{c}{FMS} & \multicolumn{2}{c}{Final iteration} & \multicolumn{2}{c}{Time [s]} \\
      \cmidrule(lr){3-4}\cmidrule(lr){5-6}\cmidrule(l){7-8}
   \(\eta\) & \(J\) &             CMF &              CP &              CMF &               CP &             CMF &               CP \\
      \cmidrule(lr){3-3} \cmidrule(r){4-4} \cmidrule(lr){5-5} \cmidrule(r){6-6} \cmidrule(lr){7-7} \cmidrule(){8-8}
    0.33 & \(10^1\)   & \(0.98 \pm 0\) &  \(0.98 \pm 0\) &  \(195 \pm 77\) &   \(192 \pm 64\) &   \(11 \pm 5\) &    \(10 \pm 3\) \\
         & \(10^2\)  & \(0.98 \pm 0\) &  \(0.98 \pm 0\) &  \(179 \pm 81\) &  \(188 \pm 116\) &  \(50 \pm 24\) &   \(46 \pm 30\) \\
         & \(10^3\) &  \(0.98 \pm 0\) &  \(0.98 \pm 0\) &  \(144 \pm 65\) &   \(160 \pm 71\) &  \(98 \pm 44\) &  \(104 \pm 46\) \vspace{0.5em}\\
     
    0.5 & \(10^1\)   &  \(0.96 \pm 0.01\) &  \(0.96 \pm 0.01\) &   \(301 \pm 289\) &  \(320 \pm 389\) &   \(14 \pm 9\) &  \(13 \pm 11\) \\
        & \(10^2\)  &  \(0.96 \pm 0.01\) &  \(0.96 \pm 0.01\) &   \(150 \pm 56\) &   \(151 \pm 37\) &  \(41 \pm 16\) &   \(33 \pm 8\) \\
        & \(10^3\) &  \(0.96 \pm 0.01\) &  \(0.96 \pm 0.01\) &   \(141 \pm 56\) &   \(141 \pm 52\) &  \(93 \pm 35\) &  \(86 \pm 32\) \\
    \bottomrule
    \end{tabular}
    \label{sup.tab:update.comparison.selected.inits}
\end{table}


\begin{table}[]
    \centering
    \caption{Setup SM1: Mean performance (\(\pm\) one standard deviation) across all initializations and all simulated data tensors with the CMF-based and CP-based AO-ADMM scheme.}
    \begin{tabular}{@{}r@{\hspace{1em}}rl@{\hspace{0.5em}}ll@{\hspace{0.5em}}ll@{\hspace{0.5em}}l@{}}
    \toprule
    &  & \multicolumn{2}{c}{FMS} & \multicolumn{2}{c}{Final iteration} & \multicolumn{2}{c}{Time [s]} \\
      \cmidrule(lr){3-4}\cmidrule(lr){5-6}\cmidrule(l){7-8}
   \(\eta\) & \(J\) &             CMF &              CP &              CMF &               CP &             CMF &               CP \\
      \cmidrule(lr){3-3} \cmidrule(r){4-4} \cmidrule(lr){5-5} \cmidrule(r){6-6} \cmidrule(lr){7-7} \cmidrule(){8-8}
    0.33 & \(10^1\)   &  \(0.98 \pm 0\) &  \(0.98 \pm 0\) &  \(261 \pm 302\) &  \(342 \pm 461\) &   \(15 \pm 18\) &    \(18 \pm 24\) \\
         & \(10^2\)  &  \(0.98 \pm 0\) &  \(0.98 \pm 0\) &  \(179 \pm 142\) &  \(227 \pm 280\) &   \(51 \pm 46\) &    \(59 \pm 83\) \\
         & \(10^3\) &  \(0.98 \pm 0\) &  \(0.98 \pm 0\) &   \(152 \pm 65\) &  \(167 \pm 149\) &  \(105 \pm 43\) &  \(109 \pm 106\) \vspace{0.5em}\\
     
    0.5 & \(10^1\)   &  \(0.96 \pm 0.01\) &  \(0.96 \pm 0.01\) &  \(525 \pm 647\) &  \(699 \pm 777\) &  \(27 \pm 35\) &  \(33 \pm 39\) \\
        & \(10^2\)  &  \(0.96 \pm 0.01\) &  \(0.96 \pm 0.01\) &   \(156 \pm 88\) &  \(182 \pm 185\) &  \(43 \pm 29\) &  \(47 \pm 60\) \\
        & \(10^3\) &  \(0.96 \pm 0.01\) &  \(0.96 \pm 0.01\) &   \(140 \pm 60\) &   \(142 \pm 62\) &  \(93 \pm 41\) &  \(87 \pm 38\) \\
    \bottomrule
    \end{tabular}
    \label{sup.tab:update.comparison.all.inits}
\end{table}

\section{Additional details for the simulation experiments}
\subsection{Setup 2}\label{sup.sec:setup2}
To generate \(\Bk\)-factor matrices, solved the optimization problem
\begin{equation}
\begin{aligned}
\min_{\Bk} \quad & ||\Bk \Tra \Bk - \M{X}\Tra\M{X}||^2 \\
\st \quad & [\Bk]_{jr} \geq 0
\end{aligned}
\end{equation}
using projected gradient descent with different standard normal random initializations for each \(\Bk\)-factor matrix and a fixed \(\M{X}\) matrix with elements from a truncated standard normal distribution. Then, if projected gradient descent did not obtain a loss-value less than \(10^{-12}\) within 10000 iterations, we re-initialized \(\Bk\), reduced the step-size by a factor 10 and tried again. If no suitable \(\Bk\) factor matrix could be found after 10 attempts, we selected a new \(\M{X}\)-matrix and restarted the process.

\begin{figure}
    \centering
    \begin{subfigure}{\textwidth}
        \includegraphics{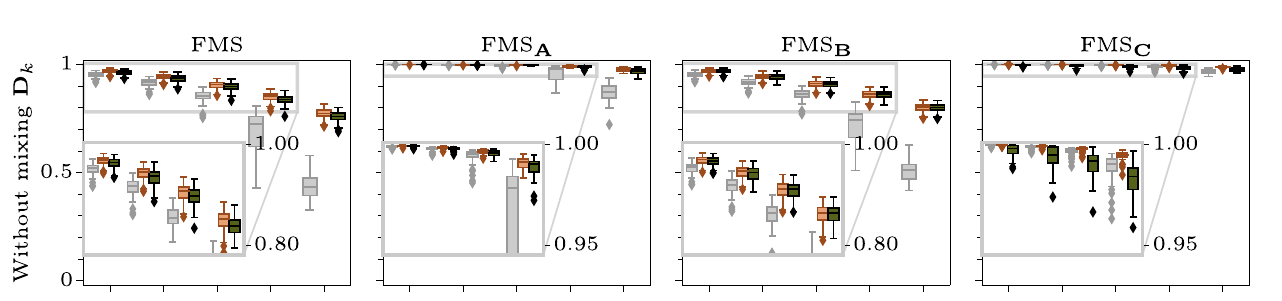}
    \end{subfigure}
    \begin{subfigure}{\textwidth}
        \includegraphics{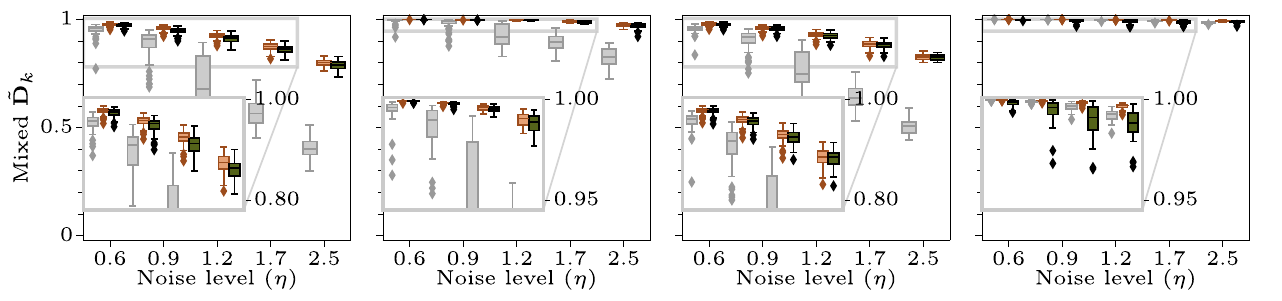}
    \end{subfigure}
    \begin{subfigure}{\textwidth}
        \includegraphics{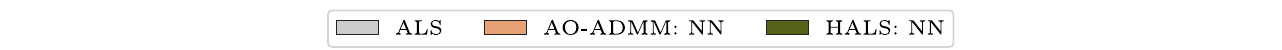}
    \end{subfigure}
    \caption{Setup 2: Boxplots showing the FMS for the different datasets.}
    \label{sup.fig:sim.nn.boxplot.remaining}
\end{figure}

\Cref{sup.fig:sim.nn.boxplot.remaining} shows the results for noise levels \(\eta = 0.6\), \(0.85\), \(1.2\), \(1.7\) and \(2.5\), and \cref{sup.tab:sim.nn.num.valid} gives an overview of the number of datasets left after removing datasets where ALS gave only degenerate solutions. 

\begin{table}[h]
    \centering
    \caption{Setup 2: Number of datasets where ALS gave at least one non-degenerate solution (of a total of 50 datasets)}
    \begin{tabular}{lrrrrrrrrrr}
\toprule
Noise level &  0.5 &  0.6 &  0.7 &  0.9 &  1.0 &  1.2 &  1.5 &  1.7 &  2.1 &  2.5 \\
Data &            &           &           &           &           &           &           &           &           &           \\
\midrule
No mixing &        50 &        50 &        50 &        50 &        50 &        50 &        50 &        49 &        49 &        39 \\
Mixing C  &        50 &        50 &        50 &        50 &        50 &        50 &        49 &        39 &        32 &        23 \\
\bottomrule
\end{tabular}
    \label{sup.tab:sim.nn.num.valid}
\end{table}

\begin{figure}
    \centering
    \begin{subfigure}{0.48\textwidth}
        \includegraphics{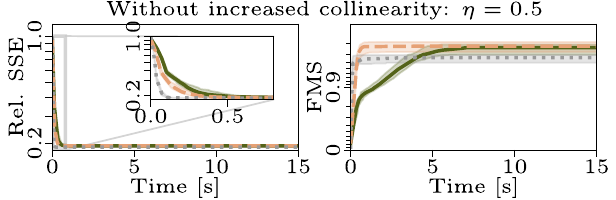}
    \end{subfigure}
    \begin{subfigure}{0.48\textwidth}
        \includegraphics{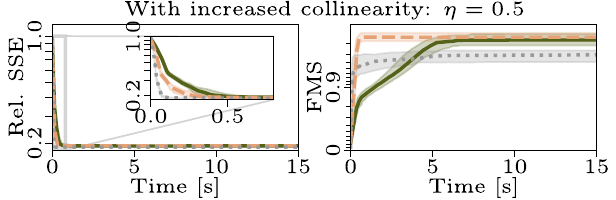}
    \end{subfigure}
    \begin{subfigure}{0.48\textwidth}
        \includegraphics{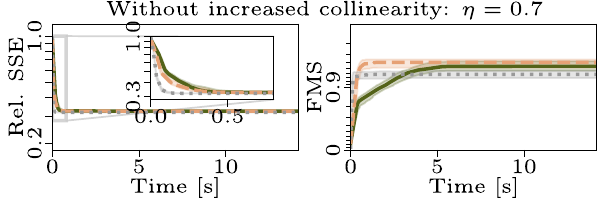}
    \end{subfigure}
    \begin{subfigure}{0.48\textwidth}
        \includegraphics{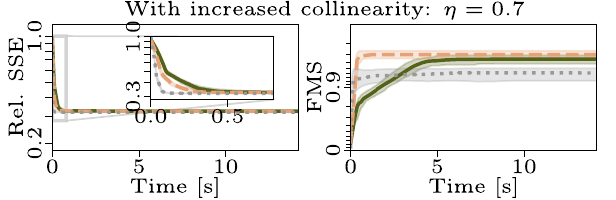}
    \end{subfigure}
    \begin{subfigure}{0.48\textwidth}
        \includegraphics{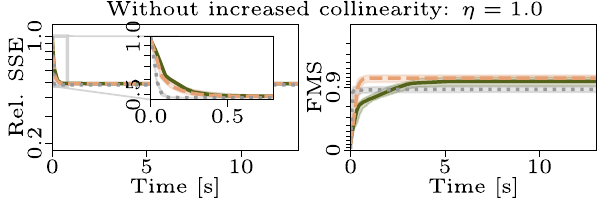}
    \end{subfigure}
    \begin{subfigure}{0.48\textwidth}
        \includegraphics{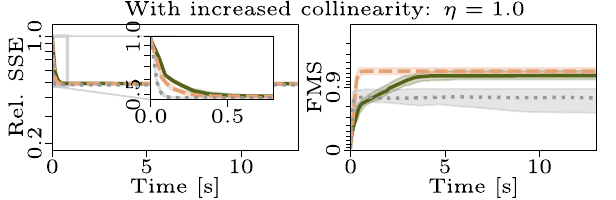}
    \end{subfigure}
    \begin{subfigure}{0.48\textwidth}
        \includegraphics{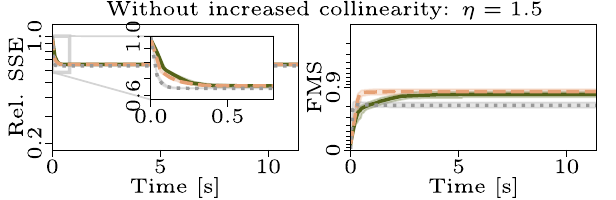}
    \end{subfigure}
    \begin{subfigure}{0.48\textwidth}
        \includegraphics{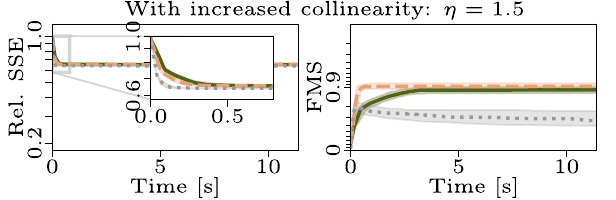}
    \end{subfigure}
    \begin{subfigure}{0.48\textwidth}
        \includegraphics{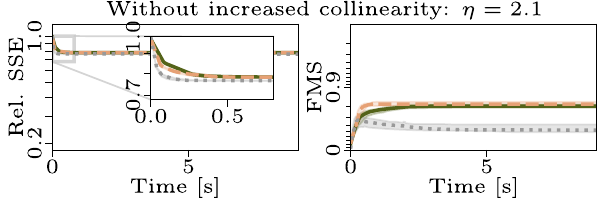}
    \end{subfigure}
    \begin{subfigure}{0.48\textwidth}
        \includegraphics{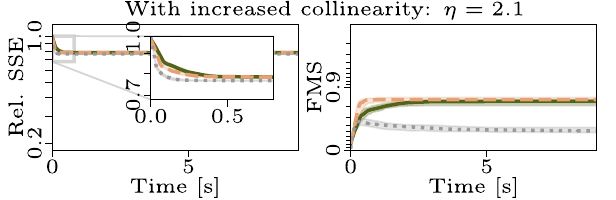}
    \end{subfigure}
    \begin{subfigure}{\textwidth}
    \includegraphics{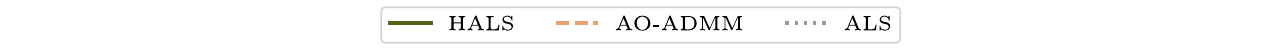}
    \end{subfigure}
    \caption{Setup 2: Performance plots showing FMS and relative SSE plotted against time. Each row corresponds to a different noise level.}
    \label{sup.fig:setup2.time.performance}
\end{figure}
    
\begin{figure}
    \centering
    \begin{subfigure}{0.48\textwidth}
        \includegraphics{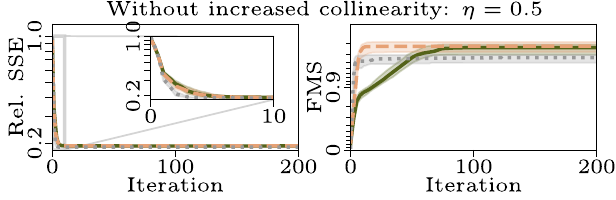}
    \end{subfigure}
    \begin{subfigure}{0.48\textwidth}
        \includegraphics{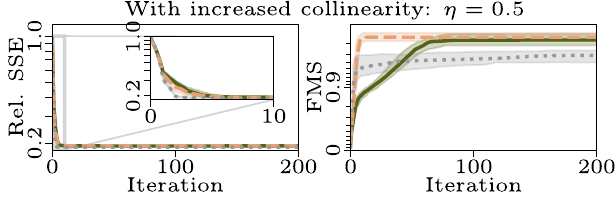}
    \end{subfigure}
    \begin{subfigure}{0.48\textwidth}
        \includegraphics{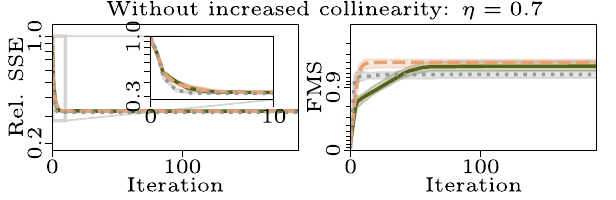}
    \end{subfigure}
    \begin{subfigure}{0.48\textwidth}
        \includegraphics{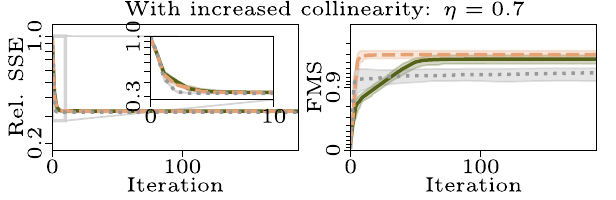}
    \end{subfigure}
    \begin{subfigure}{0.48\textwidth}
        \includegraphics{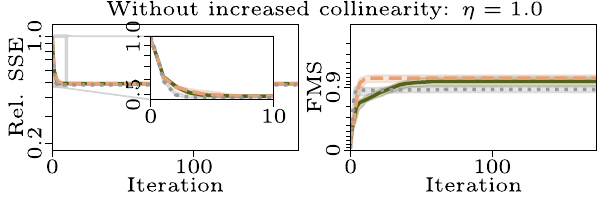}
    \end{subfigure}
    \begin{subfigure}{0.48\textwidth}
        \includegraphics{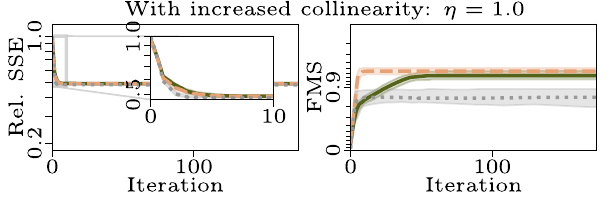}
    \end{subfigure}
    \begin{subfigure}{0.48\textwidth}
        \includegraphics{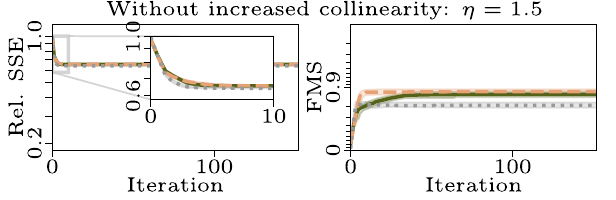}
    \end{subfigure}
    \begin{subfigure}{0.48\textwidth}
        \includegraphics{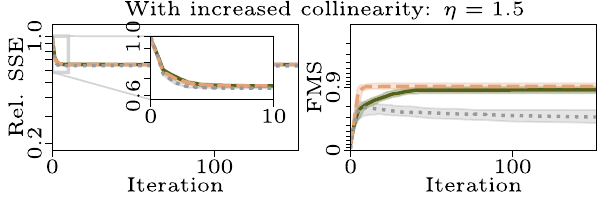}
    \end{subfigure}
    \begin{subfigure}{0.48\textwidth}
        \includegraphics{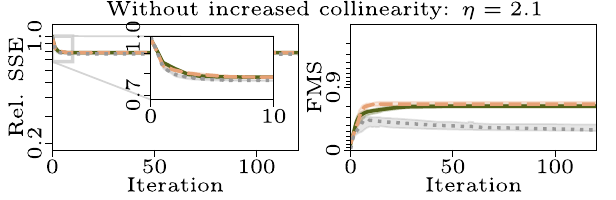}
    \end{subfigure}
    \begin{subfigure}{0.48\textwidth}
        \includegraphics{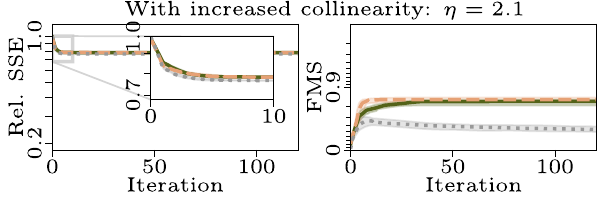}
    \end{subfigure}
    \begin{subfigure}{\textwidth}
    \includegraphics{figures/supplement/M145003_SUPPLEMENT_setup2_timing_legend.pdf}
    \end{subfigure}
    \caption{Setup 2: Performance plots showing FMS and relative SSE plotted against iteration. Each row corresponds to a different noise level.}
    \label{sup.fig:setup2.iteration.performance}
\end{figure}
\subsection{Setup 3}
\begin{figure}
    \centering
    \begin{subfigure}{0.48\textwidth}
        \includegraphics{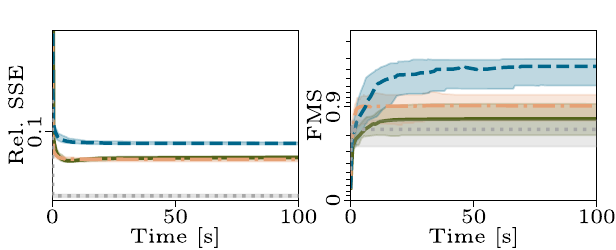}
    \end{subfigure}
    \begin{subfigure}{0.48\textwidth}
        \includegraphics{figures/supplement/M145003_SUPPLEMENT_setup3_timing_time.pdf}
    \end{subfigure}
    \begin{subfigure}{\textwidth}
    \includegraphics{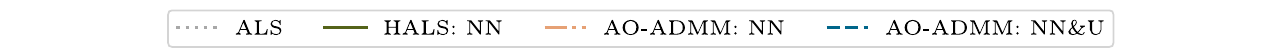}
    \end{subfigure}
    \caption{Setup 3: Performance plots showing FMS and relative SSE plotted against time and iteration.}
    \label{sup.fig:sim.unimodal.timing}
\end{figure}

{In \cref{sup.fig:sim.unimodal.timing}, we see that the models fitted with AO-ADMM with only non-negativity constraints converged faster than those fitted with the HALS scheme. Furthermore, we observe that while the models fitted with unimodality constraint took longer to converge than the other models, they obtained a much higher FMS.}
\subsection{Setup 3b}\label{sup.sec:setup3b}
\begin{figure}
    \centering
    \includegraphics{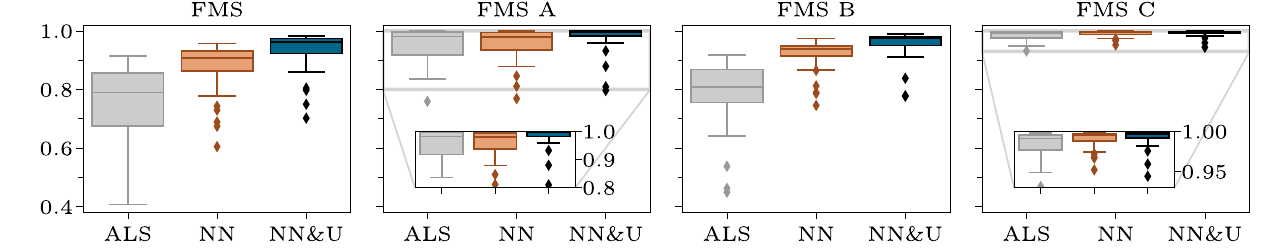}
    
    \caption{Setup 3: Boxplots showing the FMS for different models fitted to datasets with unimodal \(\Bk\)-matrices. NN represents models fitted with non-negativity imposed on all modes using AO-ADMM, NN\&U represents models fitted with non-negativity on all modes and unimodality imposed on \(\BAll\) using AO-ADMM, and ALS represents models fitted with non-negativity imposed on \(\A\) and \(\DAll\).}
    \label{sup.fig:sim.unimodal.boxplot}
\end{figure}

\begin{figure}
    \centering
    \begin{subfigure}{\textwidth}
        \includegraphics{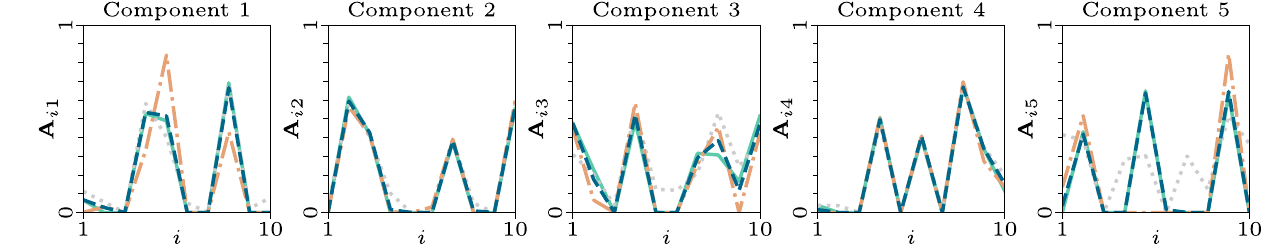}
    \end{subfigure}
    \begin{subfigure}{\textwidth}
        \includegraphics{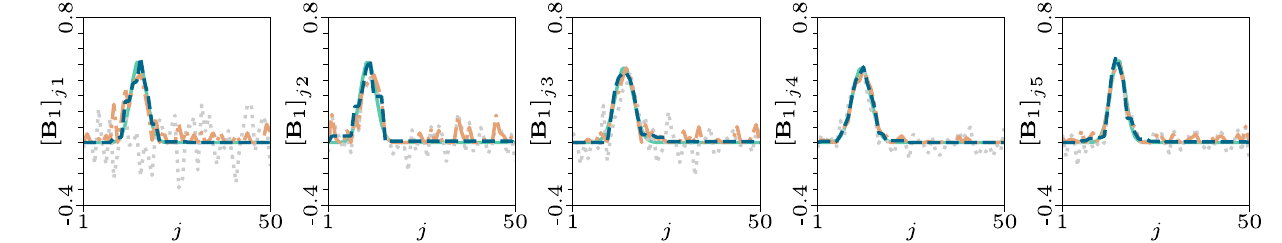}
    \end{subfigure}
    \begin{subfigure}{\textwidth}
        \includegraphics{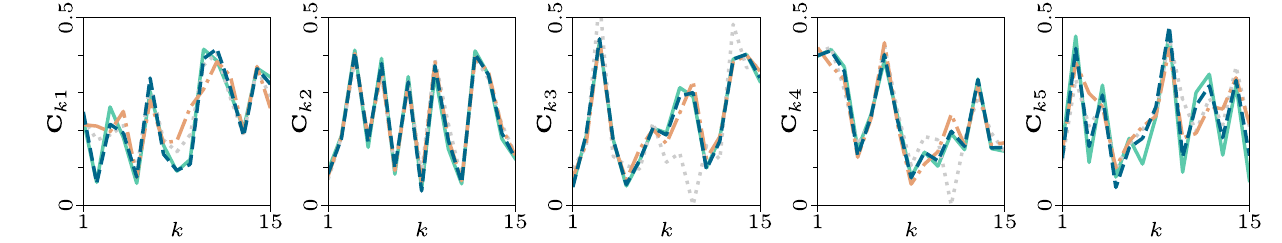}
    \end{subfigure}
    \begin{subfigure}{\textwidth}
        \includegraphics{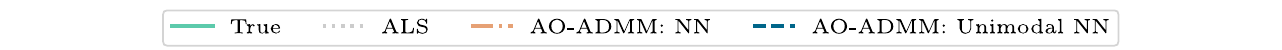}
    \end{subfigure}
    
    \caption{Setup 3: Plots showing the true and estimated components for one of the datasets. NN represents the model fitted with non-negativity imposed on all modes, Unimodal NN represents the model fitted with non-negativity on all modes and unimodality imposed on \(\BAll\) and ALS represents the model fitted with non-negativity imposed on \(\A\) and \(\DAll\). The cwSNRs (left to right) are \(-2.4, -9.3, -12, 4.5\) and \(-11\)~dB for this frontal slice.}
    \label{sup.fig:sim.unimodal.factors}
\end{figure}

{
\paragraph{Data generation}
We also ran experiments using the same data generation setup as for Setup 3, but with a constant \(\sigma_{kr}= \sigma_r\) for all \(k\), which means that the only change is a constant shifting for all components. Thus, the \(\Bk\)-components followed the PARAFAC2 constraint.

\paragraph{Experiment settings}
We used the same experimental settings as for Setup 3 to decompose these datasets (including the initialization scheme and the scaling of \(\rho_{\Bk}\) and the extra ADMM iterations).

\paragraph{Results}
In \cref{sup.fig:sim.unimodal.boxplot}, we see the results from these experiments. For six datasets, ALS yielded degenerate solutions for all 20 initializations, leaving 44 out of 50 datasets. We see from the figures that the results are similar to those from Setup 3. Specifically, we see that imposing constraints on the evolving mode improves factor recovery, especially when the component vectors are constrained to be unimodal. \Cref{sup.fig:sim.unimodal.factors} shows the true and estimated components for one of the datasets.
}

\subsection{Setup 4 and 5}
\begin{figure}
    \centering
    \includegraphics{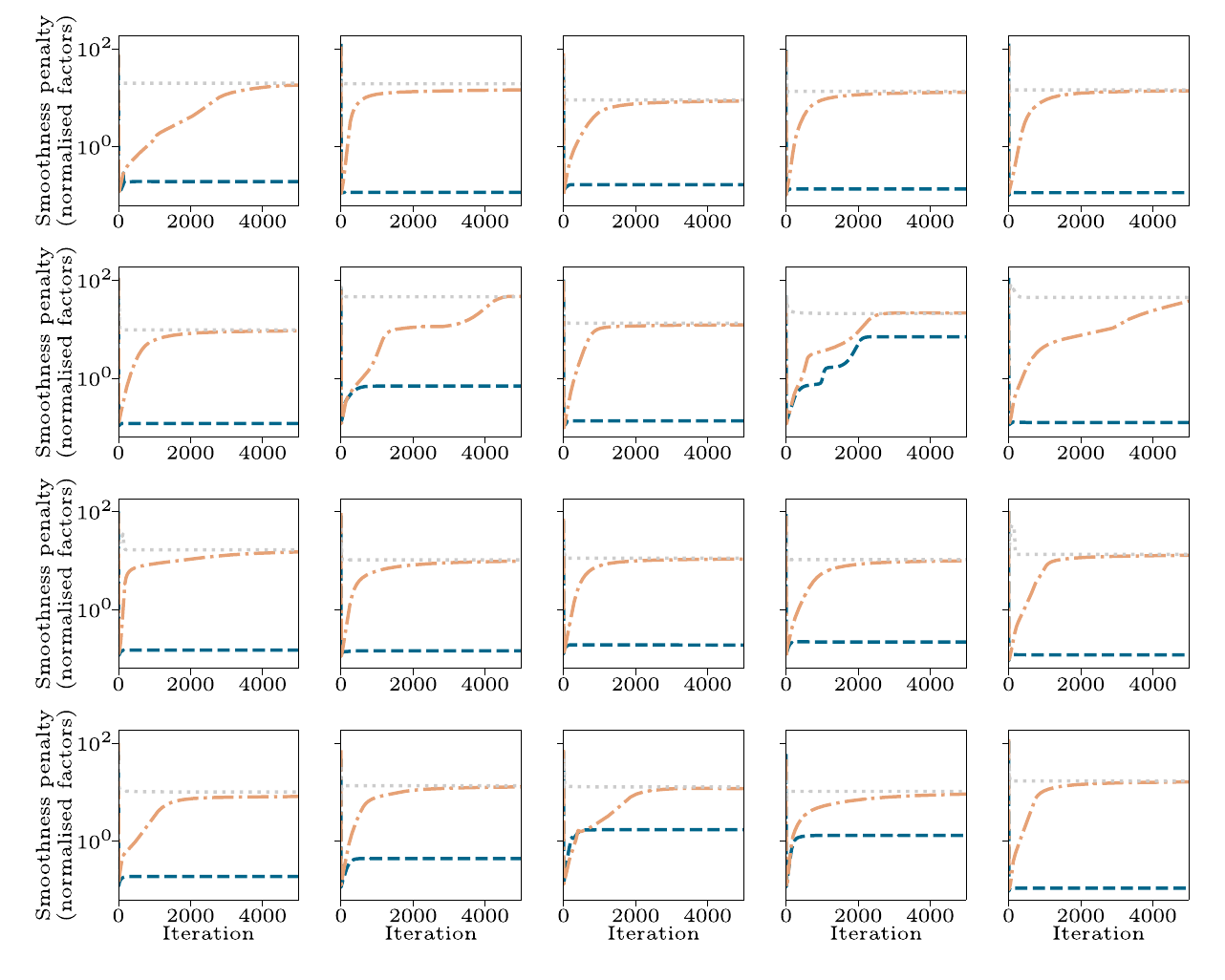}
    \includegraphics{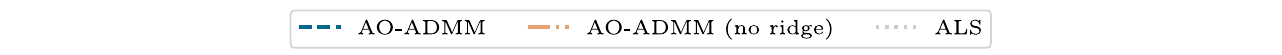}
    \caption{Setup 4: Graph Laplacian penalty of components (after normalizing) as a function of iteration number.  Each plot corresponds to one dataset.}
    \label{sup.fig:sim.smooth.reg.convergence}
\end{figure}

\begin{figure}
    \centering
    \includegraphics{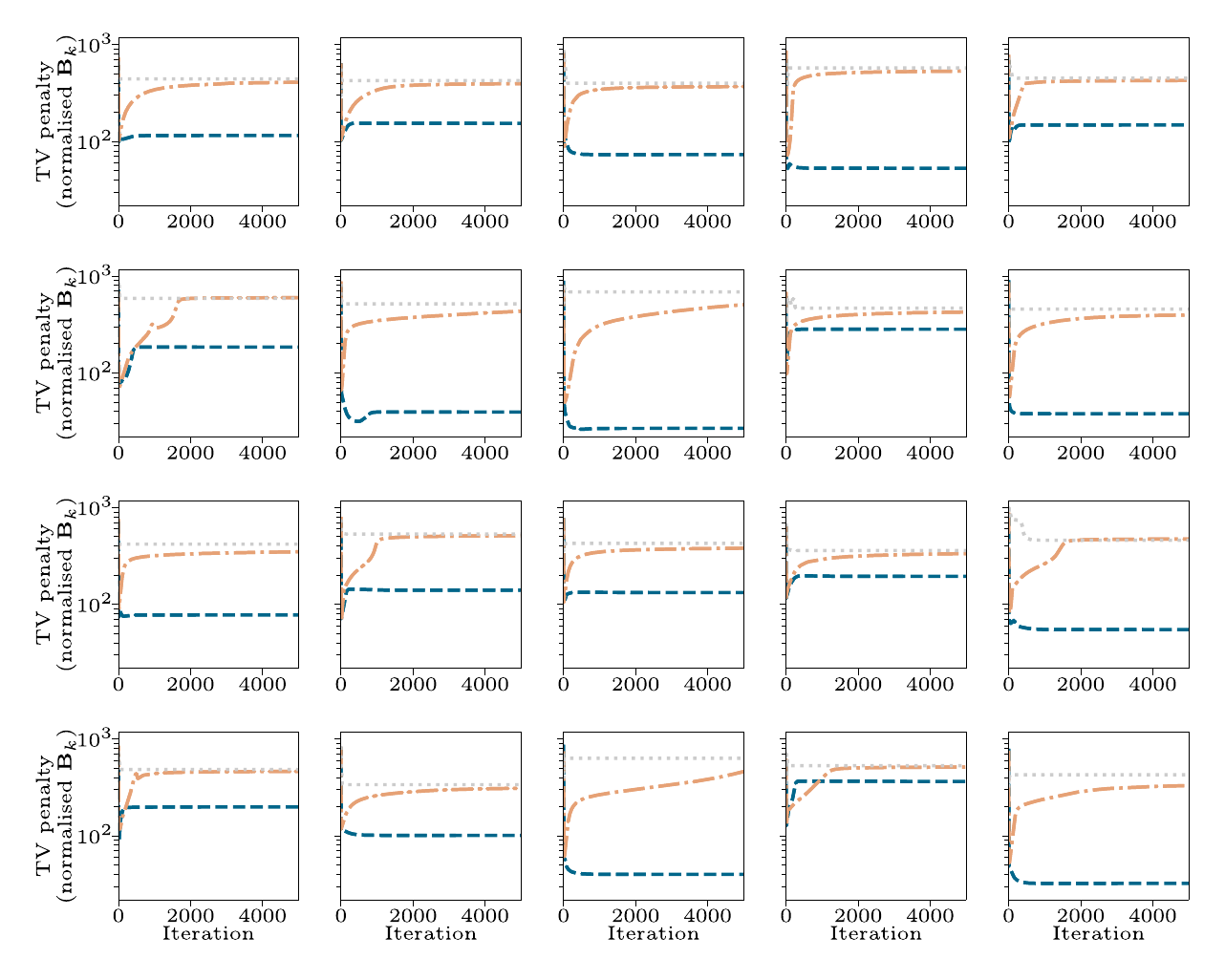}
    \includegraphics{figures/supplement/M145003_SUPPLEMENT_setup45_convergence_legend.pdf}
    \caption{Setup 5: Total variation of components (after normalizing) as a function of iteration number.  Each plot corresponds to one dataset.}
    \label{sup.fig:sim.tv.reg.convergence}
\end{figure}

\begin{figure}
    \centering
    \begin{subfigure}{0.49\textwidth}
     \includegraphics{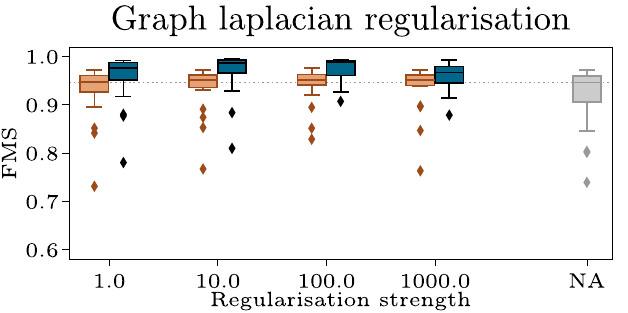}
    \end{subfigure}
    \begin{subfigure}{0.49\textwidth}
     \includegraphics{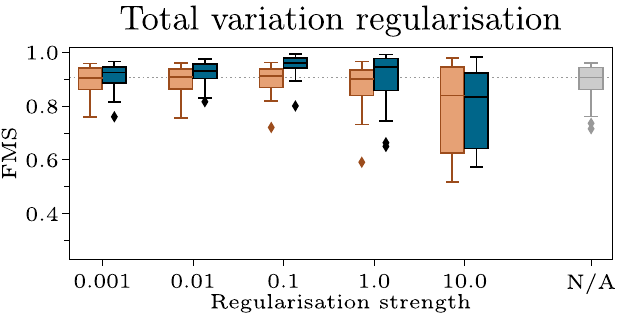}
    \end{subfigure}
     \includegraphics{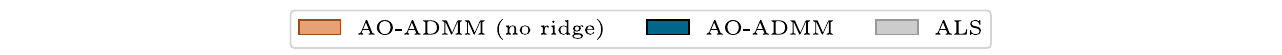}
    \caption{Setup 4 and 5: The FMS for the different models fitted with penalty based regularization when the initialization that obtained the highest overall FMS is chosen for each simulated dataset.}
    \label{sup.fig:sim.penalty.boxplot}
\end{figure}

In \cref{sup.tab:sim.smooth.convergence,sup.tab:sim.tv.convergence}, we see the number of initializations that converged and the number of simulated datasets where at least one initialization converged for the models fitted with graph Laplacian regularization and TV regularization, respectively. Also, in \cref{sup.fig:sim.smooth.reg.convergence,sup.fig:sim.tv.reg.convergence}, we see the regularization penalty (after normalizing the \(\Bk\)-matrices) as a function of iteration number for the selected initialization for each simulated dataset. Once an algorithm has converged, its final value is set as constant for the rest of the iterations, making it easier to compare the behavior of the ALS algorithm and AO-ADMM algorithm without ridge penalty on \(\A\) and \(\DAll\). Finally, in \cref{sup.fig:sim.penalty.boxplot}, we see a boxplot depicting the FMS of the model fitted with graph Laplacian regularization and TV regularization when we select the initialization that obtained the highest overall FMS.

\begin{table}
    \centering
    \caption{Setup 4 and 5: Number of converged initializations and datasets with at least one converged initialization for the models fitted with graph Laplacian regularization, the \emph{Successful decompositions} column shows the number of datasets where at least one initialization converged. Convergence for ALS is measured in the same way using the same tolerance as AO-ADMM (with no feasibility gaps due to ALS not being a splitting method).}
    \label{sup.tab:sim.smooth.convergence}
    \begin{tabular}{llrr}
    \toprule
        &         &  Converged init. &  Successful decompositions \\
    Method & Smooth &                     &            \\
    \midrule
    ADMM - Ridge=0.0 &  1    &                 0/400 &          0/20 \\
         &  10   &                 0/400 &          0/20 \\
         &  100  &                 0/400 &          0/20 \\
         &  1000 &                 0/400 &          0/20 \\
    ADMM - Ridge=0.1 &  1    &               272/400 &         19/20 \\
         &  10   &               255/400 &         18/20 \\
         &  100  &               261/400 &         18/20 \\
         &  1000 &               261/400 &         17/20 \\
    ALS & N/A    &               393/400 &         20/20 \\
    \bottomrule
    \end{tabular}
\end{table}

\begin{table}
    \centering
    \caption{Setup 4 and 5: Number of converged initializations and datasets with at least one converged initialization for the models fitted with TV regularization, the \emph{Successful decompositions} column shows the number of datasets where at least one initialization converged. Convergence for ALS is measured in the same way using the same tolerance as AO-ADMM (with no feasibility gaps due to ALS not being a splitting method).}
    \label{sup.tab:sim.tv.convergence}
    \begin{tabular}{llrr}
\toprule
                 &         &     Converged init. &  Successful decompositions \\
Method & Reg &                     &            \\
\midrule
ADMM - Ridge=0.0 &  0.001  &               0/400 &       0/20 \\
                 &  0.01  &                0/400 &       0/20 \\
                 &  0.1  &                 0/400 &       0/20 \\
                 &  1  &                   0/400 &       0/20 \\
                 &  10 &                   0/400 &       0/20 \\
ADMM - Ridge=0.1 &  0.001  &              38/400 &       7/20 \\
                 &  0.01  &              198/400 &      16/20 \\
                 &  0.1  &               230/400 &      20/20 \\
                 &  1  &                 199/400 &      18/20 \\
                 &  10 &                  54/400 &       8/20 \\
ALS              & N/A  &                398/400 &      20/20 \\
\bottomrule
\end{tabular}
\end{table}

\clearpage
\printbibliography